\documentclass[letterpaper]{article} 
\usepackage[draft]{aaai25}  
\usepackage{times}  
\usepackage{helvet}  
\usepackage{courier}  
\usepackage[hyphens]{url}  
\usepackage{graphicx} 
\usepackage{natbib}  
\usepackage{caption} 
\usepackage{algorithm}
\usepackage{algorithmic}
\usepackage{amsmath}

\usepackage{newfloat}
\usepackage{listings}
\DeclareCaptionStyle{ruled}{labelfont=normalfont,labelsep=colon,strut=off} 
\floatstyle{ruled}
\newfloat{listing}{tb}{lst}{}
\floatname{listing}{Listing}
\usepackage{booktabs}
\usepackage{subcaption}
\usepackage{amsfonts}
\usepackage{xcolor}
\usepackage[symbol]{footmisc}

\title{CoRe: Context-Regularized Text Embedding Learning for \\ Text-to-Image Personalization}
\author{
    Feize Wu\textsuperscript{\rm 1}\footnotemark[1],
    Yun Pang\textsuperscript{\rm 1}\footnotemark[1],
    Junyi Zhang\textsuperscript{\rm 1}\footnotemark[1],
    Lianyu Pang\textsuperscript{\rm 1}\footnotemark[1],
    Jian Yin\textsuperscript{\rm 1}, \\
    Baoquan Zhao\textsuperscript{\rm 1},
    Qing Li\textsuperscript{\rm 2},
    Xudong Mao\textsuperscript{\rm 1}\footnotemark[2]
}
\affiliations{
    \textsuperscript{\rm 1}Sun Yat-sen University\ \ \ 
    \textsuperscript{\rm 2}The Hong Kong Polytechnic University
}

\usepackage{bibentry}

\begin{document}

\maketitle
\footnotetext[1]{Equal contribution.}
\footnotetext[2]{Corresponding author (xudong.xdmao@gmail.com).}

%

\begin{abstract}
Recent advances in text-to-image personalization have enabled high-quality and controllable image synthesis for user-provided concepts. However, existing methods still struggle to balance identity preservation with text alignment. Our approach is based on the fact that generating prompt-aligned images requires a precise semantic understanding of the prompt, which involves accurately processing the interactions between the new concept and its surrounding context tokens within the CLIP text encoder. To address this, we aim to embed the new concept properly into the input embedding space of the text encoder, allowing for seamless integration with existing tokens. We introduce Context Regularization (CoRe), which enhances the learning of the new concept's text embedding by regularizing its context tokens in the prompt. This is based on the insight that appropriate output vectors of the text encoder for the context tokens can only be achieved if the new concept's text embedding is correctly learned. CoRe can be applied to arbitrary prompts without requiring the generation of corresponding images, thus improving the generalization of the learned text embedding. Additionally, CoRe can serve as a test-time optimization technique to further enhance the generations for specific prompts. Comprehensive experiments demonstrate that our method outperforms several baseline methods in both identity preservation and text alignment. Code will be made publicly available.
\end{abstract}

\section{Introduction}
Text-to-image personalization involves adapting a pre-trained diffusion model to generate novel images based on user-provided concepts and text prompts. The goal of personalization techniques is to produce high-quality images that not only accurately preserve the concept's identity but also align well with the text prompt. However, balancing the trade-off between identity preservation and text alignment remains a core challenge in personalization of diffusion models.

\begin{figure}[h]
 \centering
 \includegraphics[width=1.\linewidth]{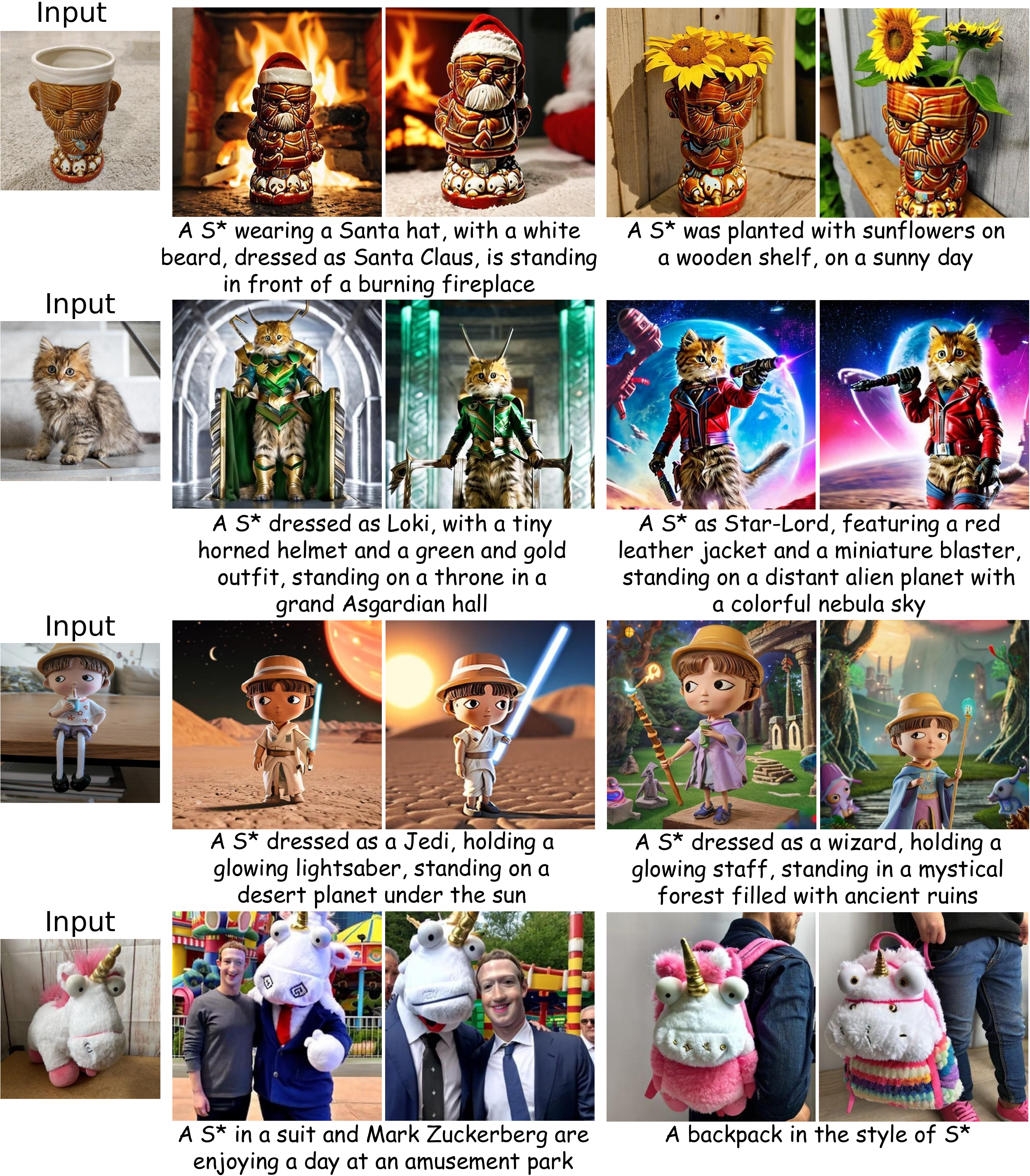}
\caption{CoRe enables text-aligned personalized generations, allowing for high visual variability of the user-provided concept.}

\label{fig:teaser}
\vspace{-0.3cm}
\end{figure}
In this work, we focus on improving text alignment for text-to-image personalization. Our investigation is based on the fact that a precise semantic understanding of the prompts is the premise for aligning the generated images with the prompts. The semantic understanding of the prompts is managed by the CLIP text encoder, which involves processing the text embeddings of all tokens and their interactions. Therefore, we aim to learn a proper text embedding for the new concept, which not only accurately represents the concept but also seamlessly integrates with existing tokens.

\begin{figure*}[h]
    \centering
    \hspace{-0.016\linewidth} 
    \begin{subfigure}[b]{0.50\linewidth}
        \centering
        \vfill
    \includegraphics[width=\linewidth]{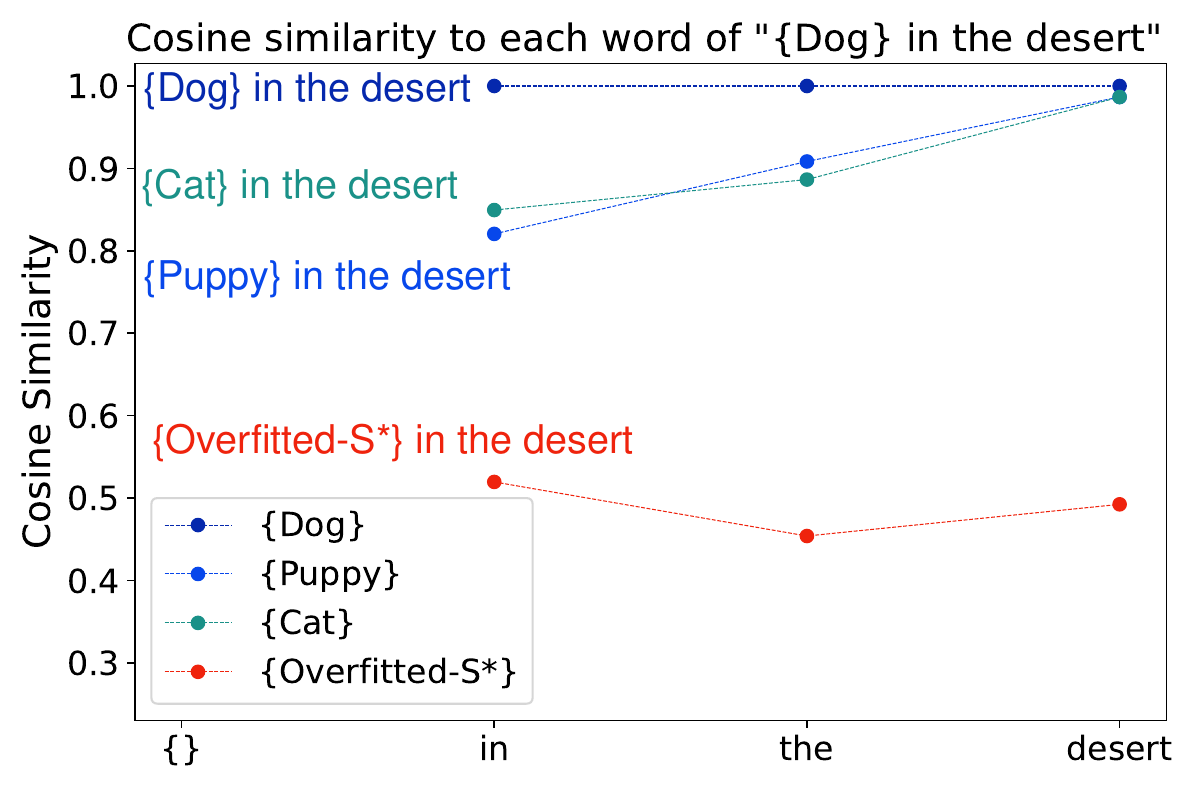}
        \label{fig:sub1}
    \end{subfigure}
    \hfill
    \begin{subfigure}[b]{0.50\linewidth}
        \centering
        \raisebox{0.022\height}{
        \includegraphics[width=\linewidth]{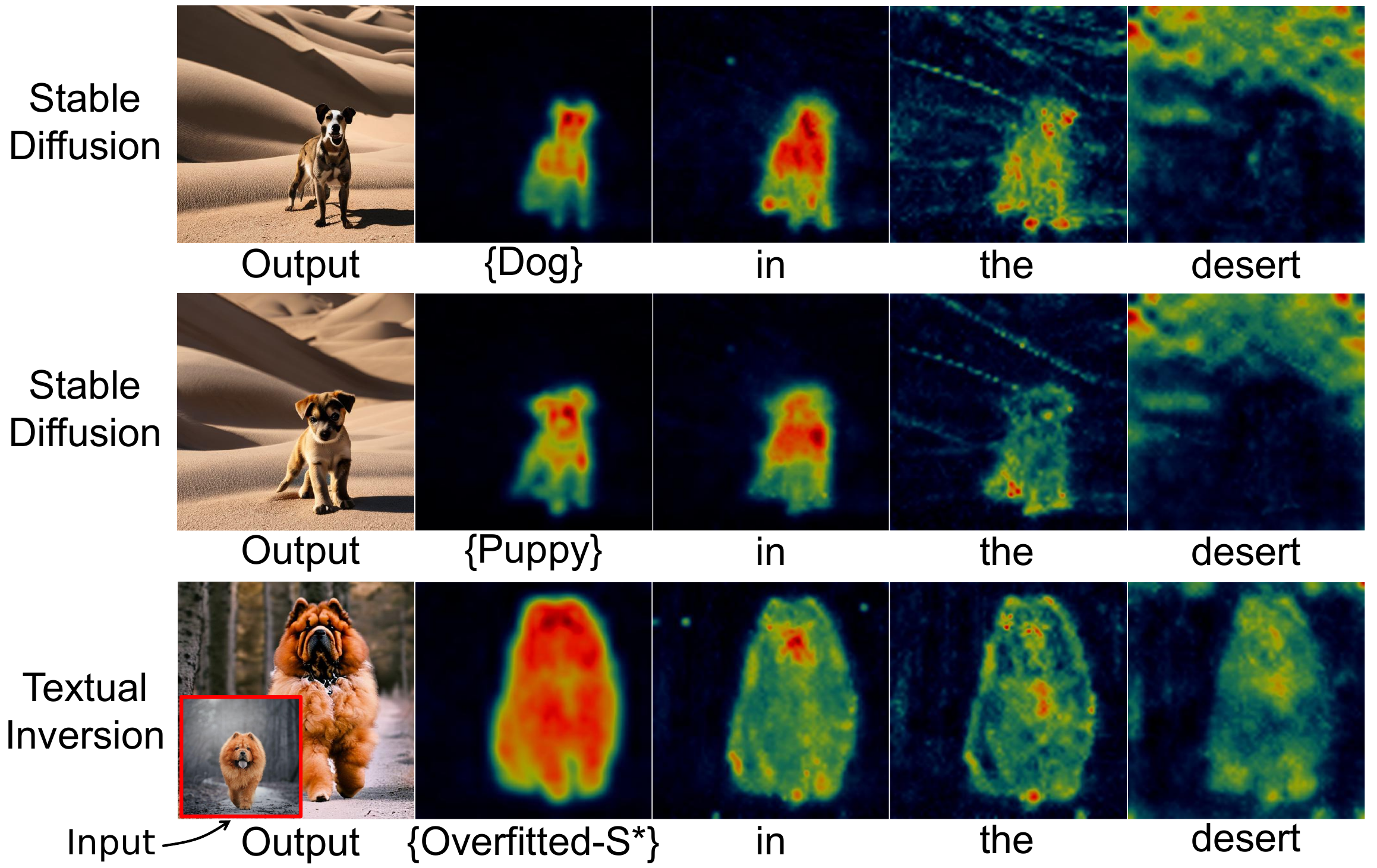}}
        \label{fig:sub2}
    \end{subfigure}
    \vspace{-0.4cm}
    \caption{For the four similar prompts (``\{\} in the desert''), we show the cosine similarity between the output embeddings of each token (left), and the cross-attention map visualization of each token (right). Replacing ``dog'' with ``puppy'' or ``cat'' results in similar output embeddings and attention maps for other tokens. In contrast, using the overfitted $S_*$ by Textual Inversion significantly alters the output embeddings and attention maps for other tokens.}
    \label{fig:motivation}
\end{figure*}

Instead of investigating the text embedding of the new concept itself~\cite{alaluf2023neural,pang2024attndreambooth}, we shift our focus to the context tokens surrounding the new concept in prompts. Here, the term \textit{text embedding} refers to the input to the CLIP text encoder. Moreover, for clarity and following the terminology used in~\cite{lu2022prompt}, we refer to the embeddings before and after the CLIP text encoder also as the \textit{input embedding} and \textit{output embedding}, respectively. As illustrated in Figure~\ref{fig:motivation}, consider a scenario where the object token in a prompt is switched from ``dog'' to ``cat''; the output embeddings of the other tokens largely remain consistent. However, using an overfitted text embedding by Textual Inversion~\cite{textual-inversion} significantly alters the output embeddings of its context tokens. This alteration occurs because the overfitted embedding adversely affects the output of the context tokens within the text encoder. As shown, these inappropriate output embeddings subsequently lead to incorrect allocations in the attention maps for the context tokens.

Based on these observations, we introduce a new method named Context Regularization (CoRe), which enhances text embedding learning for a new concept by regularizing its context tokens. CoRe can improve the compatibility of the new concept's text embedding, thereby facilitating a more precise semantic understanding of the prompt. As indicated in Figure~\ref{fig:motivation}, replacing the object token ``dog'' with a compatible embedding of the new concept should yield similar output embeddings for the context tokens. Therefore, we propose a regularization strategy that encourages the output embeddings of the context tokens to align with those from a reference prompt containing a super-category token. As the attention maps play a crucial role in generation, we also impose constraints on the attention maps for the context tokens. We avoid imposing constraints directly between the new concept and its super-category, due to the typically substantial differences between them.

As our proposed CoRe is applied only to the output embeddings and attention maps, without the need for generating images, it can be used with arbitrary prompts. Therefore, we construct a regularization prompt set that covers a broad range of prompts to improve the generalization of the new concept's text embedding. During training, a prompt is randomly selected from this set for regularization purposes. Moreover, at test time, CoRe can serve as a test-time optimization approach to further refine the generations for specific prompts.

We demonstrate the effectiveness of our method by comparing it with four state-of-the-art personalization methods through both qualitative and quantitative evaluations. Our method shows superior performance in identity preservation and text alignment compared to the baselines, especially for prompts requiring high visual variability. Moreover, in addition to personalizing general objects, our method also works well for face personalization, generating more identity-preserved face images compared to three recent face personalization methods.

\section{Related Work}
\paragraph{Text-to-Image Generation.} 
Text-to-image generation involves creating visual images from textual prompts, a task that has seen significant advances with diffusion models~\cite{diffusion,ddpm,improved_ddpm}. To achieve high-resolution text-to-image generation, various methods have been developed. DALL-E 2~\cite{dalle2} proposes converting a CLIP text embedding into a CLIP image embedding, and then transforming this image embedding into the final image. Imagen~\cite{imagen} employs a cascaded diffusion model that learns to generate images from low-resolution to high-resolution. Latent diffusion models~\cite{stable-diffusion} utilize an autoencoder to map the image into a lower-dimensional latent representation, performing progressive denoising steps in this latent space.

\paragraph{Text-to-Image Personalization.} 
Text-to-image personalization focuses on adapting pre-trained diffusion models to incorporate new concepts with a few user-provided images. Textual Inversion~\cite{textual-inversion} inverts the new concept into the text embedding space, while DreamBooth~\cite{dreambooth} fine-tunes the entire U-Net to learn the new concept. Custom Diffusion~\cite{kumari2022customdiffusion} fine-tunes the text embedding and a few parameters in the U-Net, enabling fast tuning for multi-concept customization. Recent advancements in text-to-image personalization have significantly improved identity preservation~\cite{voynov2023p,alaluf2023neural,zhou2023enhancing} and text alignment~\cite{tewel2023keylocked,qiu_oft}. Some works~\cite{arar2024palp,huang2024realcustom} enhance text alignment by optimizing generations for specific prompts at test time. Moreover, tuning-free approaches~\cite{wei2023elite, shi2023instantbooth, li2023blipdiffusion,ye2023ipa} focus on accelerating the personalization process. Additionally, many studies~\cite{xiao2023fastcomposer,wang2024instantid,li2023photomaker,wang2024stableidentity} concentrate on the personalized synthesis of widely interested human faces.

\paragraph{Text Embedding Learning.}
Customizing a concept by inverting it into the text embedding space was first introduced in Textual Inversion~\cite{textual-inversion}. XTI~\cite{voynov2023p} extends this space to be more expressive by using multiple tokens, assigning one token per attention layer. NeTI~\cite{alaluf2023neural} further expands the text embedding space to depend on both the denoising timestep and the U-Net layer. E4T~\cite{gal2023encoderbased} employs an encoder-based tuning method that efficiently inverts the concept into the text embedding space. AttnDreamBooth~\cite{pang2024attndreambooth} suggests optimizing the text embedding for the new concept with very few steps, as it is prone to overfitting. A concurrent work, ClassDiffusion~\cite{huang2024classdiffusion}, also utilizes a super-category token to guide the learning of text embeddings for the new concept. Our work differs in several aspects. First, we use context tokens to indirectly regularize the text embedding learning without including the super-category token, whereas concurrent work focuses on narrowing the gap between the new concept and its super-category. Second, we further regularize the attention maps of the context tokens. Third, we construct a regularization prompt set to cover a broad range of prompts, making the learned text embeddings more generalizable.

\paragraph{Cross-Attention Control.}
The control of the cross-attention maps has demonstrated the effectiveness in image synthesis for diffusion models~\cite{AttendandExcite}. Several studies~\cite{Jin2023Image, Nam2024DreamMatcher,ma2023subjectdiffusionopen,Zhang2024compositional} have also investigated controlling the attention maps for text-to-image personalization. Custom Diffusion~\cite{kumari2022customdiffusion} illustrates how incorrect attention maps can lead to failed compositions involving multiple concepts. ViCo~\cite{Hao2023ViCo} proposes regularizing the attention maps to focus on meaningful regions. Break-A-Scene~\cite{Avrahami2023BreakAScene} enhances the generation of multiple concepts by using segmentation masks to guide the learning of the attention maps.

\begin{figure*}[htbp]
 \centering
 \includegraphics[width=1\linewidth]{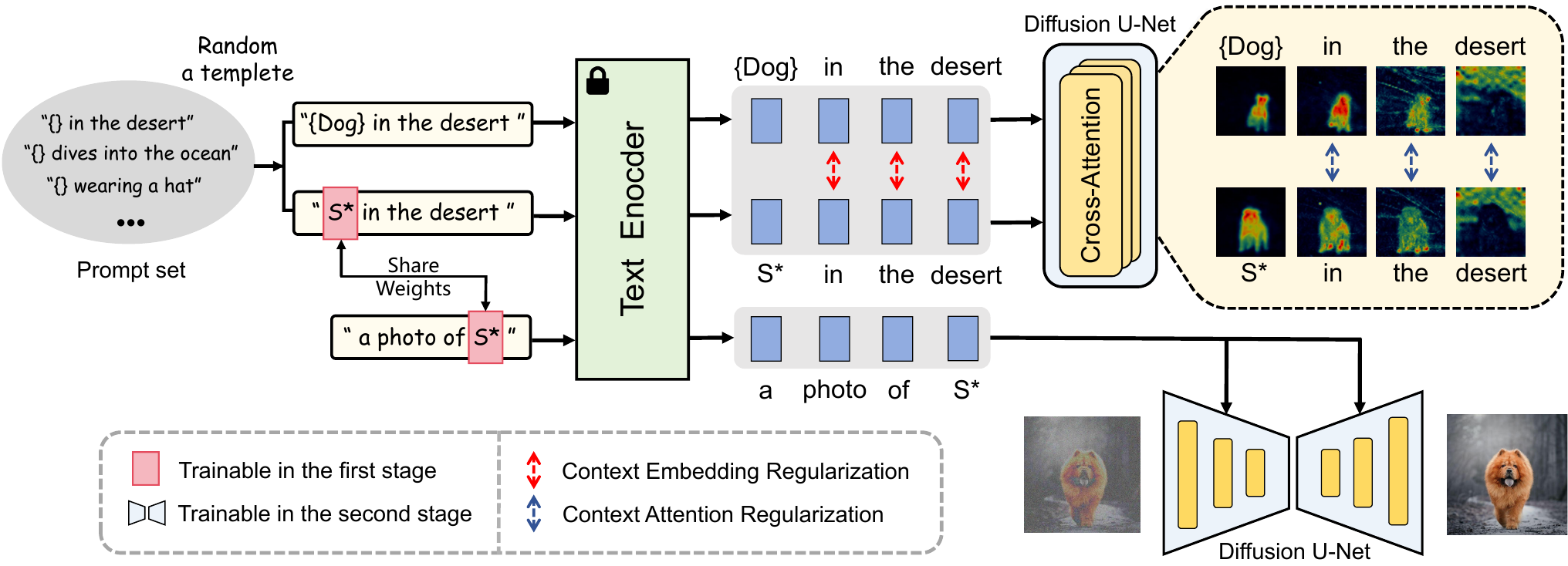}
\caption{Overview of the proposed CoRe. Our method enhances the text embedding learning for $S_*$ by regularizing its context tokens. Specifically, we randomly select a regularization prompt (e.g., ``$S_*$ in the desert'') and a reference prompt (e.g., ``Dog in the desert'') from the prompt set. During training, the proposed context embedding regularization and context attention regularization are applied together with the diffusion loss, which encourages the representations of the context tokens surrounding $S_*$ to align with those in the reference prompt. These regularization terms make the text embedding of $S_*$ more compatible with existing tokens.}
\label{fig:method_pic}
\end{figure*}

\section{Preliminaries}
\paragraph{Latent Diffusion Models.}
In Latent Diffusion Model (LDM)~\citep{stable-diffusion}, an encoder $\mathcal{E}$ transforms an image $x$ into a latent representation $z = \mathcal{E}(x)$ in lower-dimensional space, and a decoder $\mathcal{D}$ reconstructs the image from this latent code, i.e., $\mathcal{D}(\mathcal{E}(x)) \approx x$. Furthermore, a denoising diffusion probabilistic model~\citep{ddpm} is utilized to generate latent codes within the autoencoder's latent space. To create images from textual descriptions, the model relies on a conditioning input vector $c(y)$, which is derived from the given text prompt $y$. The training objective of LDM is expressed as follows: 
\begin{equation}
  \mathcal{L}_{\text{diffusion}}={E}_{z \sim \mathcal{E}(x), y, \varepsilon \sim 
  \mathcal{N}(0,1), t}\left[\left\|\varepsilon-\varepsilon_{\theta}\left(z_{t}, t, c(y)\right)\right\|_{2}^{2}\right],
  \label{eq:ldm}
\end{equation}
where the denoising network $\varepsilon_\theta$ is used to remove the noise added to the latent code given the noised latent $z_t$, the timestep $t$ and the conditioning vector $c(y)$.

\paragraph{Textual Inversion.}
Given several image examples of a target concept, Textual Inversion (TI)~\cite{textual-inversion} learns the concept by inverting it into the text embedding space. TI introduces a new token $S_*$ and a corresponding embedding $v_*$. During the learning process, $v_*$ is optimized to minimize the diffusion loss (Eq.~\ref{eq:ldm}) as follows:
\begin{equation}
    v_* = \arg\min_v {E}_{z, y, \epsilon, t} \left[ \left\| \varepsilon - \varepsilon_\theta(z_t, t, c(y, v)) \right\|_2^2 \right],
  \label{eq:ti_objective}
\end{equation}
where $c(y, v)$ denotes the the conditioning vector using the optimized text embedding $v$.

\paragraph{DreamBooth.}
DreamBooth~\citep{dreambooth} fine-tunes the entire U-Net of the diffusion model to learn the target concept. It employs a rarely used token to represent the concept and fixes its text embedding during optimization. Since the entire U-Net and possibly the text encoder are fine-tuned, DreamBooth usually achieves better identity preservation than Textual Inversion.

\section{Method} 
\subsection{Text Embedding Learning with CoRe} 
To achieve text-aligned generations, we aim to learn an appropriate text embedding for the new concept that is compatible with and seamlessly integrates into existing tokens. This is because text-aligned generations depend on a precise semantic understanding of the prompt, which in turn depends on the correct interactions between the new concept and the other tokens. Instead of directly improving the new concept's embedding, we focus on constraining the context tokens surrounding the new concept. Our method derives from two key insights. First, proper output embeddings of the context tokens can only be achieved if the new concept’s input embedding is correctly learned; otherwise, it adversely impacts the output embeddings of the context tokens within the text encoder. Second, when replacing the object token in a prompt with another, the output embeddings and attention maps of the context tokens should largely remain consistent. We verify these insights through experiments illustrated in Figure~\ref{fig:motivation}. For instance, in the prompt ``dog in the desert'', replacing ``dog'' with an overfitted embedding by Textual Inversion significantly alters the output embeddings and attention maps of the other tokens. In contrast, replacing ``dog'' with ``cat'' maintains the consistency of the output embeddings and attention maps.

Based on these insights, we propose Context Regularization (CoRe) that enhances the text embedding learning for the new concept by regularizing its context tokens. For a training prompt containing the new concept, we construct a reference prompt by replacing the new concept token with a super-category token. We then enforce a similarity constraint on the output embeddings and attention maps of the context tokens between these two prompts. It is important to note that our context regularization can be used with arbitrary prompts because it is applied only to the output embeddings and attention maps, without the need for generating images. Therefore, we construct a regularization prompt set designed to cover a broad range of prompts, with details provided in the Appendix.

\begin{figure*}[t]
 \centering
 \includegraphics[width=.94\linewidth]{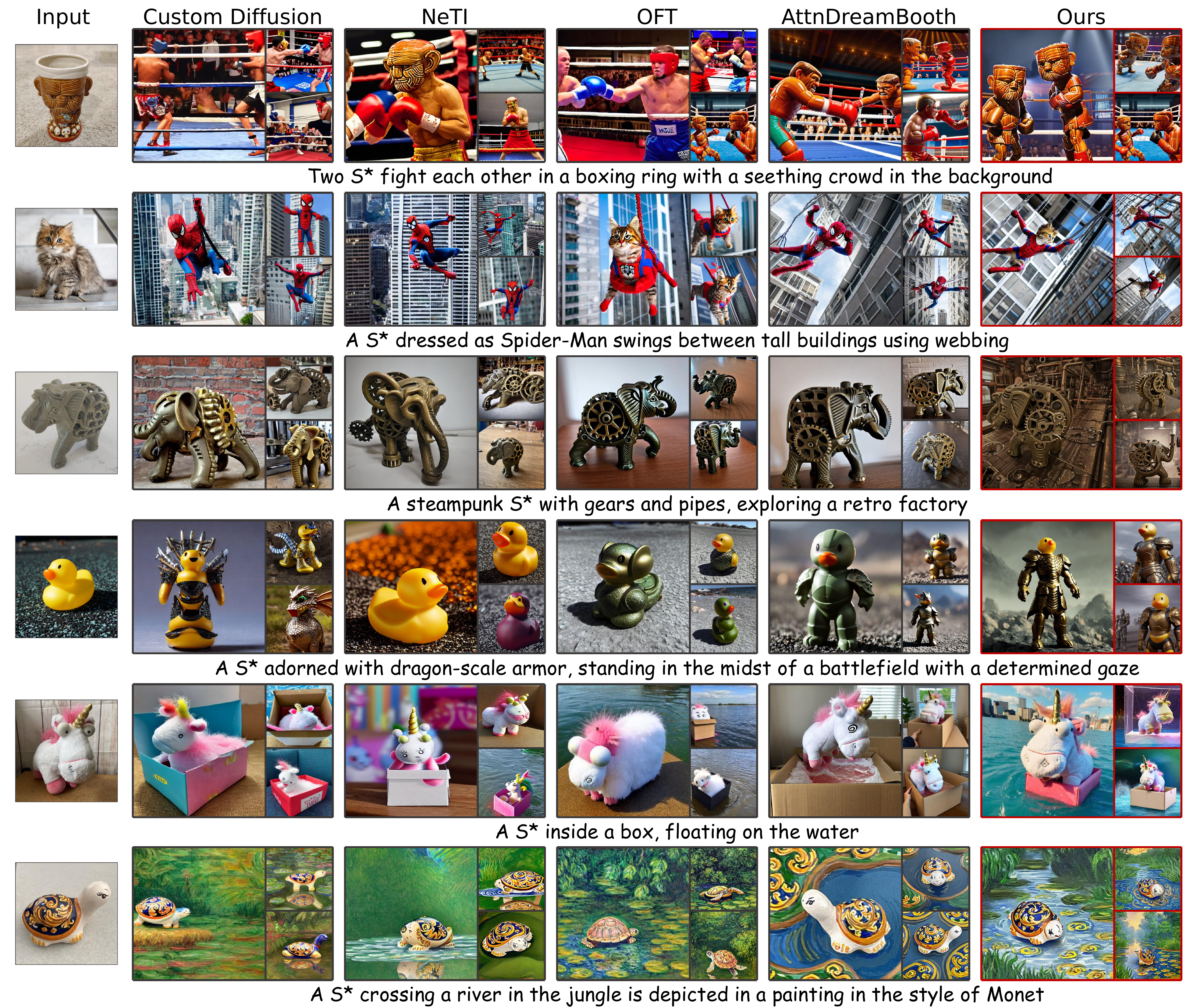}
\vspace{-0.1cm}
\caption{Qualitative comparison. We present personalization results of our method and four baseline methods, including Custom Diffusion~\cite{kumari2022customdiffusion}, NeTI~\cite{alaluf2023neural}, OFT~\cite{qiu_oft}, and AttnDreamBooth~\cite{pang2024attndreambooth}. Our method demonstrates superior performance in both text alignment and identity preservation compared to these baselines, especially for the prompts that require high visual variability of the concept.}

\label{fig:qualitative_comparsion}
\vspace{-0.3cm}
\end{figure*}

\paragraph{Context Embedding Regularization.}
Formally, we randomly select a prompt template (e.g., ``a \{\} in the jungle'') from the regularization prompt set, and fill it with the new concept token and its super-category token, respectively, producing a pair of prompts $y_*$ and $y_\text{cat}$ (e.g., ``a $S_*$ in the jungle'' and ``a [super-category] in the jungle''). The input embeddings of these two prompts, $\{v_i\}$ and $\{v_i’\}$, are then fed into the text encoder $E$, producing corresponding output embeddings $\{E(v_i)\}$ and $\{E(v_i')\}$. We minimize the average cosine distance between these two sets of output embeddings:
\begin{equation}
  \mathcal{L}_{\text{emb}}=\frac{1}{n-1}\sum_{i=1,i\not=k}^{n}\big(1-\text{cos}(E(v_i),E(v_i'))\big),
  \label{eq:emb}
\end{equation}
where $k$ is the index corresponding to $S_*$ and [super-category], $n$ is the length of the output embeddings, and $\text{cos}(\cdot,\cdot)$ denotes the cosine similarity. Note that we avoid imposing constraints between $S_*$ and [super-category] due to the observed significant degradation in identity preservation, as the new concept and its super-category usually exhibit substantial differences.

\paragraph{Context Attention Regularization.}
As illustrated in Figure~\ref{fig:motivation}, the overfitted embedding of the new concept subsequently results in incorrect attention maps for the context tokens. Therefore, we utilize attention maps to further regularize the text embedding learning for the new concept. We introduce an additional regularization term that enforces a similarity constraint between the attention maps of the two prompts, $y_*$ and $y_\text{cat}$. Formally, the output embeddings $\{E(v_i)\}$ and $\{E(v_i')\}$ for $y_*$ and $y_\text{cat}$ are fed into the 16 different cross-attention layers of the U-Net, generating 16 attention maps $\{M_i^{1:16}\}$ and $\{M_i'^{1:16}\}$, respectively. We minimize the average squared difference between the mean values of these attention maps as follows:
\begin{equation}
  \mathcal{L}_{\text{attn}}=\frac{1}{n-1}\sum_{i=1,i\not=k}^{n}\bigl(\mu(M_i^{1:16})-\mu(M_i'^{1:16})\bigr)^2,
  \label{eq:attn}
\end{equation}
where $\mu(M^{1:16})$ denotes the mean of all values across the 16 attention maps, $k$ is the index corresponding to $S_*$, and $n$ is the length of the prompt.

Overall, the full optimization objective of our method is defined as:
\begin{equation}
    v_* = \arg\min_{v_k} \mathcal{L}_{\text{diffusion}}+\lambda_{\text{emb}}\mathcal{L}_{\text{emb}}+\lambda_{\text{attn}}\mathcal{L}_{\text{attn}}.
  \label{eq:ti_objective}
\end{equation}

\paragraph{Embedding Rescaling.}
As identified in~\cite{alaluf2023neural,pang2024cross}, during optimization, the scale of the new concept's text embedding tends to become excessively large, leading to significant degradation in text alignment. Inspired by~\cite{alaluf2023neural}, we propose rescaling the norm of the text embedding during optimization to mitigate this issue. Specifically, after one optimization step, we reset the norm of the updated embedding to match the norm from the previous step. The rescaled embedding is given by:
\begin{equation}
  v_*^s = \frac{v_*^s}{\| v_*^s \|} \| v_*^{s-1} \|,
  \label{eq:rescale}
\end{equation}
where $s$ denotes the $s$-th optimization step. In practice, we apply this rescaling strategy only during the intermediate phase of the optimization, as we empirically find that rescaling at the beginning or end phases can lead to degraded identity preservation, likely due to the information loss introduced by rescaling.

\subsection{Embedding-to-Identity Training Strategy}
Solely optimizing the text embedding is insufficient to capture the concept identity. Inspired by~\cite{roich2022pivotal,pang2024attndreambooth}, we propose a two-stage training strategy. Initially, we employ CoRe to learn a text embedding for the new concept that is compatible with existing tokens. This yields an editable embedding but provides a coarse depiction of the concept identity. In the second stage, we freeze the text embedding and fine-tune all layers of the U-Net to precisely capture the concept identity.

\subsection{Test-Time Optimization}
At test time, our proposed method, CoRe, can optionally serve as a test-time optimization technique to enhance generation for specific prompts. Specifically, given a prompt for generation, we refine the output embeddings and attention maps associated with this prompt by performing a few additional optimization steps using CoRe. This refinement is done without employing the diffusion loss. Note that in our experiments, this test-time optimization strategy is not applied when comparing with the baselines to ensure a fair comparison.

\section{Experiments}

\paragraph{Datasets.}
For a comprehensive evaluation, we collect 24 concepts from previous studies~\cite{textual-inversion,dreambooth,kumari2022customdiffusion}. Following ~\cite{tewel2023keylocked}, we categorize these concepts into two groups: animate objects (e.g., ``cat'' and ``child doll'') and inanimate objects (e.g., ``clock'' and ``berry bowl''). Accordingly, we use two sets of prompts for these two groups, respectively. Some prompts are shared across all concepts, including background change, concept color change, and artistic style, while others are specific to animate objects, such as action and outfit change.

\begin{figure}[t]
    \centering
    \renewcommand{\arraystretch}{0.3} 
    \setlength{\tabcolsep}{0.5pt} 

    {\footnotesize
    \begin{tabular}{c@{\hspace{4pt}}c c c c}
         Input &
        \multicolumn{1}{c}{CI} &
        \multicolumn{1}{c}{PM} &
        \multicolumn{1}{c}{FD} &
        \multicolumn{1}{c}{Ours} \\

        \includegraphics[width=0.09\textwidth]{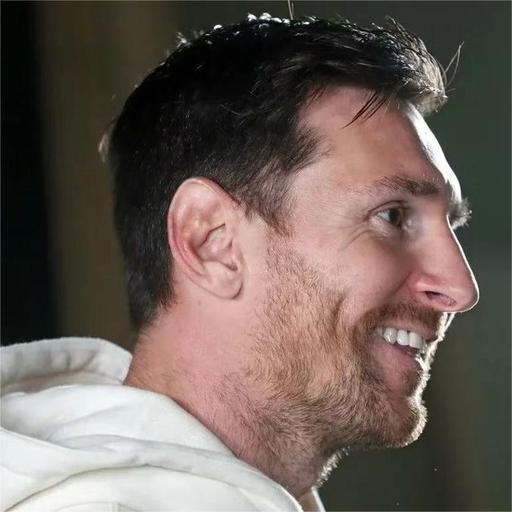} &
        \includegraphics[width=0.09\textwidth]{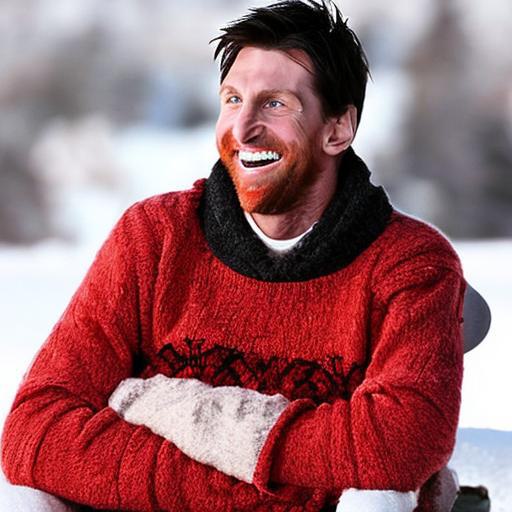} &
        \includegraphics[width=0.09\textwidth]{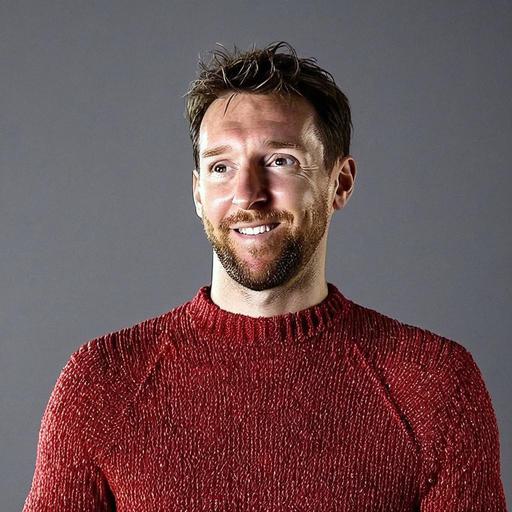} &
        \includegraphics[width=0.09\textwidth]{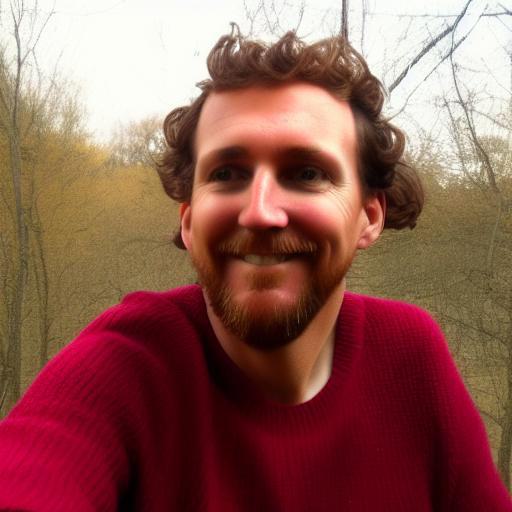} &
        \includegraphics[width=0.09\textwidth]{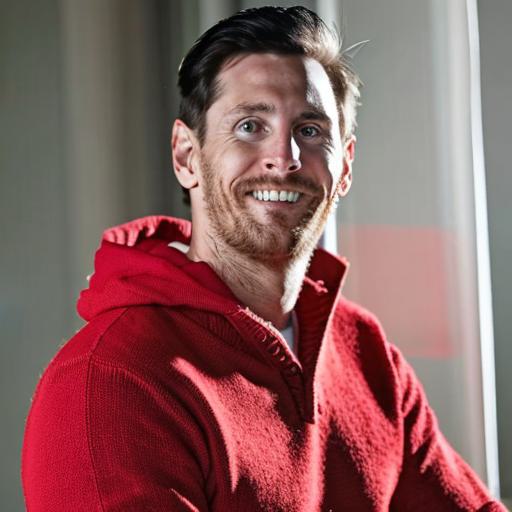} \\
        & \multicolumn{4}{c}{{S*} wearing a red sweater\vspace{0.5mm}} \\

        \includegraphics[width=0.09\textwidth]{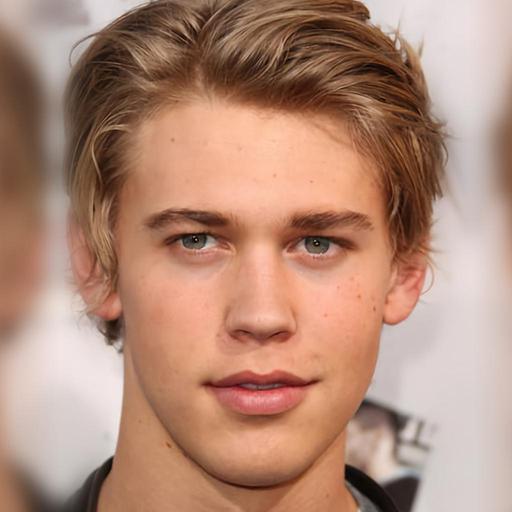} &
        \includegraphics[width=0.09\textwidth]{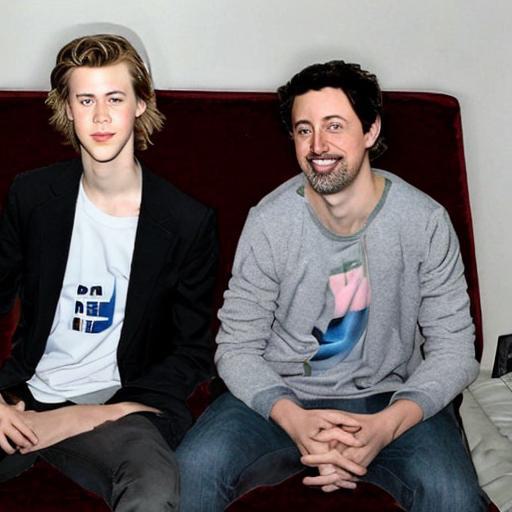} &
        \includegraphics[width=0.09\textwidth]{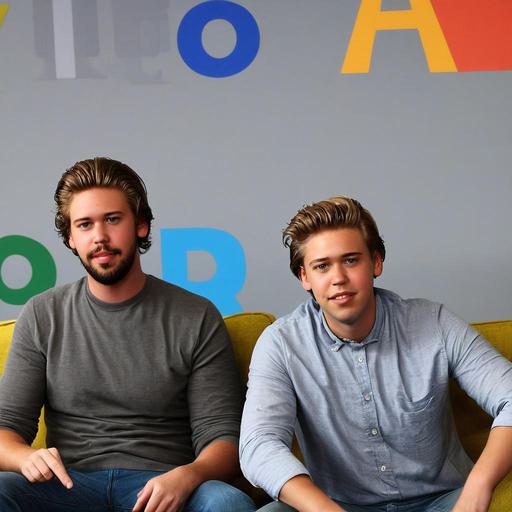} &
        \includegraphics[width=0.09\textwidth]{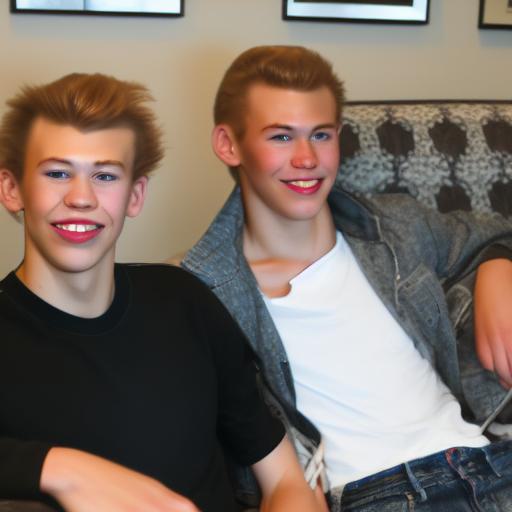} &
        \includegraphics[width=0.09\textwidth]{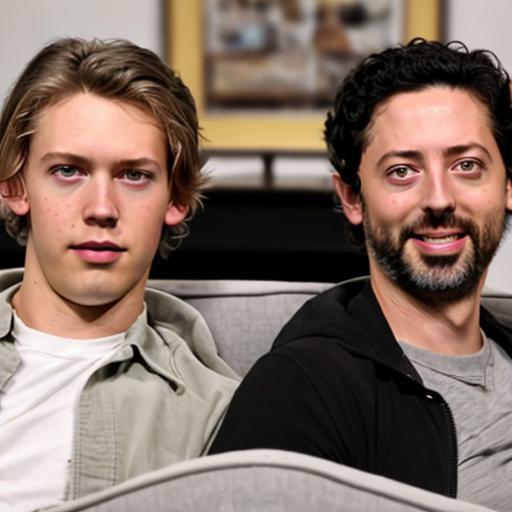} \\
        
        & \multicolumn{4}{c}{ {S*} and Sergey Brin sit on a sofa\vspace{0.5mm}} \\

        \includegraphics[width=0.09\textwidth]{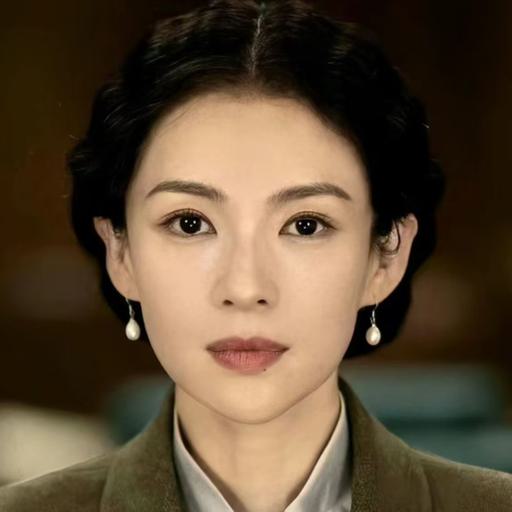} &
        \includegraphics[width=0.09\textwidth]{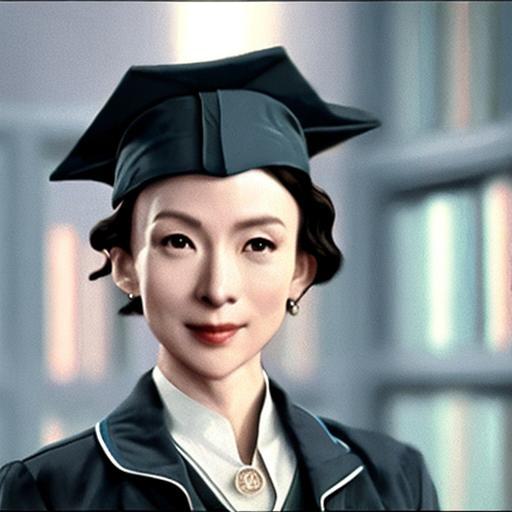} &
        \includegraphics[width=0.09\textwidth]{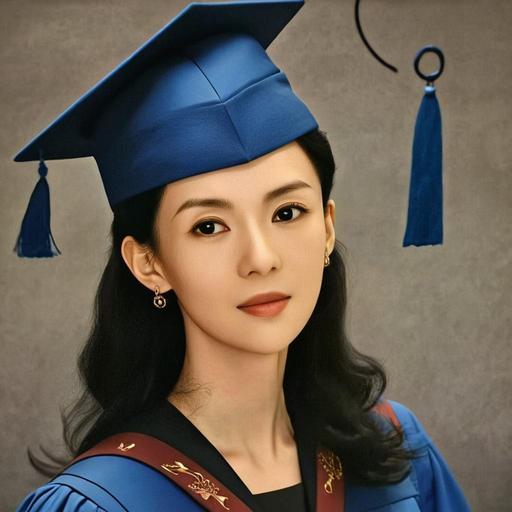} &
        \includegraphics[width=0.09\textwidth]{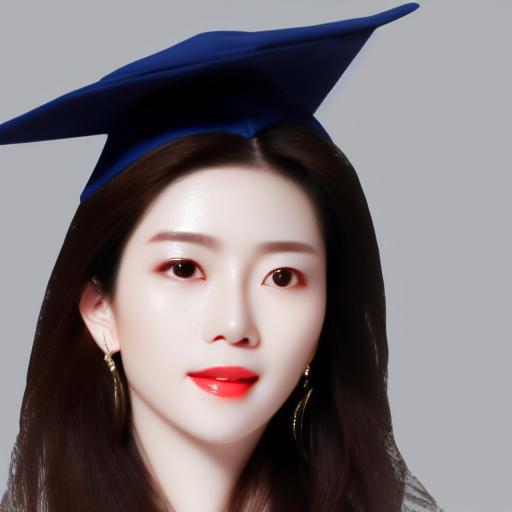} &
        \includegraphics[width=0.09\textwidth]{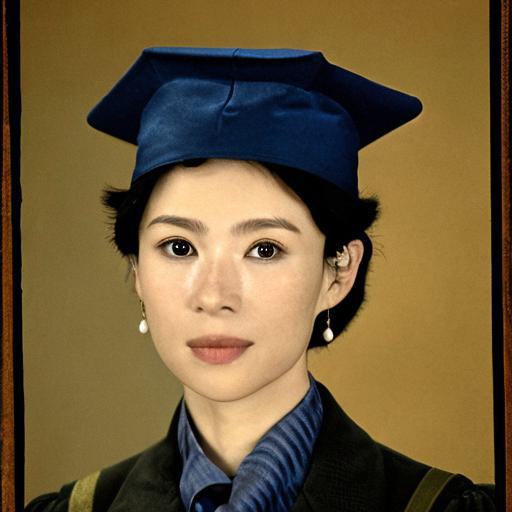} \\
        
        & \multicolumn{4}{c}{ {S*} wearing a doctoral cap\vspace{0.5mm}} \\
        
        \includegraphics[width=0.09\textwidth]{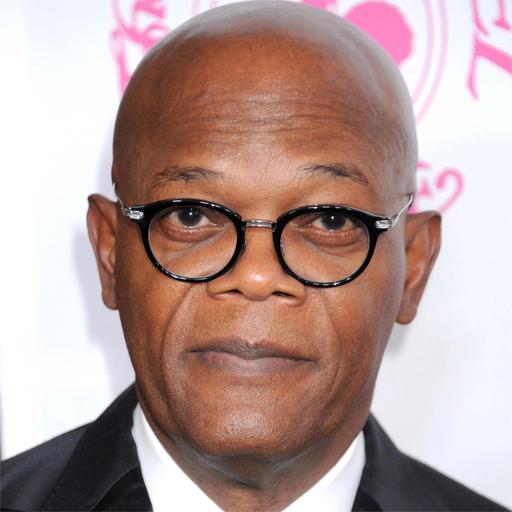} &
        \includegraphics[width=0.09\textwidth]{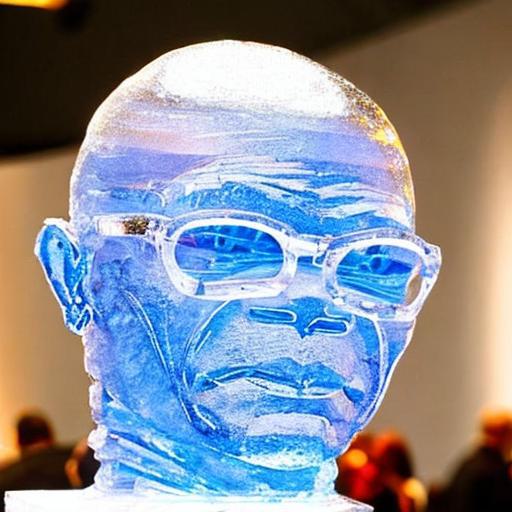} &
        \includegraphics[width=0.09\textwidth]{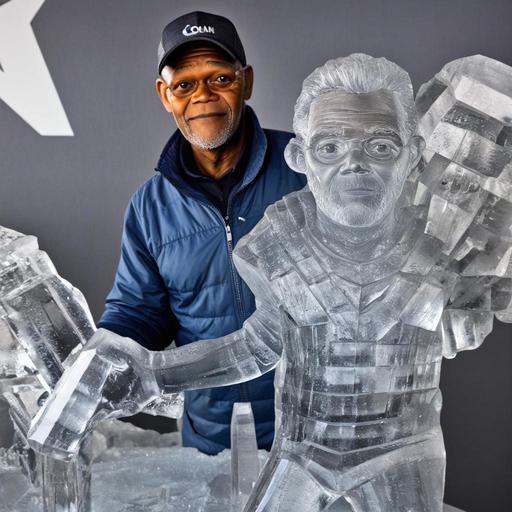} &
        \includegraphics[width=0.09\textwidth]{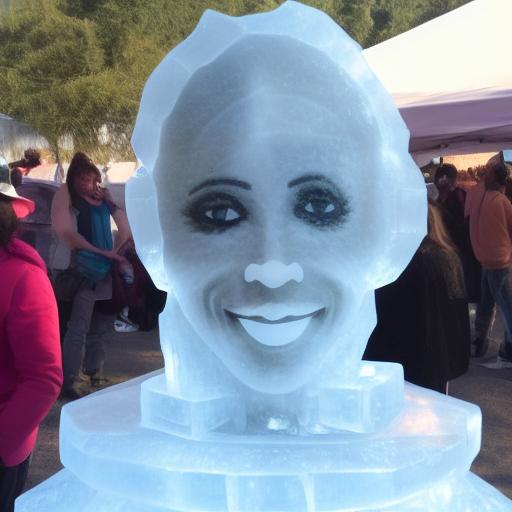} &
        \includegraphics[width=0.09\textwidth]{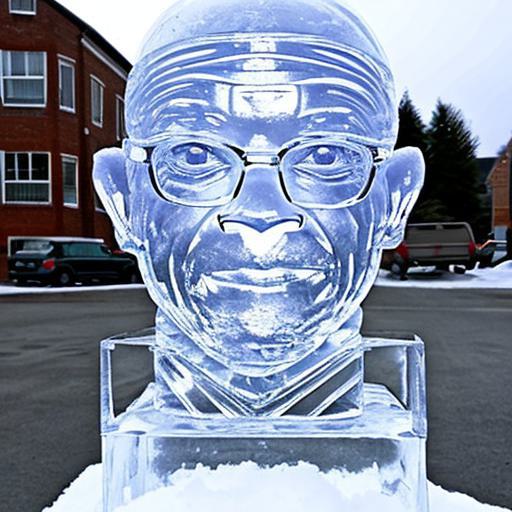} \\
        & \multicolumn{4}{c}{ Ice sculpture of {S*}\vspace{0.5mm}} \\

    \end{tabular}
    }

    \caption{Face personalization results of our method and three baseline methods, including Cross Initialization (CI)~\cite{pang2024cross}, PhotoMaker (PM)~\cite{li2023photomaker}, and Face2Diffusion (FD)~\cite{shiohara2024face2diffusion}. Our method achieves more identity-preserved face generations compared to the baselines, especially when the input image is a side face.}
    \label{fig:face_comparison}
    \vspace{-0.1cm}
\end{figure}

\paragraph{Evaluation Setup.}
We compare our method against four recent baseline methods: Custom Diffusion~\cite{kumari2022customdiffusion}, NeTI~\cite{alaluf2023neural}, OFT~\cite{qiu_oft}, and AttnDreamBooth~\cite{pang2024attndreambooth}. For quantitative evaluation, we employ a set of 20 prompts, detailed in the Appendix, using the following metrics: (1) identity preservation, measured by the visual similarity between the generated and input images in the CLIP-I~\cite{clip} and DINO~\cite{caron2021emerging} feature spaces; and (2) text alignment, measured by the CLIP-T similarity between the generated images and the prompts. Following~\cite{zeng2024jedi}, the CLIP-I and DINO scores are exclusively calculated on foreground-masked images to eliminate background variations and better reflect concept identity similarity. Additionally, prompts involving stylization or outfit change are excluded from the CLIP-I and DINO score calculations because these modifications can significantly alter the concept's appearance. The implementation details of our method and the baselines are provided in the Appendix.

\begin{table}[t]
    \centering
    \caption{Quantitative comparison. CLIP-I and DINO evaluate identity preservation by measuring the similarity between the generated and input images. CLIP-T evaluates text alignment by measuring the similarity between the generated images and the text prompts.}

        \begin{tabular}{lccc}
            \toprule
            Methods & CLIP-T$\uparrow$ & CLIP-I$\uparrow$ & DINO$\uparrow$ \\
            \midrule
            Custom Diffusion & 0.2537& 0.6706& 0.5163\\
            NeTI & 0.2386& \textbf{0.7104}&  0.5623\\
            OFT & 0.2397& 0.7018& 0.5612\\
            AttnDreamBooth & \underline{0.2547}& 0.6918&\underline{0.5641}\\
            \midrule
            Ours & \textbf{0.2568}& \underline{0.7054}& \textbf{0.5842}\\
            \bottomrule
        \end{tabular}
    
    \label{tab:quantitative_evaluation}
\end{table}

\begin{table}[t]
    \centering
    \caption{User study. For each paired comparison, our method is preferred over the baselines.}
    \begin{tabular}{l c c }
      \toprule
      Baselines & Prefer Baseline    & Prefer Ours  \\
      \midrule
      Custom Diffusion    & 14.3\%  & \textbf{85.7\%}  \\
      NeTI                      & 23.7\%  & \textbf{76.3\%}  \\
      OFT                           & 28.4\%  & \textbf{71.6\%}  \\
      AttnDreamBooth                                    & 35.3\%  & \textbf{64.7\%}  \\
      \bottomrule
          \label{tab:user_study}
    \end{tabular}
    \vspace{-0.3cm}
\end{table}

\subsection{Results}
\paragraph{Qualitative Evaluation.}
In Figure~\ref{fig:qualitative_comparsion}, we present a visual comparison of personalized generations for various concepts. We employ a set of complex prompts for evaluation, such as depicting the pets in a human-like posture and dressing (e.g., ``$S_*$ dressed as Spider-Man swings between tall buildings''), complex spatial relationships (e.g., ``$S_*$ inside a box, floating on the water''), and composition of multiple changes (e.g., ``A steampunk $S_*$ with gears and pipes, exploring a retro factory''). As observed, Custom Diffusion fails to generate text-aligned images and sometimes discards the new concept in the generation. NeTI and OFT struggle to accurately adapt the given concept in new scenes. AttnDreamBooth achieves improved personalized generations, but still fails to generate identity-preserved and text-aligned images, especially for prompts requires high visual variability (e.g., ``A cat$_*$ dressed as Spider-Man''). In contrast, the generations by our method accurately preserve the concept identity and align with the complex prompts. Additional qualitative results are provided in the Appendix.

Although our method is primarily designed for personalizing general objects, it also performs well in personalizing human faces. Figure~\ref{fig:face_comparison} shows our personalization results on human faces compared with three specialized face personalization methods, including Cross Initialization~\cite{pang2024cross}, PhotoMaker~\cite{li2023photomaker}, and Face2Diffusion~\cite{shiohara2024face2diffusion}. Our method demonstrates superior identity preservation compared to these baselines.

\paragraph{Quantitative Evaluation.}
We quantitatively evaluate each method using 24 concepts and 20 text prompts, generating 32 samples per prompt for each concept. The results are presented in Table~\ref{tab:quantitative_evaluation}. Note that prompts requiring high visual variability are excluded from quantitative evaluation due to the limitations of quantitative metrics in accurately assessing the quality of generated images for these prompts, for two main reasons. First, such prompts significantly alter the concept's appearance, which makes them unsuitable for measuring identity similarity to the input images. Second, methods that neglect to incorporate the new concept in generations tend to achieve high text alignment scores, as these scores are calculated without considering the new concept. Consequently, using relatively simple prompts, our method achieves slightly higher CLIP-T scores than AttnDreamBooth. In terms of CLIP-I and DINO scores, our method outperforms AttnDreamBooth, likely due to the insufficient text embedding learning in AttnDreamBooth. NeTI achieves the highest CLIP-I score but ranks lowest in text alignment, indicating a tendency to overfit the new concept. Overall, the results demonstrate that our method achieves a superior balance between identity preservation and text alignment compared to the baselines.

\begin{figure}[t]
    \centering
    \renewcommand{\arraystretch}{0.3}
    \setlength{\tabcolsep}{0.5pt}
    {\footnotesize

    \begin{tabular}{c c c c c }

        \begin{tabular}{c} Input \end{tabular} &
        \begin{tabular}{c} w/o CER \end{tabular} &
        \begin{tabular}{c} w/o CAR\end{tabular} &
        \begin{tabular}{c} w/o Rescale\end{tabular} &
        \begin{tabular}{c} Full \end{tabular} \\

        \includegraphics[width=0.093\textwidth]{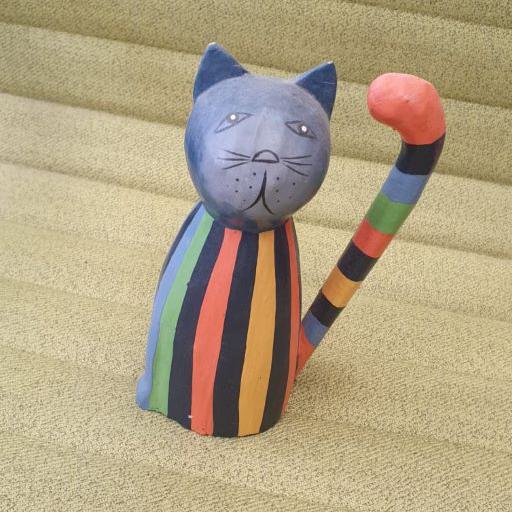} &
        \includegraphics[width=0.093\textwidth]{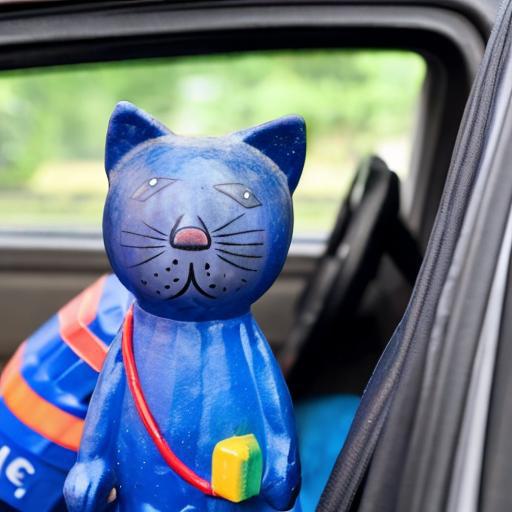} &
        \includegraphics[width=0.093\textwidth]{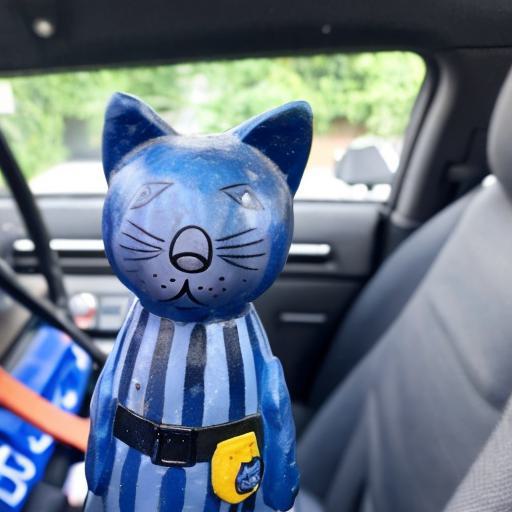} &
        \includegraphics[width=0.093\textwidth]{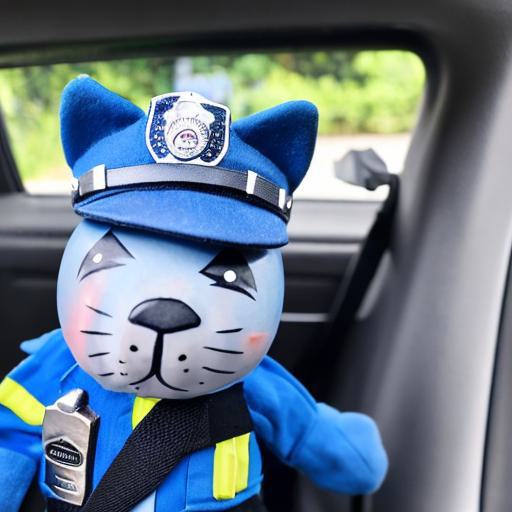}&
        \includegraphics[width=0.093\textwidth]{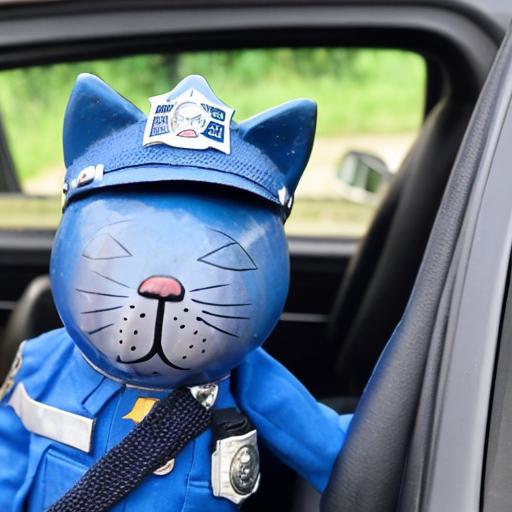} \\
        
        & \multicolumn{4}{c}{A {S*} wearing a police cap in a police car\vspace{0.5mm}} \\

        \includegraphics[width=0.093\textwidth]{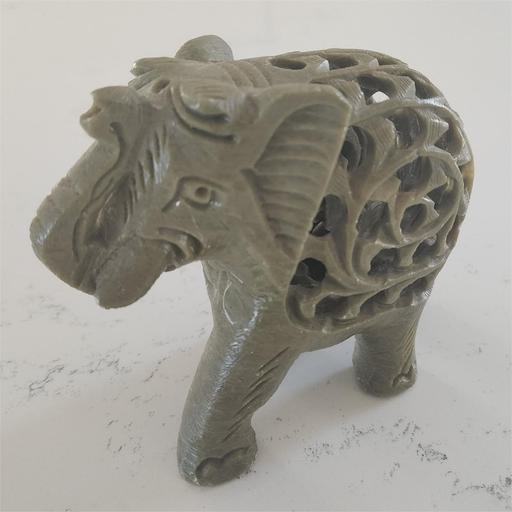} &
        \includegraphics[width=0.093\textwidth]{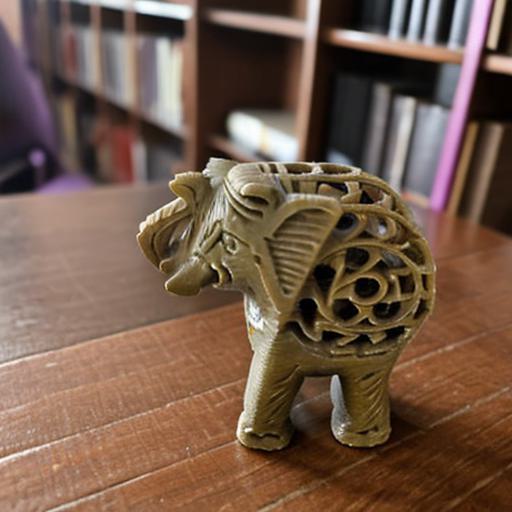} &
        \includegraphics[width=0.093\textwidth]{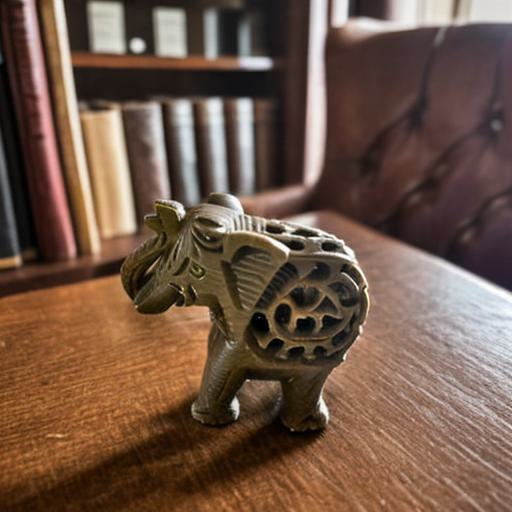} &
        \includegraphics[width=0.093\textwidth]{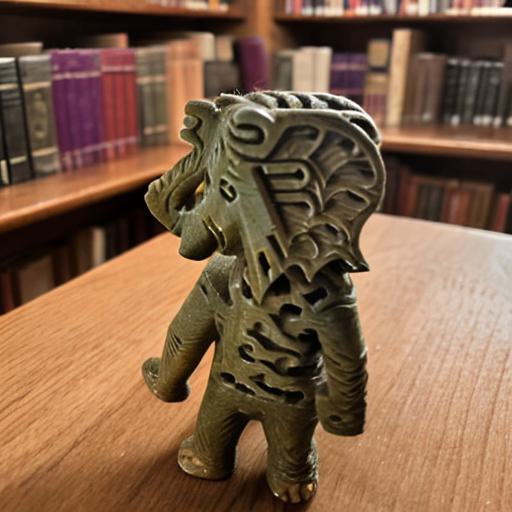}&
        \includegraphics[width=0.093\textwidth]{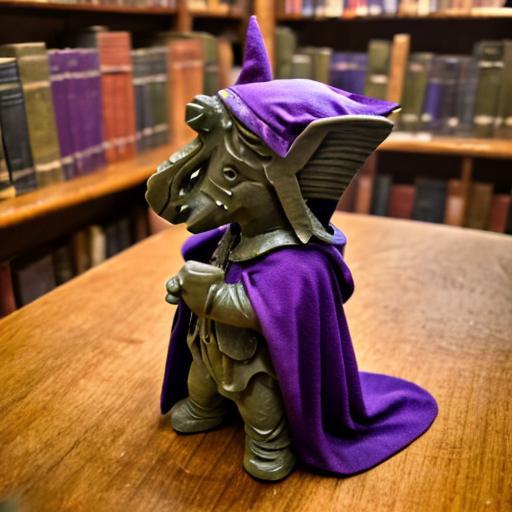} \\
        & \multicolumn{4}{c}{A {S*} dressed as a purple wizard on a desk} \\
        & \multicolumn{4}{c}{in a medieval library\vspace{0.5mm}} \\

        \includegraphics[width=0.093\textwidth]{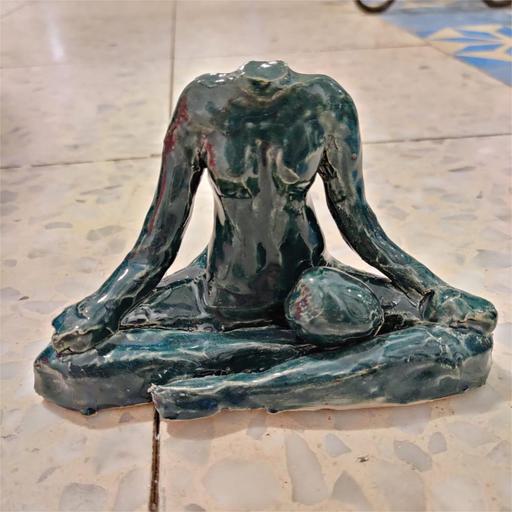} &
        \includegraphics[width=0.093\textwidth]{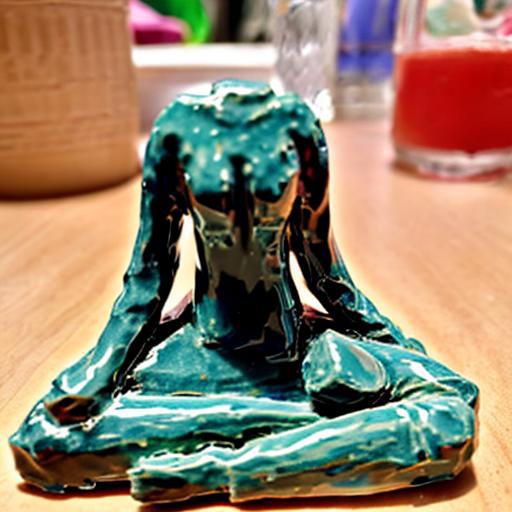} &
        \includegraphics[width=0.093\textwidth]{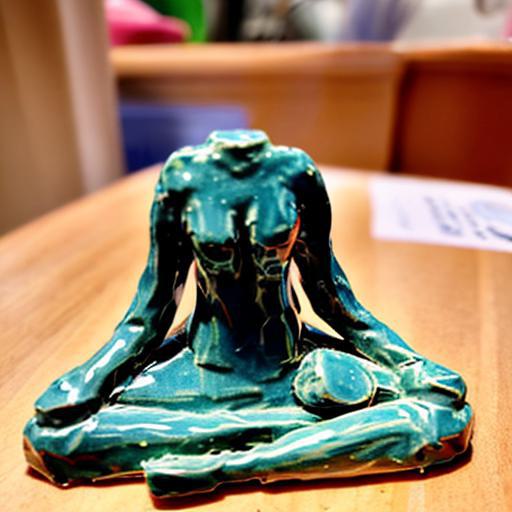} &
        \includegraphics[width=0.093\textwidth]{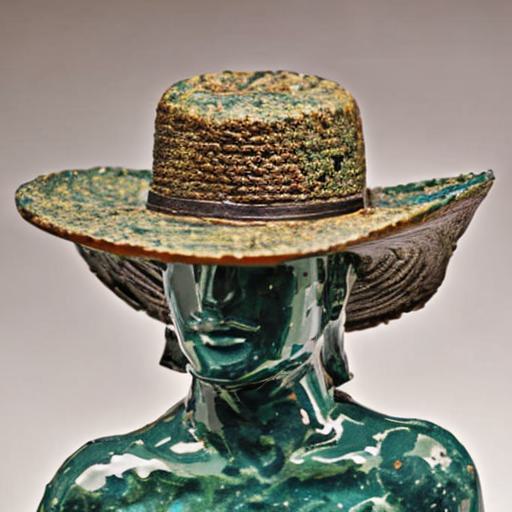}&
        \includegraphics[width=0.093\textwidth]{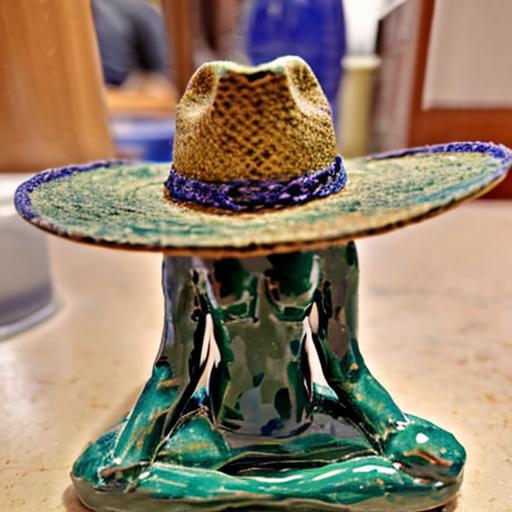} \\
        
        & \multicolumn{4}{c}{A {S*} wearing a sombrero\vspace{0.5mm}} \\

    \end{tabular}
    }
    \caption{Ablation study. We compare models trained without Context Embedding Regularization (w/o CER), without Context Attention Regularization (w/o CAR), and without embedding rescaling strategy (w/o Rescale). All sub-modules are essential for achieving identity-preserved and text-aligned personalized generations.}
    \label{fig:ablation_study}

\end{figure}

\paragraph{User Study.}
We conduct a paired human preference study to compare CoRe with the baseline methods. In each question, we present two generated images, one from our method and one from a baseline, using the same prompt. Participants are asked to evaluate the generated images based on identity preservation and text alignment. We collect 1200 responses from 60 participants. As shown in Table~\ref{tab:user_study}, our method is clearly preferred over the baselines, indicating its superiority in identity preservation and text alignment.

\subsection{Ablation Study}
In this section, we ablate each sub-module of our method to demonstrate its contribution. Figures~\ref{fig:ablation_study} shows the results of the ablation study. As shown, the absence of the contextual embedding regularization leads to degradation in both identity preservation and text alignment. The model without the context attention regularization tends to generate images that are similar to the input, indicating a potential overfit to the concept. Additionally, without applying the embedding rescaling strategy, the model exhibits slight degradation in both text alignment and identity preservation. Additional ablation study results can be found in the Appendix.

\begin{figure}[t]
 \centering
 \includegraphics[width=1.\linewidth]{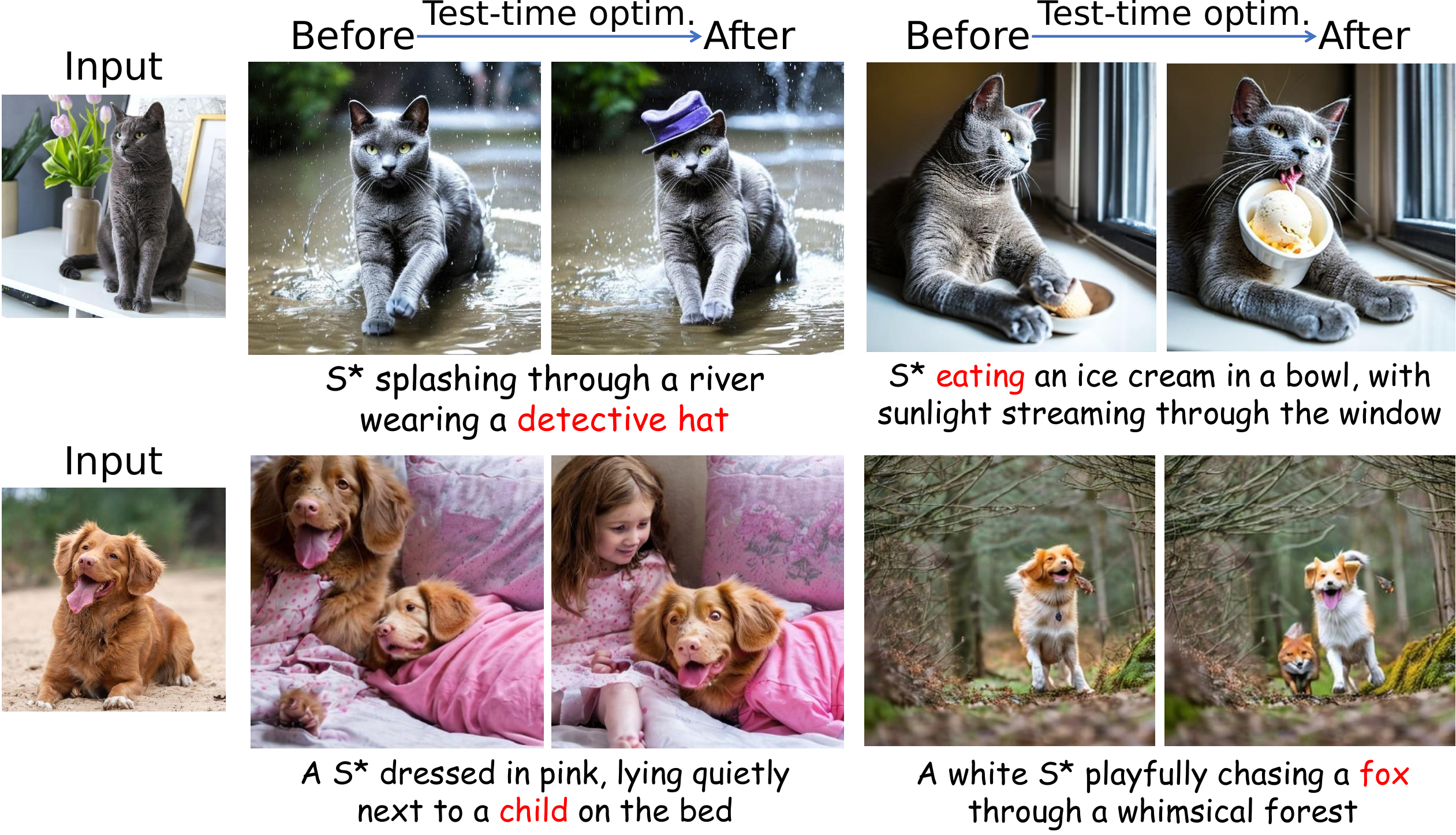}
\caption{Serving as a test-time optimization technique, CoRe enables previously omitted words to be reflected in the generated images.}
\label{fig:test-time}

\end{figure}

\subsection{Test-Time Optimization}
In this section, we evaluate the effectiveness of CoRe for test-time optimization. Given a specific prompt for generation, we perform an additional 10 optimization steps using CoRe to refine the output embeddings and attention maps for this prompt. As illustrated in Figure~\ref{fig:test-time}, this strategy helps to better align the generations with the prompts, allowing previously omitted words to be reflected in the new images. For example, in the second row, test-time optimization effectively replaces the unintended ``dog'' with the correct ``child'', and retrieves the missing ``fox''.

\section{Conclusions and Limitations}
In conclusion, we proposed a personalization method named CoRe that enhances the text embedding learning for the new concept by regularizing context tokens. This method is based on the insight that appropriate output embeddings of context tokens are achievable only when the new concept’s text embedding is correctly learned. Our experimental results demonstrate that CoRe outperforms the baseline methods. As shown in Figure~\ref{fig:test-time}, our method still faces challenges with difficult compositions involving the learned concept and other objects, which is partly inherited from the pretrained model. CoRe can serve as a test-time optimization technique to enhance the generation of such difficult compositions.

\clearpage

\bibliographystyle{aaai25}
\bibliography{ref}
\clearpage
\clearpage

\appendix

\section{A\ \ \ Analysis of the Context Embedding}
In this section, we present the analysis of the context tokens' output embeddings, supplementing Figure~\ref{fig:motivation}. As illustrated, the text embedding learned by CoRe exhibits greater similarity to prompts composed solely of existing tokens, in contrast to the overfitted text embedding produced by Textual Inversion. Additionally, we present the results from CoRe when the regularization prompt set is not employed, demonstrating that this strategy significantly enhances the generalization ability of the learned text embeddings.

\section{B\ \ \ Additional Qualitative Comparisons}
\begin{figure*}[htbp]
    \centering
    \renewcommand{\arraystretch}{0.3} 
    \setlength{\tabcolsep}{0.5pt} 

    {\footnotesize
    \begin{tabular}{c c c c c @{\hspace{0.06cm}} c c}
        \normalsize Input &
        \multicolumn{1}{c}{\normalsize NeTI} &
        \multicolumn{1}{c}{\normalsize OFT} &
        \multicolumn{1}{c}{\normalsize ClassDiffusion} &
        \multicolumn{1}{c}{\normalsize ADB} &
        \multicolumn{2}{c}{\normalsize Ours} \\

         \includegraphics[width=0.126\textwidth, height=0.126\textwidth]{images/input_imgs/cat_toy.jpg} &
        \includegraphics[width=0.126\textwidth]{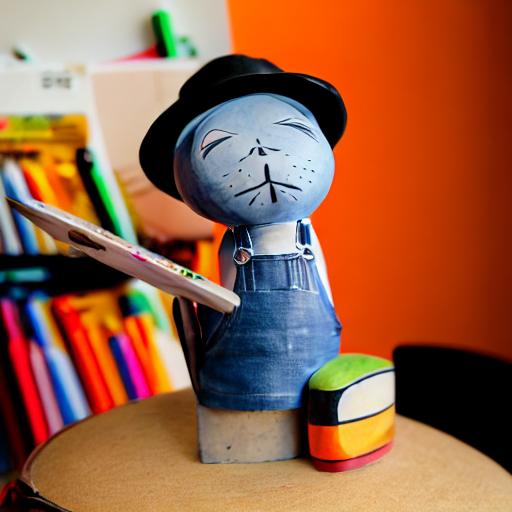} &
        \includegraphics[width=0.126\textwidth]
        {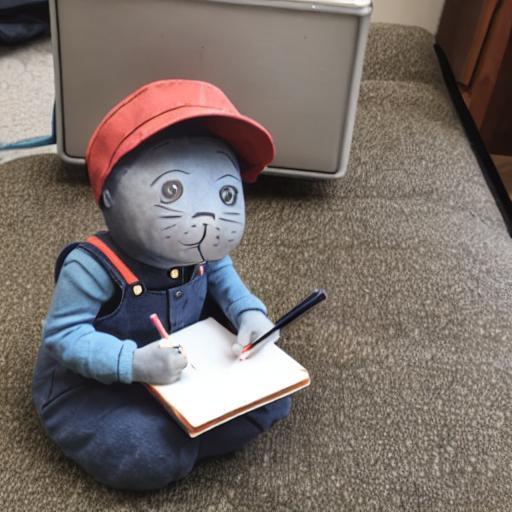} &
        \includegraphics[width=0.126\textwidth]
        {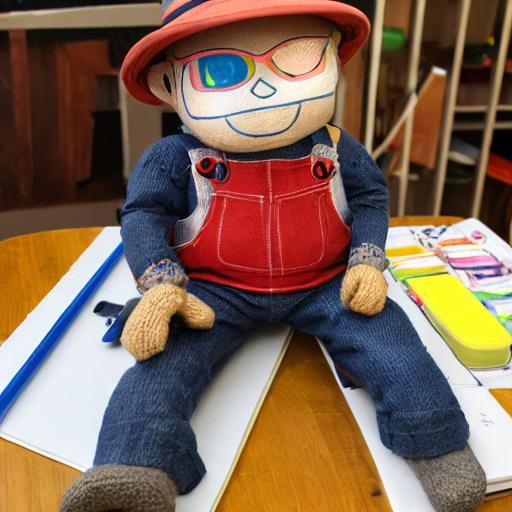} &
        \includegraphics[width=0.126\textwidth]
        {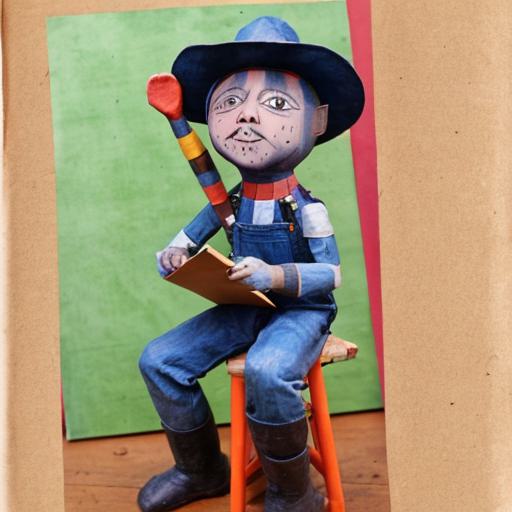} &
        \includegraphics[width=0.126\textwidth]{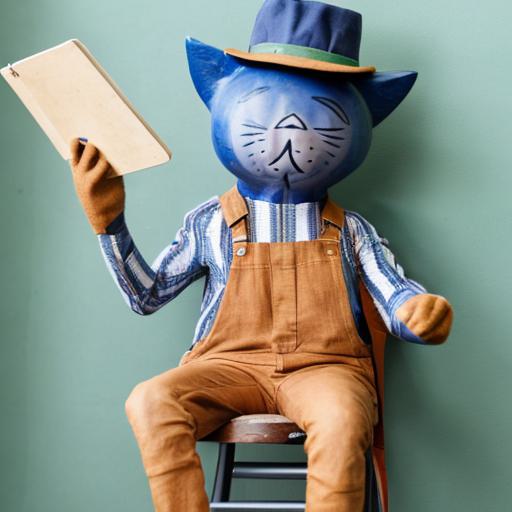} &
        \includegraphics[width=0.126\textwidth]{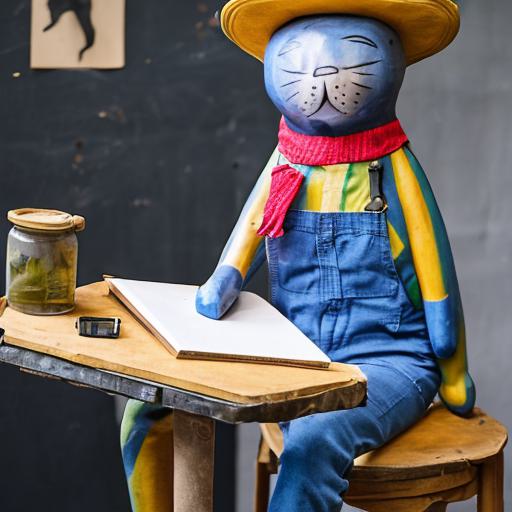} \\

        \raisebox{0.06\textwidth}{\begin{tabular}{c} A S* dressed in\\overalls and a\\wide-brimmed hat,\\holding a sketchpad\\and sitting on\\a stool.\end{tabular}} &
        \includegraphics[width=0.126\textwidth]{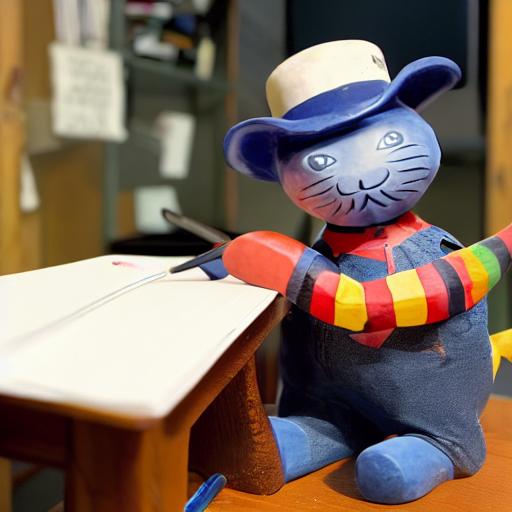} &
        \includegraphics[width=0.126\textwidth]
        {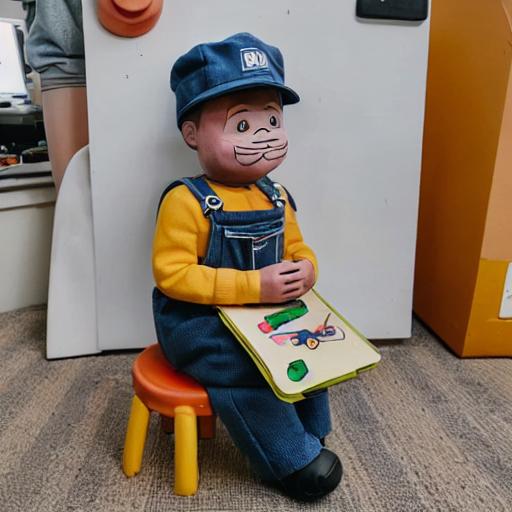} &
        \includegraphics[width=0.126\textwidth]
        {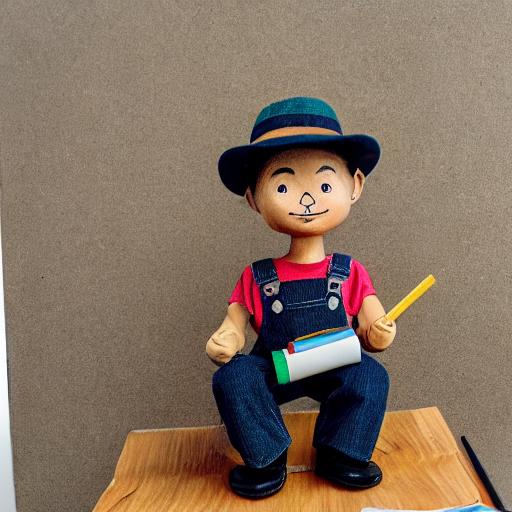} &
        \includegraphics[width=0.126\textwidth]
        {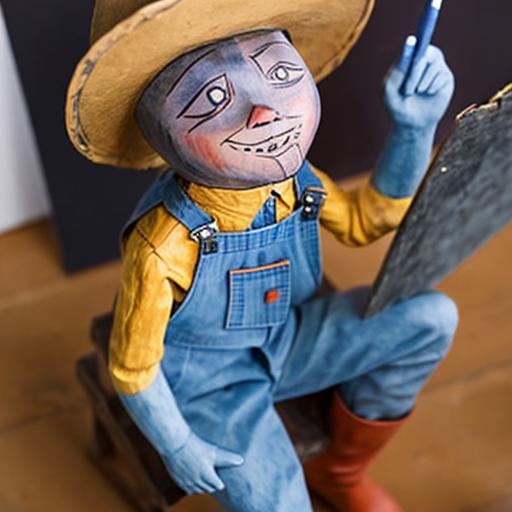} &
        \includegraphics[width=0.126\textwidth]{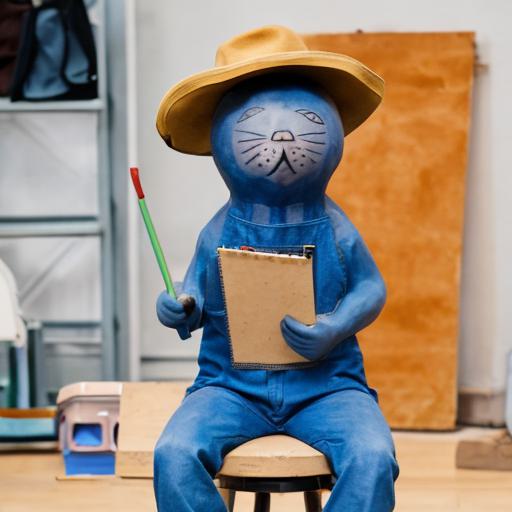} &
        \includegraphics[width=0.126\textwidth]{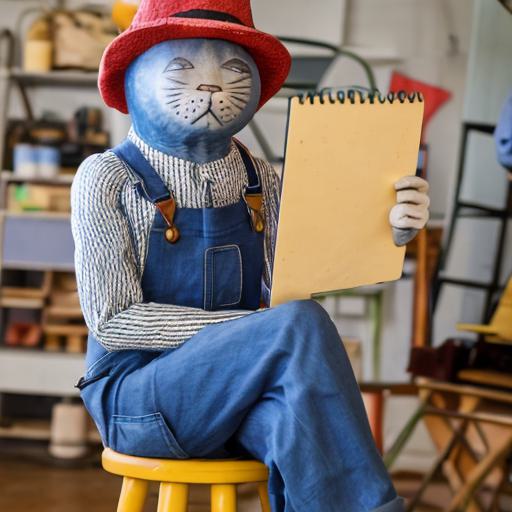} \\

        \includegraphics[width=0.126\textwidth, height=0.126\textwidth]{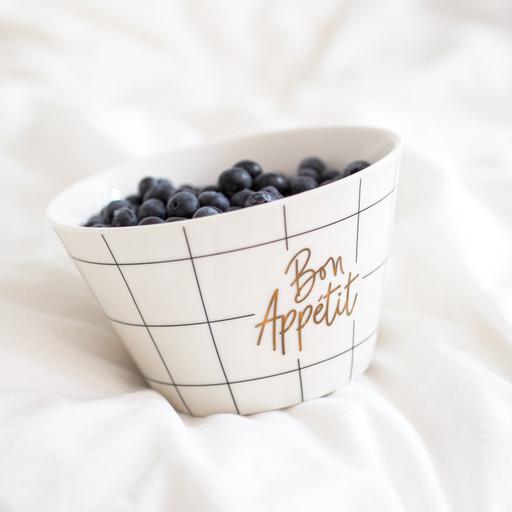} &
        \includegraphics[width=0.126\textwidth]{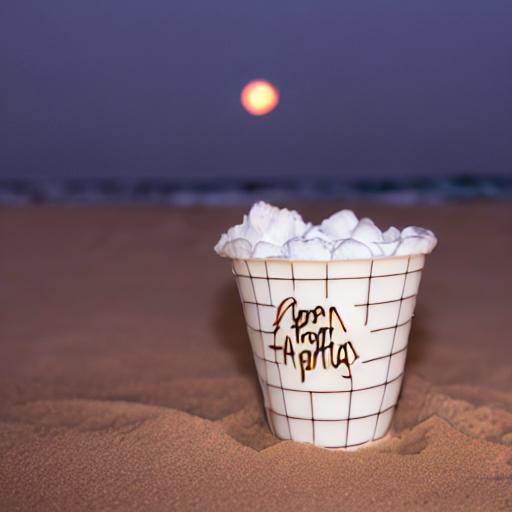} &
        \includegraphics[width=0.126\textwidth]{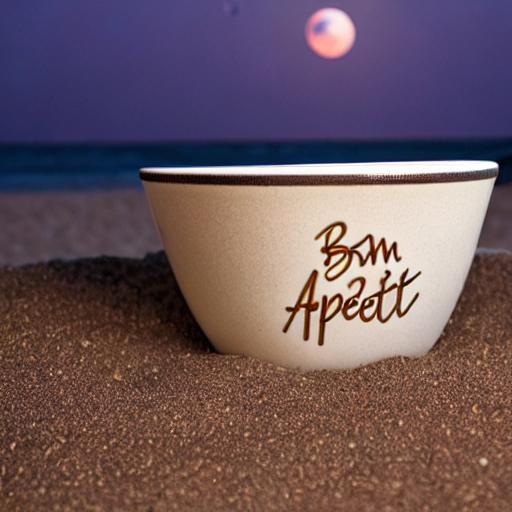} &
        \includegraphics[width=0.126\textwidth]{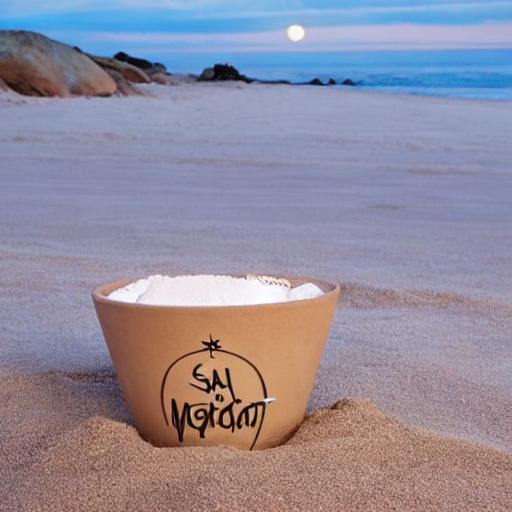} &
        \includegraphics[width=0.126\textwidth]
        {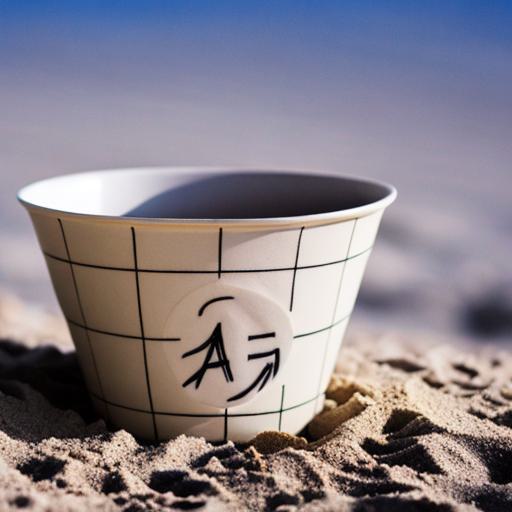} &
        \includegraphics[width=0.126\textwidth]{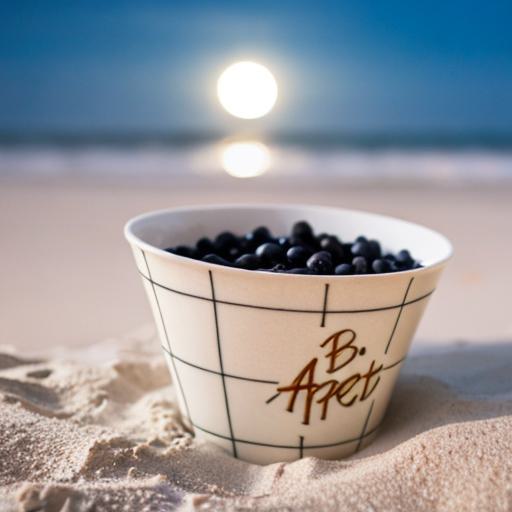} &
        \includegraphics[width=0.126\textwidth]{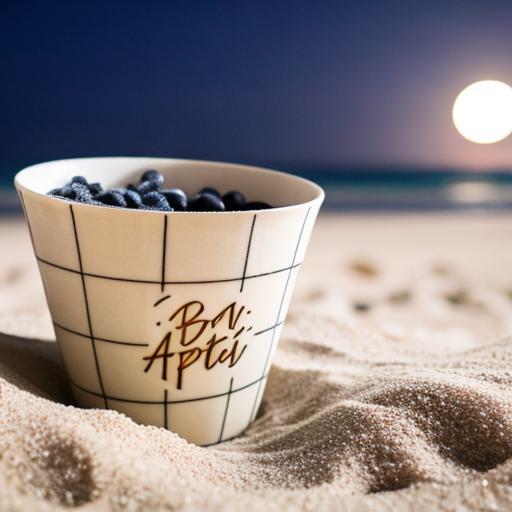} \\

        \raisebox{0.06\textwidth}{\begin{tabular}{c} A {S*} covered \\ by sand under \\ the moonlight on \\a serene beach\end{tabular}} &
        \includegraphics[width=0.126\textwidth]{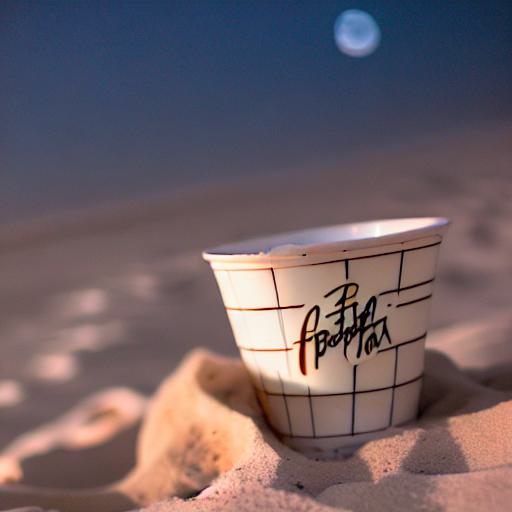} &
        \includegraphics[width=0.126\textwidth]{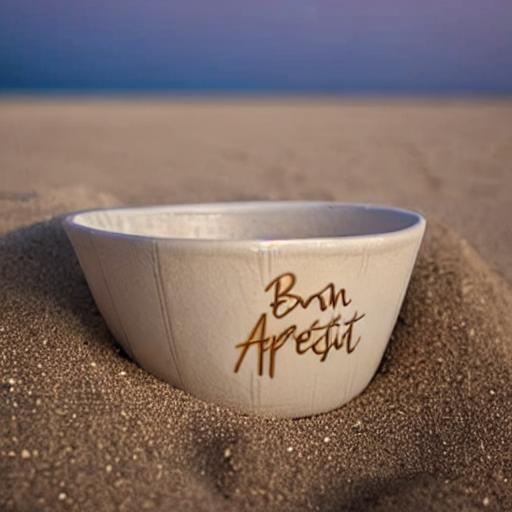} &
        \includegraphics[width=0.126\textwidth]{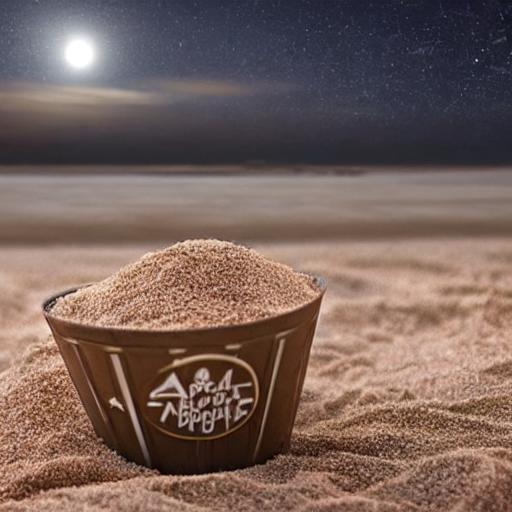} &
        \includegraphics[width=0.126\textwidth]
        {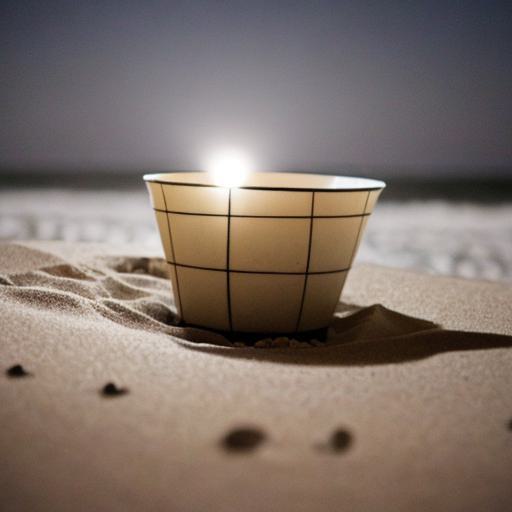} &
        \includegraphics[width=0.126\textwidth]{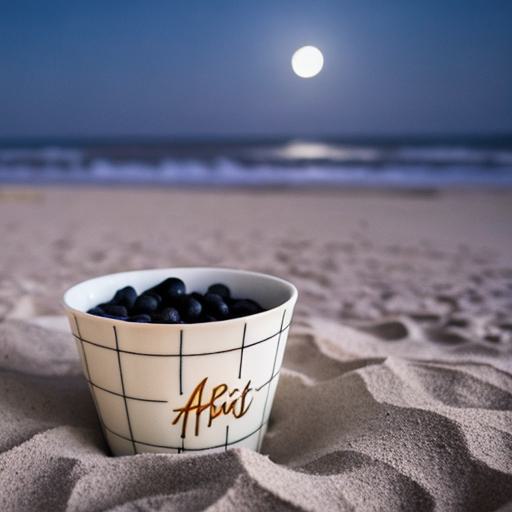} &
        \includegraphics[width=0.126\textwidth]{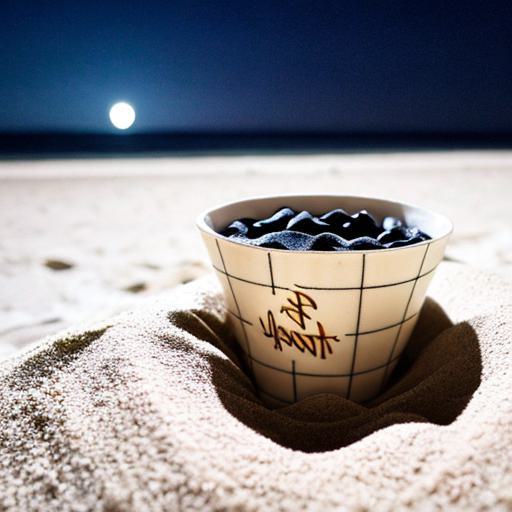} \\

        \includegraphics[width=0.126\textwidth, height=0.126\textwidth]{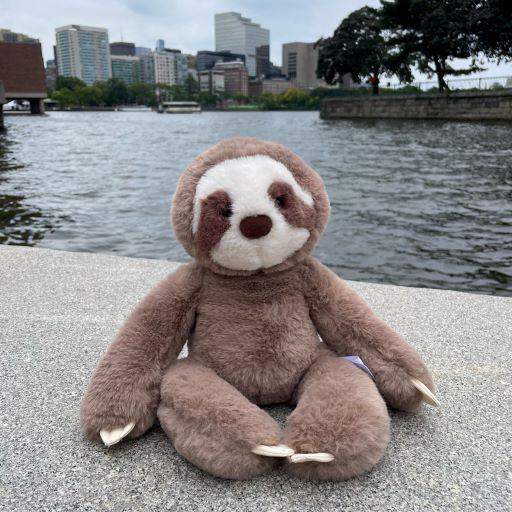} &
        \includegraphics[width=0.126\textwidth]{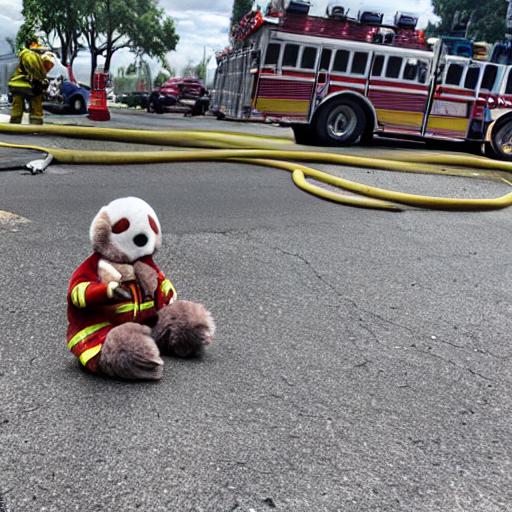} &
        \includegraphics[width=0.126\textwidth]{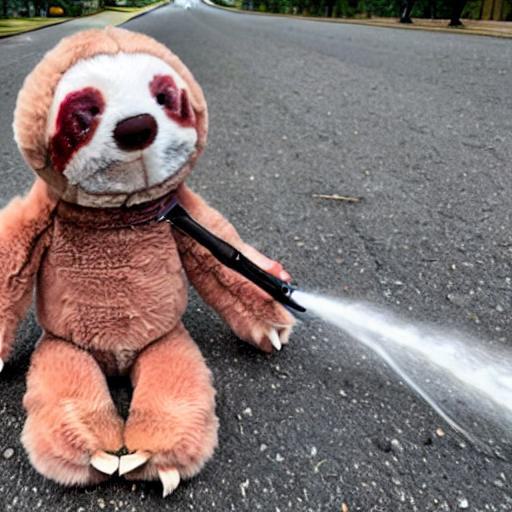} &
        \includegraphics[width=0.126\textwidth]{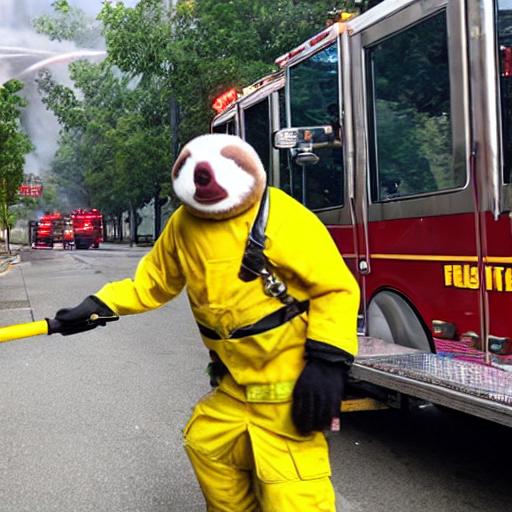} &
        \includegraphics[width=0.126\textwidth]
        {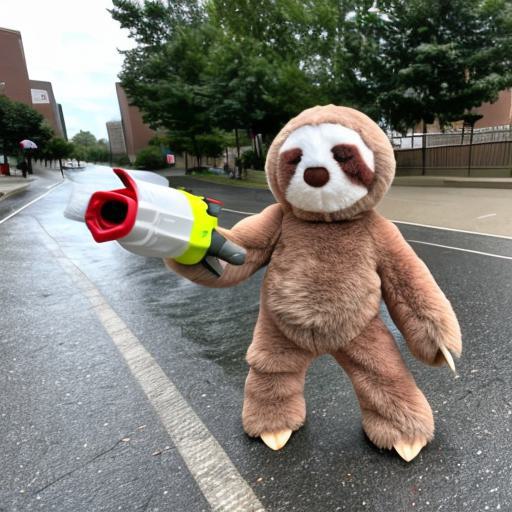} &
        \includegraphics[width=0.126\textwidth]{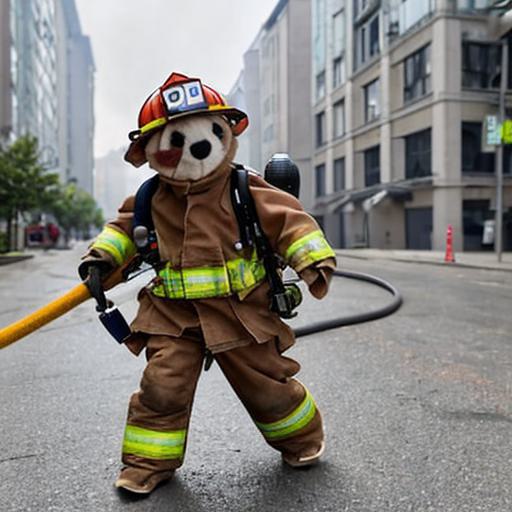} &
        \includegraphics[width=0.126\textwidth]{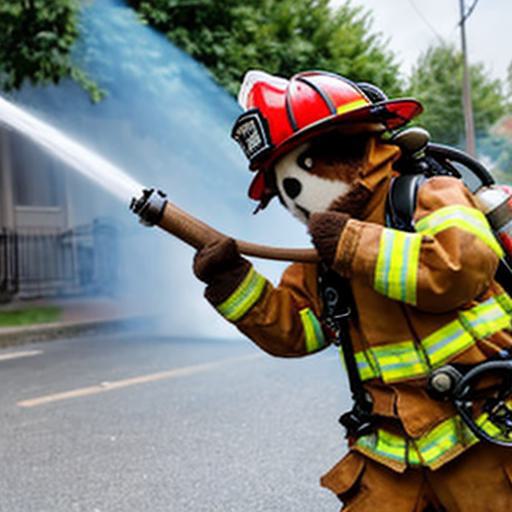} \\

        \raisebox{0.06\textwidth}{\begin{tabular}{c}A {S*} in \\ a firefighter outfit \\fights a fire\\ on the street \\with a water gun\end{tabular}} &
        \includegraphics[width=0.126\textwidth]{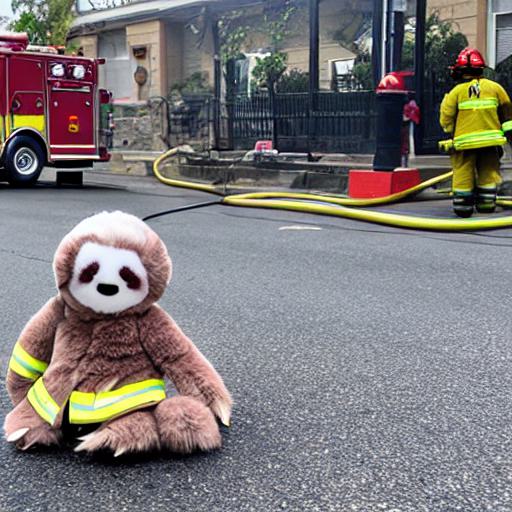} &
        \includegraphics[width=0.126\textwidth]{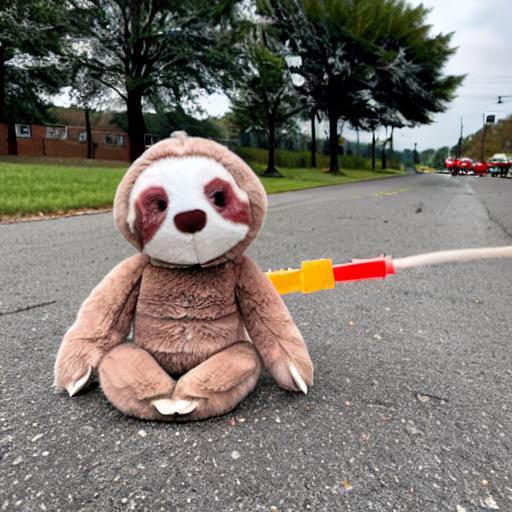} &
        \includegraphics[width=0.126\textwidth]{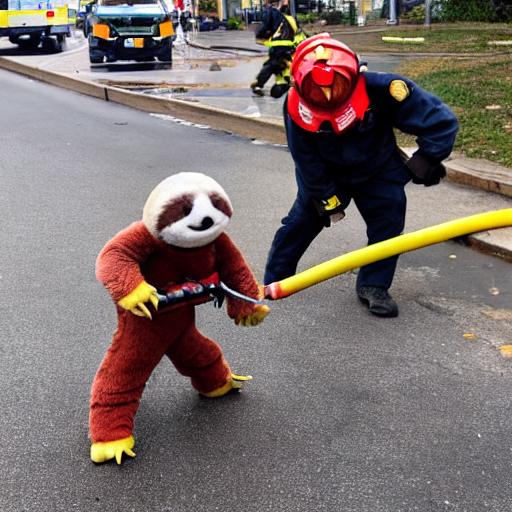} &
        \includegraphics[width=0.126\textwidth]
        {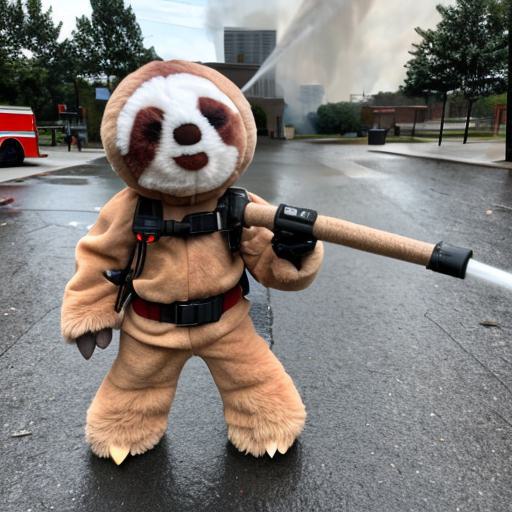} &
        \includegraphics[width=0.126\textwidth]{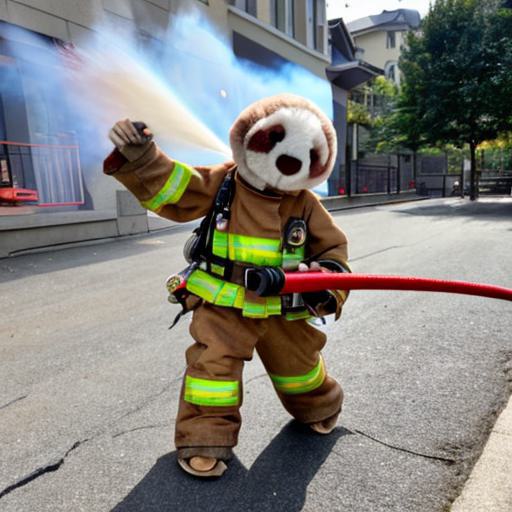} &
        \includegraphics[width=0.126\textwidth]{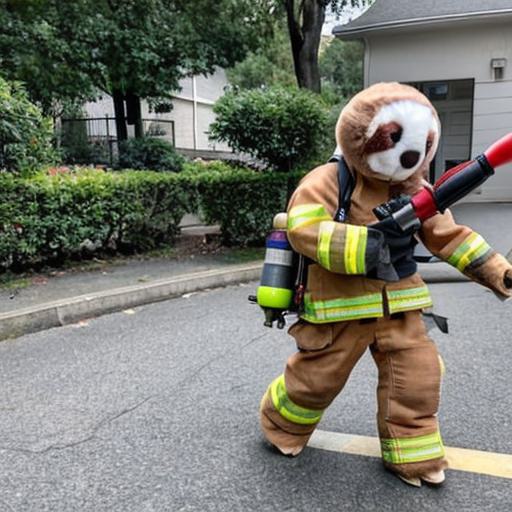} \\

        \includegraphics[width=0.126\textwidth, height=0.126\textwidth]{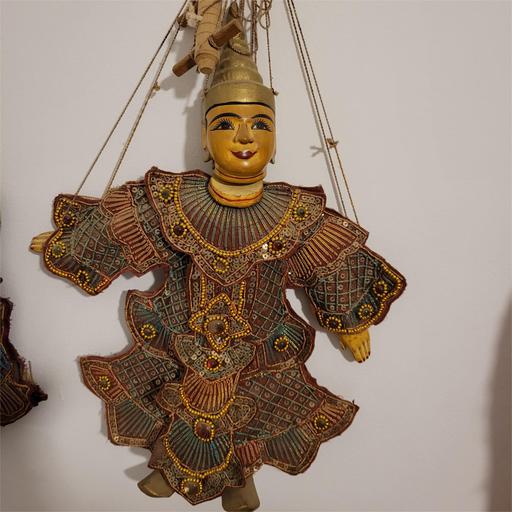} &
        \includegraphics[width=0.126\textwidth]{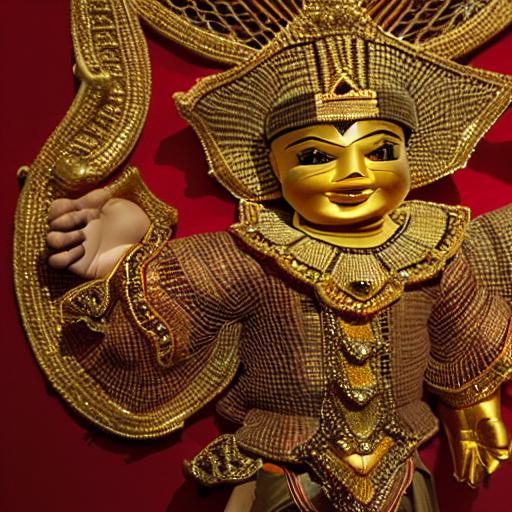} &
        \includegraphics[width=0.126\textwidth]{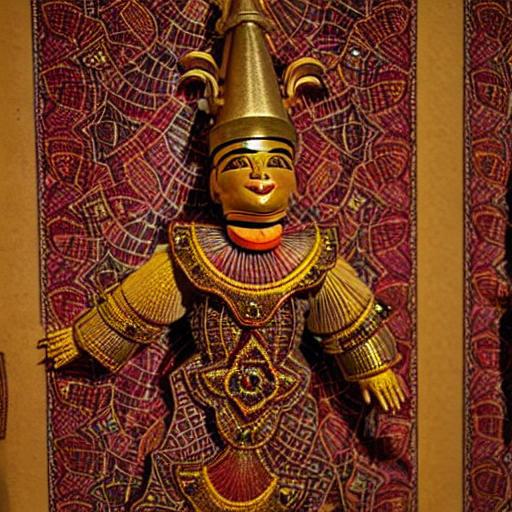} &
        \includegraphics[width=0.126\textwidth]{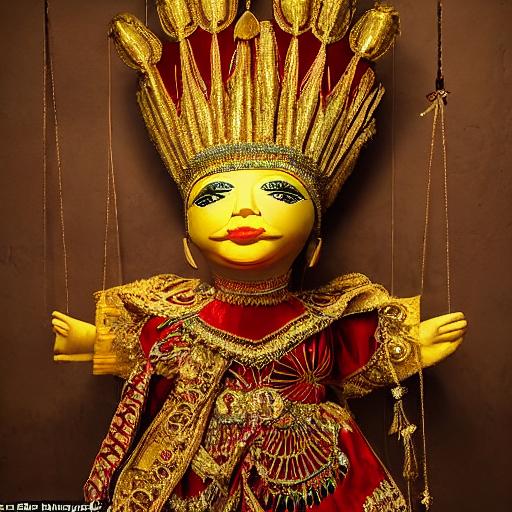} &
        \includegraphics[width=0.126\textwidth]
        {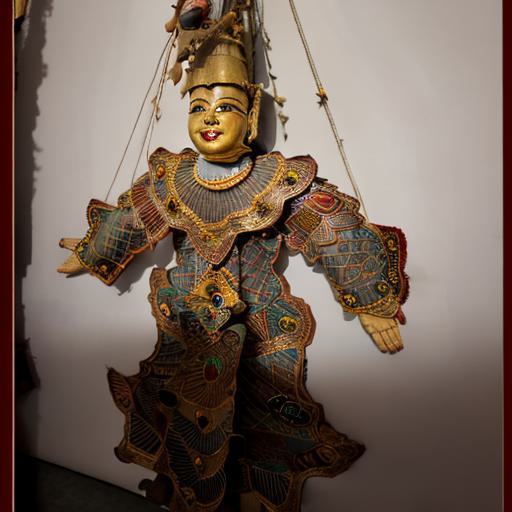} &
        \includegraphics[width=0.126\textwidth]{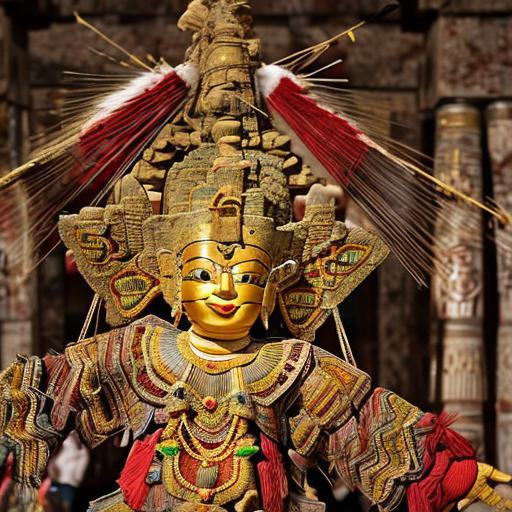} &
        \includegraphics[width=0.126\textwidth]{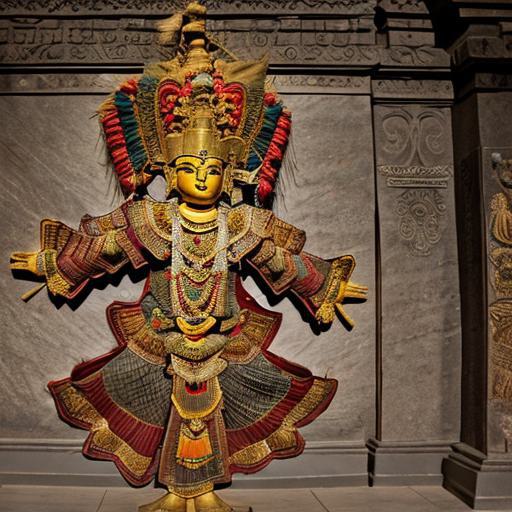} \\

        \raisebox{0.06\textwidth}{\begin{tabular}{c} A grand {S*} adorned\\with golden threads\\and a vividly red\\elaborate headdress,\\set against a\\backdrop of ancient\\stone carvings\end{tabular}} &
        \includegraphics[width=0.126\textwidth]{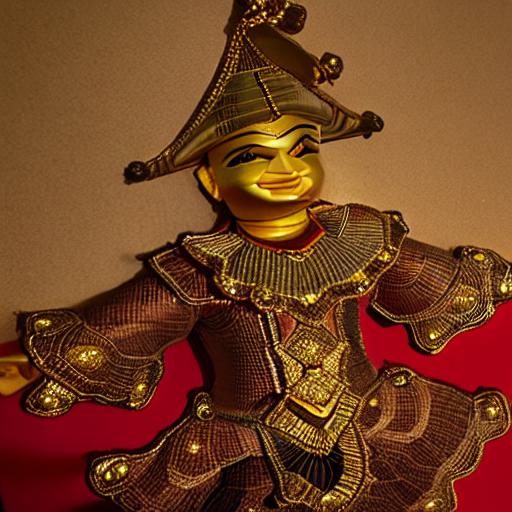} &
        \includegraphics[width=0.126\textwidth]{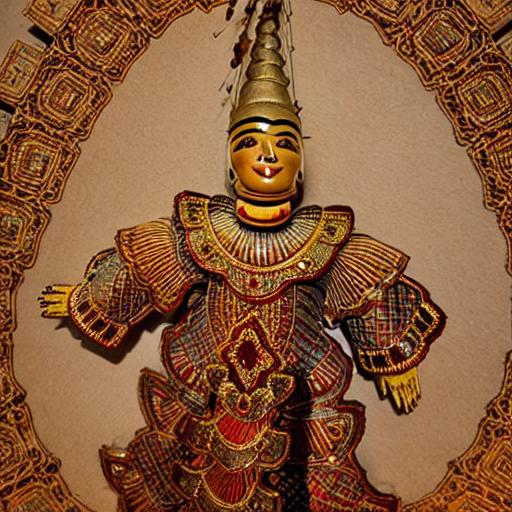} &
        \includegraphics[width=0.126\textwidth]{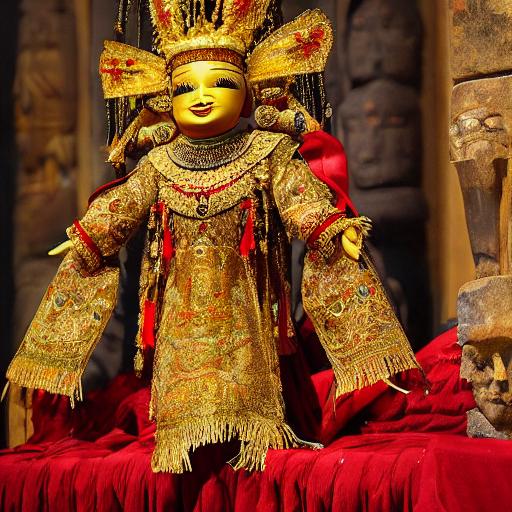} &
        \includegraphics[width=0.126\textwidth]
        {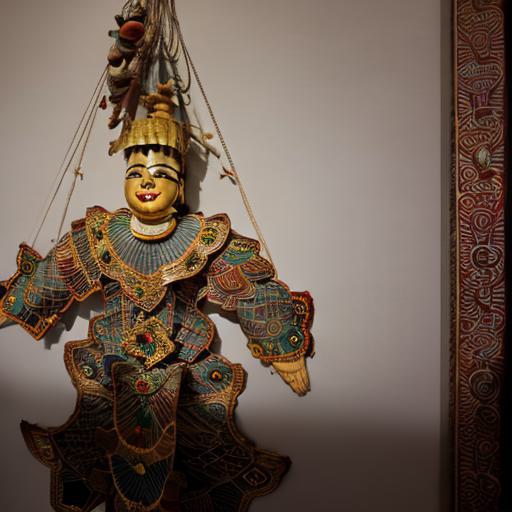} &
        \includegraphics[width=0.126\textwidth]{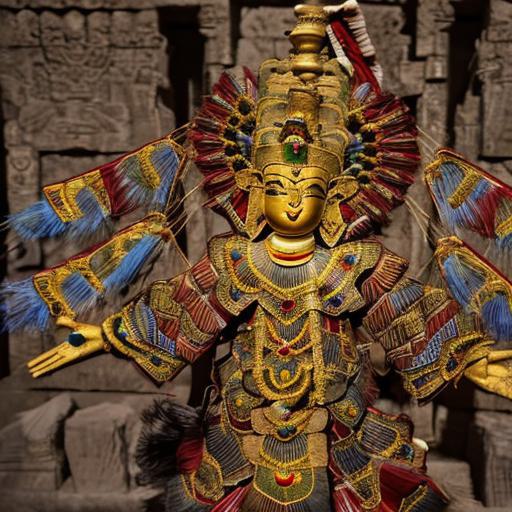} &
        \includegraphics[width=0.126\textwidth]{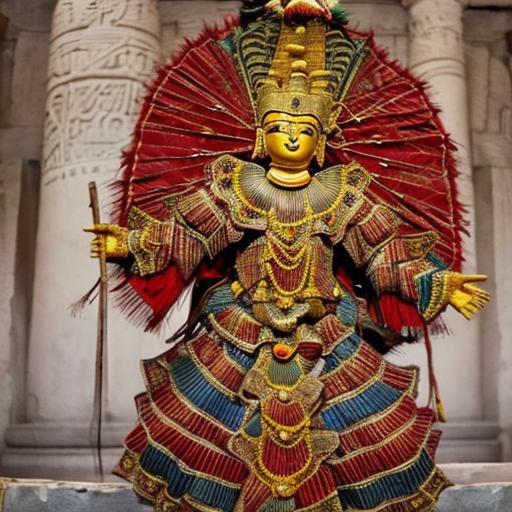} \\

        \includegraphics[width=0.126\textwidth, height=0.126\textwidth]{images/input_imgs/headless_statue.jpg} &
        \includegraphics[width=0.126\textwidth]{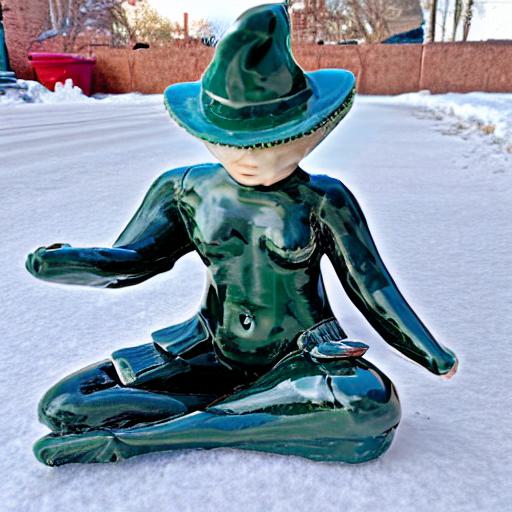} &
        \includegraphics[width=0.126\textwidth]{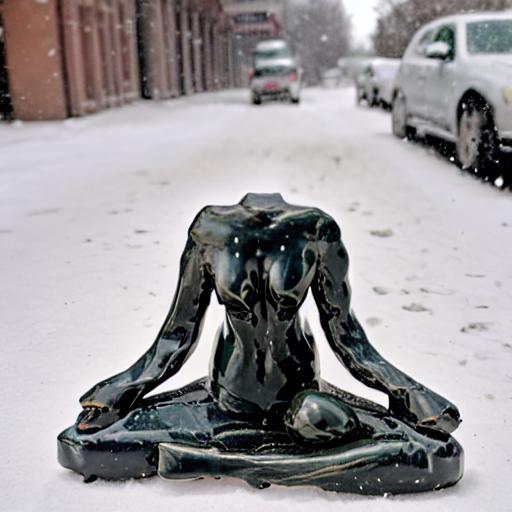} &
        \includegraphics[width=0.126\textwidth]{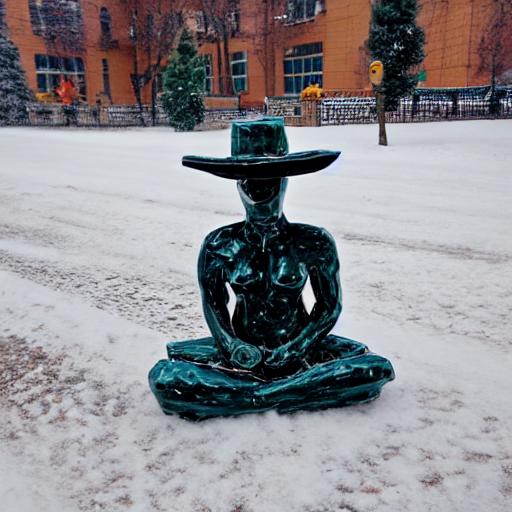} &
        \includegraphics[width=0.126\textwidth]
        {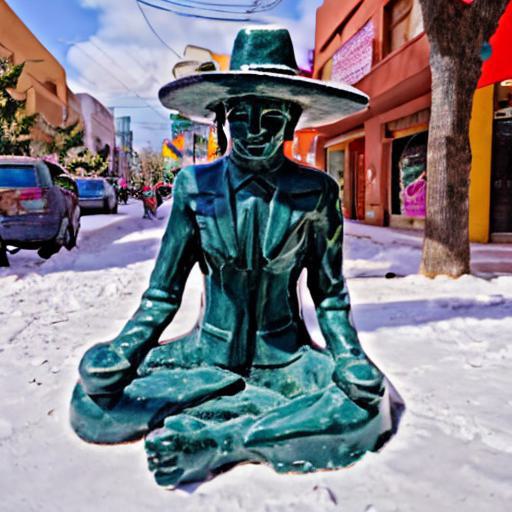} &
        \includegraphics[width=0.126\textwidth]{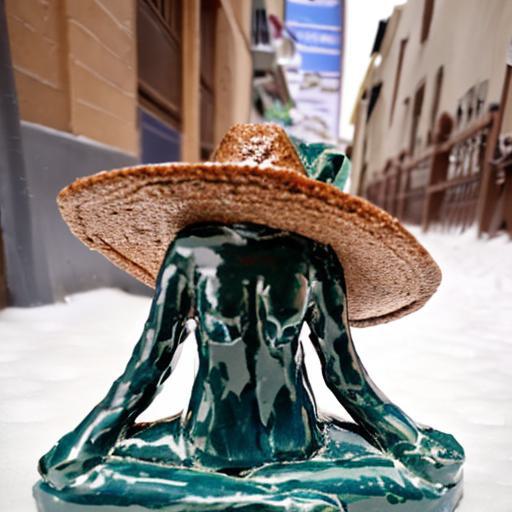} &
        \includegraphics[width=0.126\textwidth]{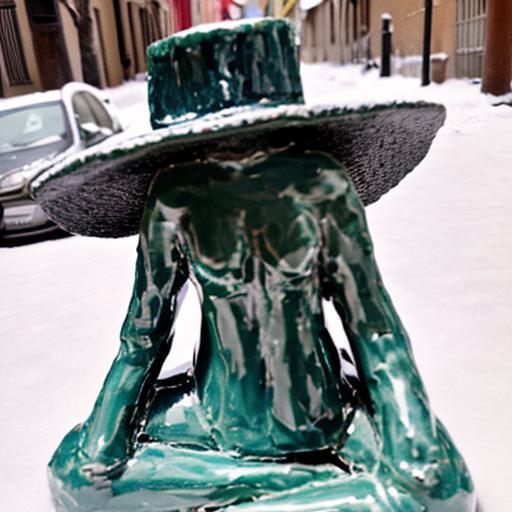} \\

        \raisebox{0.06\textwidth}{\begin{tabular}{c} a {S*} wearing \\a sombrero, \\sitting on the \\snowy street\end{tabular}} &
        \includegraphics[width=0.126\textwidth]{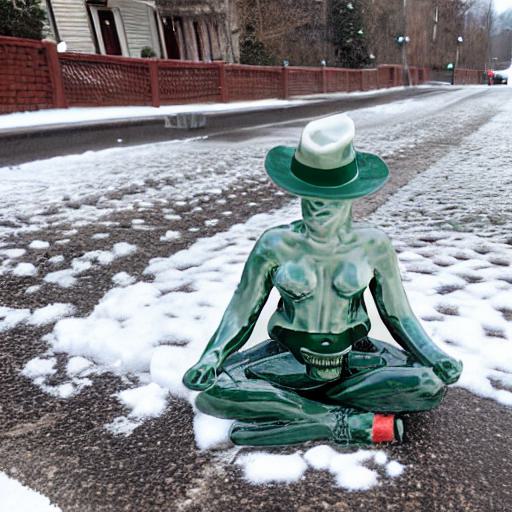} &
        \includegraphics[width=0.126\textwidth]{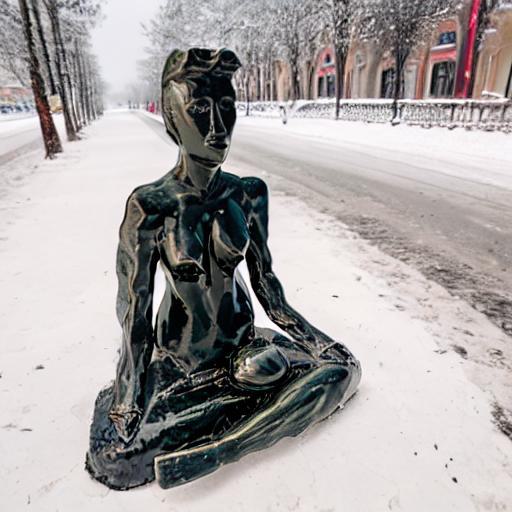} &
        \includegraphics[width=0.126\textwidth]{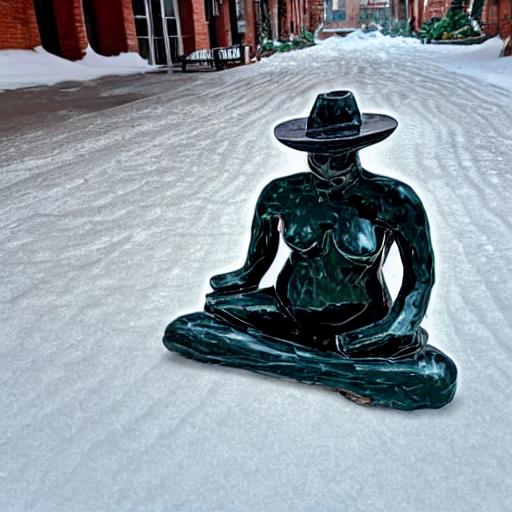} &
        \includegraphics[width=0.126\textwidth]
        {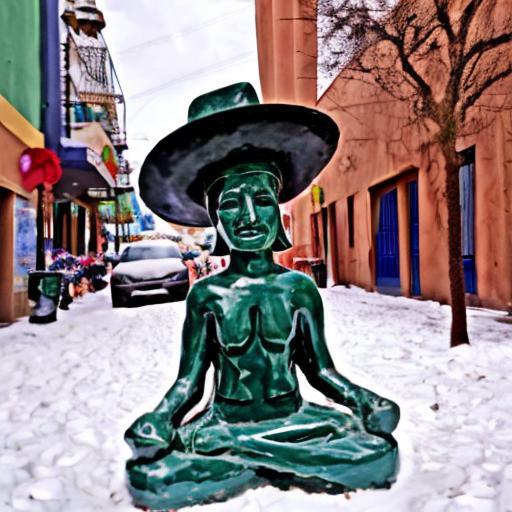} &
        \includegraphics[width=0.126\textwidth]{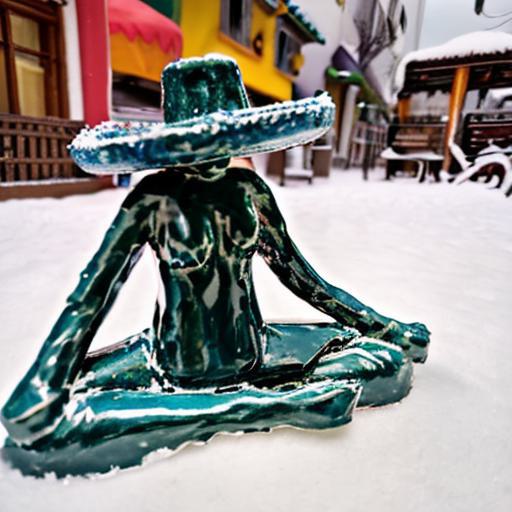} &
        \includegraphics[width=0.126\textwidth]{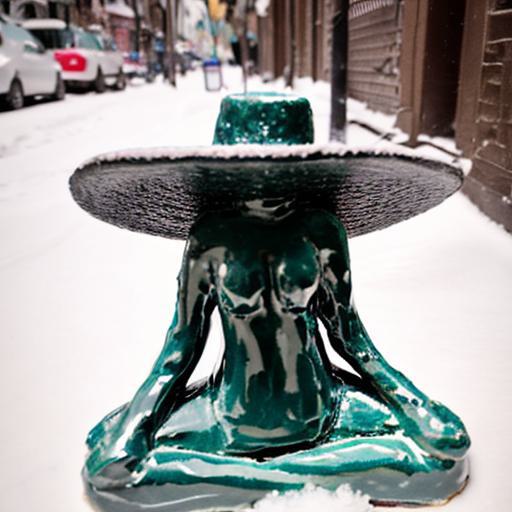} \\
    \end{tabular}
    }

    \caption{Additional qualitative comparisons with four baseline methods, including NeTI~\cite{alaluf2023neural}, OFT~\cite{qiu_oft}, ClassDiffusion~\cite{huang2024classdiffusion}, and AttnDreamBooth (ADB)~\cite{pang2024attndreambooth}.}
    \label{fig:appendix_qualitative_comparison}
\end{figure*}

In Figure~\ref{fig:appendix_qualitative_comparison}, we present additional qualitative comparisons with four baseline methods, including NeTI~\cite{alaluf2023neural}, OFT~\cite{qiu_oft}, AttnDreamBooth~\cite{pang2024attndreambooth}, and the concurrent work, ClassDiffusion~\cite{huang2024classdiffusion}.

\section{C\ \ \ Additional Qualitative Results}
\begin{figure*}[t]
    \centering
    \setlength{\tabcolsep}{0.1pt}
    {\footnotesize
    \begin{tabular}{c@{\hspace{0.3cm}} c@{\hspace{0.3cm}} c@{\hspace{0.3cm}} c@{\hspace{0.25cm}} c@{\hspace{0.18cm}} c}

        \includegraphics[width=0.15\textwidth,height=0.15\textwidth]{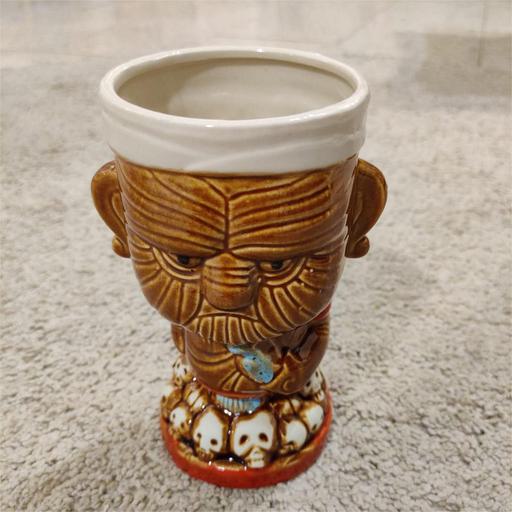} &
        \includegraphics[width=0.15\textwidth]{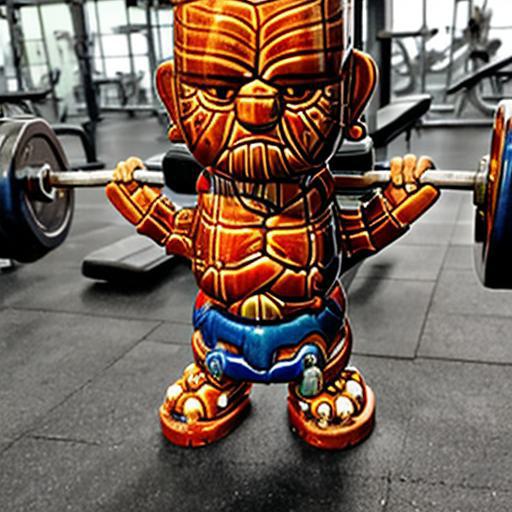} &
        \includegraphics[width=0.15\textwidth]{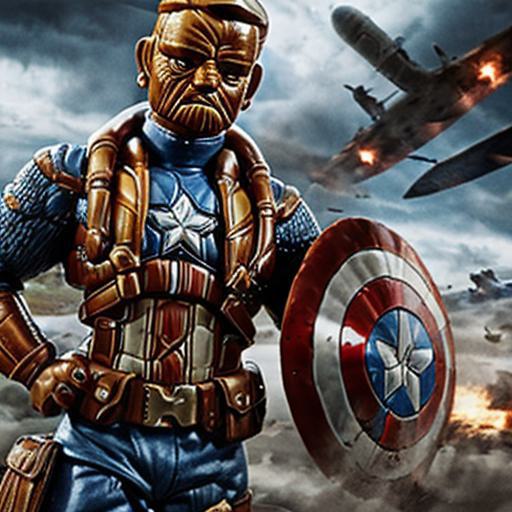} &
        \includegraphics[width=0.15\textwidth]{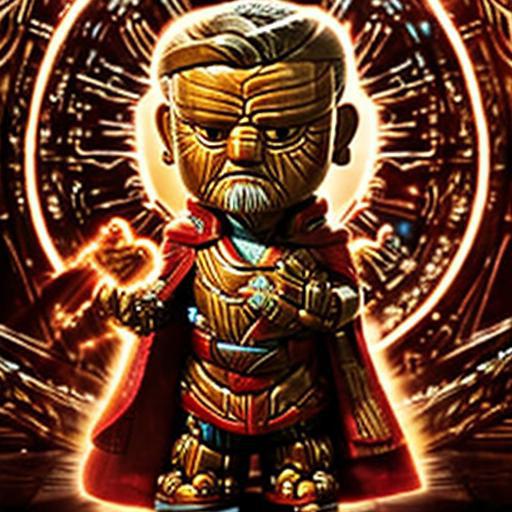} &
        \includegraphics[width=0.15\textwidth]{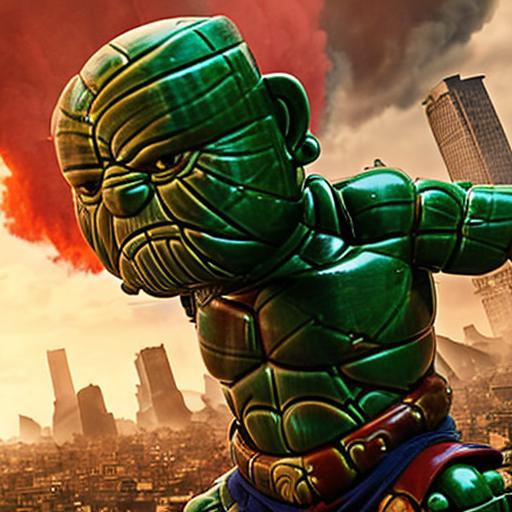} &
        \includegraphics[width=0.15\textwidth]{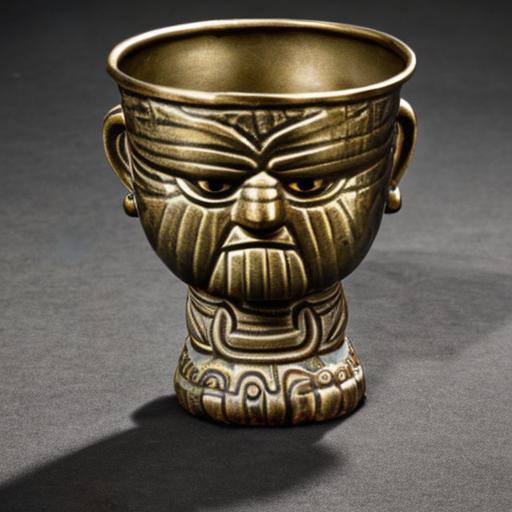} \\

        \begin{minipage}[t]{0.15\textwidth}\centering \vspace{0pt} Input Sample\end{minipage} &
        \begin{minipage}[t]{0.15\textwidth}\centering \vspace{-3mm} A S* is lifting a dumbbell at the gym\end{minipage} &
        \begin{minipage}[t]{0.15\textwidth}\centering \vspace{-3mm} A S* as Captain America in a WWII battlefield\end{minipage} &
        \begin{minipage}[t]{0.15\textwidth}\centering \vspace{-3mm} A S* as Doctor Strange in a magical sanctum\end{minipage} &
        \begin{minipage}[t]{0.15\textwidth}\centering \vspace{-3mm} A S* as Hulk, destroying a city, with Smoke in the background\end{minipage}\vspace{1mm} &
        \begin{minipage}[t]{0.15\textwidth}\centering \vspace{-3mm} A Bronze cup with two ears, in shape of S*\end{minipage} \\

        \includegraphics[width=0.15\textwidth,height=0.15\textwidth]{images/input_imgs/grey_sloth.jpg} &
        \includegraphics[width=0.15\textwidth]{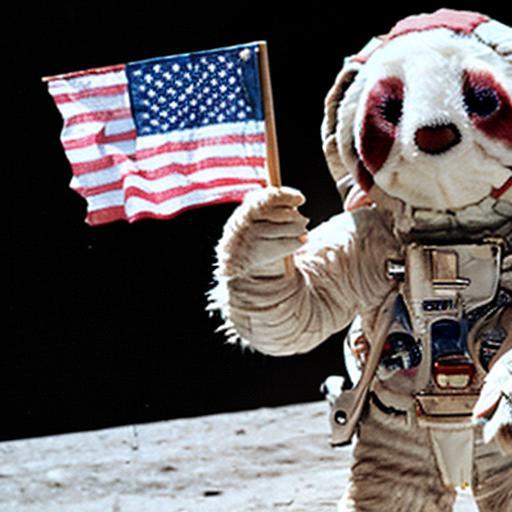} &
        \includegraphics[width=0.15\textwidth]{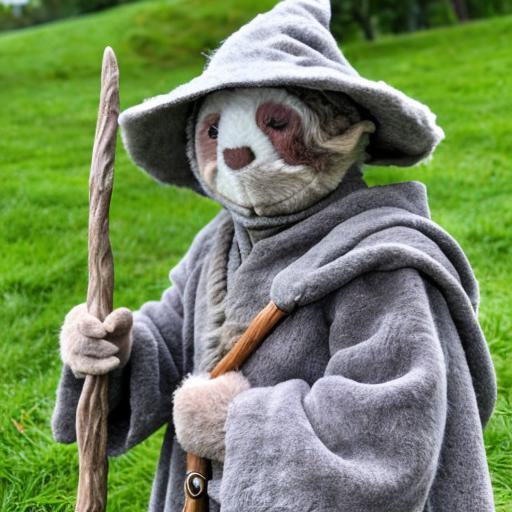} &
        \includegraphics[width=0.15\textwidth]{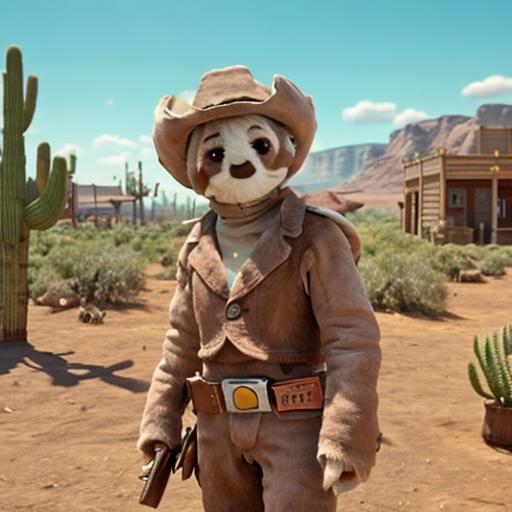} &
        \includegraphics[width=0.15\textwidth]{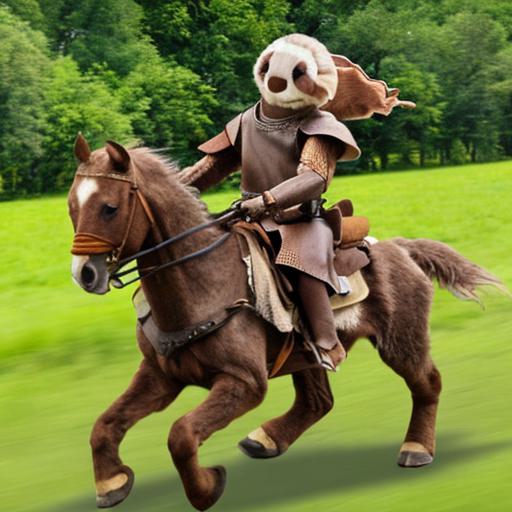} &
        \includegraphics[width=0.15\textwidth]{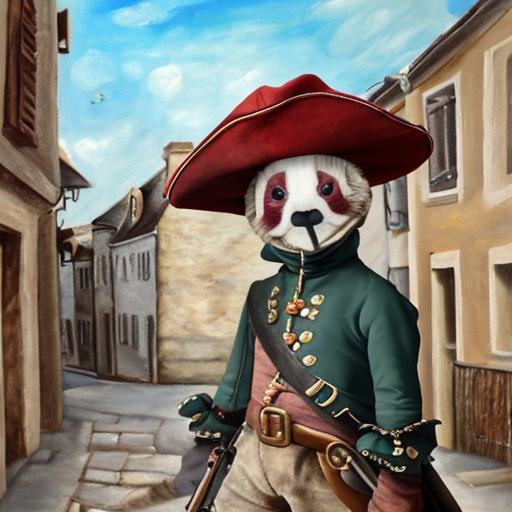} \\

        \begin{minipage}[t]{0.15\textwidth}\centering \vspace{0pt} Input Sample\end{minipage} &
        \begin{minipage}[t]{0.15\textwidth}\centering \vspace{-3mm} A S* as an astronaut holding an American flag on the moon\end{minipage} &
        \begin{minipage}[t]{0.15\textwidth}\centering \vspace{-3mm} S* is dressed as Gandalf the Grey from The Hobbit, wearing a long grey robe, a pointed hat, and holding a wooden staff\end{minipage} &
        \begin{minipage}[t]{0.15\textwidth}\centering \vspace{-3mm} S* as a cowboy in a dusty town showdown, surrounded by cacti and a saloon\end{minipage}\vspace{1mm} &
        \begin{minipage}[t]{0.15\textwidth}\centering \vspace{-3mm} A S* as a knight draped in armor, riding a brown horse and galloping through the lush fields\end{minipage} &
        \begin{minipage}[t]{0.15\textwidth}\centering \vspace{-3mm} An oil painting of S* dressed as a musketeer in an old French town\end{minipage} \\
        \includegraphics[width=0.15\textwidth,height=0.15\textwidth]{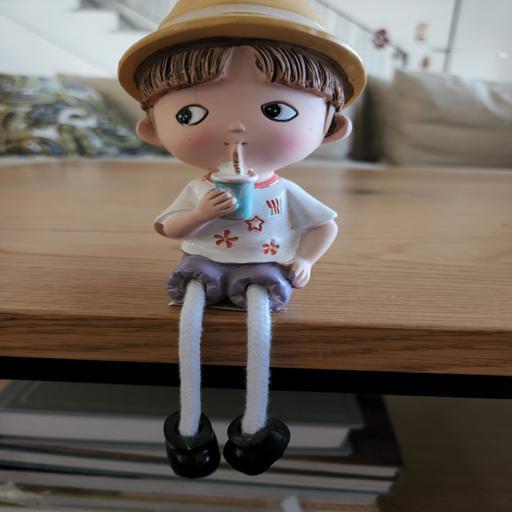} &
        \includegraphics[width=0.15\textwidth]{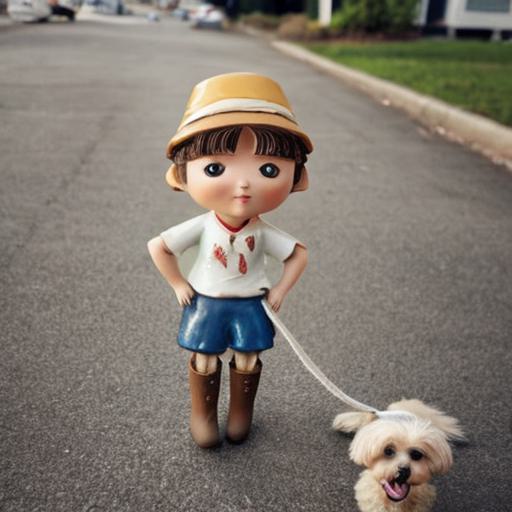} &
        \includegraphics[width=0.15\textwidth]{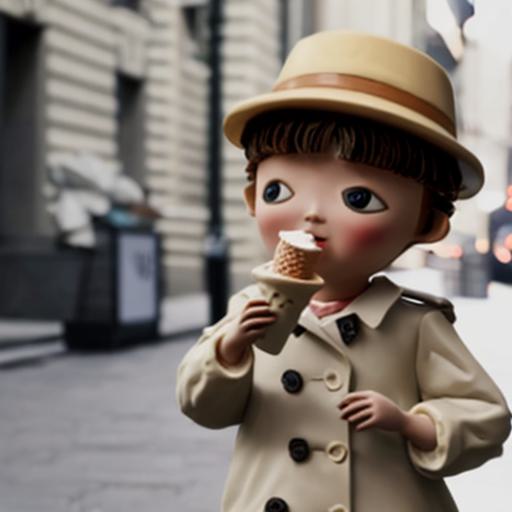} &
        \includegraphics[width=0.15\textwidth]{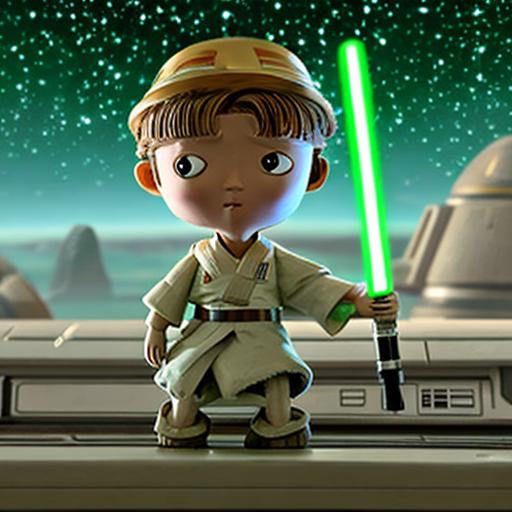} &
        \includegraphics[width=0.15\textwidth]{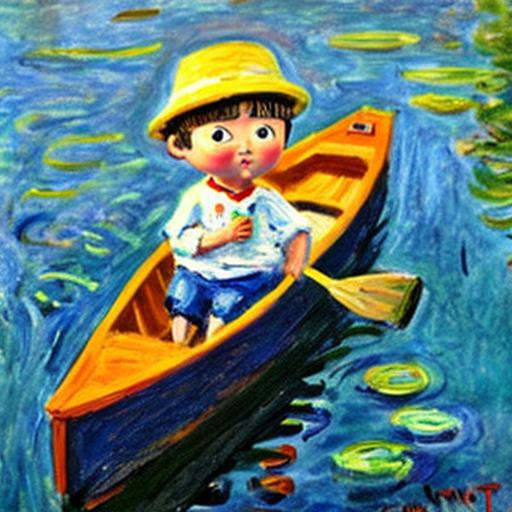} &
        \includegraphics[width=0.15\textwidth]{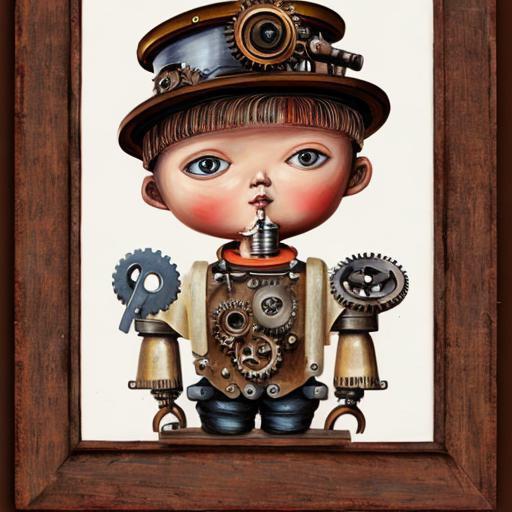} \\

        \begin{minipage}[t]{0.15\textwidth}\centering \vspace{0pt} Input Sample\end{minipage} &
        \begin{minipage}[t]{0.15\textwidth}\centering \vspace{-3mm} A S* in boots was walking its pet dog down the street with a leash\end{minipage} &
        \begin{minipage}[t]{0.15\textwidth}\centering \vspace{-3mm}A S* in a detective's trench coat and hat, eating a ice cream on the street\end{minipage} &
        \begin{minipage}[t]{0.15\textwidth}\centering \vspace{-3mm} A S* as a Jedi, with a green lightsaber, standing on a starship bridge, gazing out at distant galaxies\end{minipage}\vspace{1mm} &
        \begin{minipage}[t]{0.15\textwidth}\centering \vspace{-3mm} A painting of S* as a boatman propping a boat in the lake in the style of Monet\end{minipage} &
        \begin{minipage}[t]{0.15\textwidth}\centering \vspace{-3mm} A painting of S* as a vintage steampunk automaton, complete with gears and complex mechanical devices\end{minipage} \\

        \includegraphics[width=0.15\textwidth,height=0.15\textwidth]{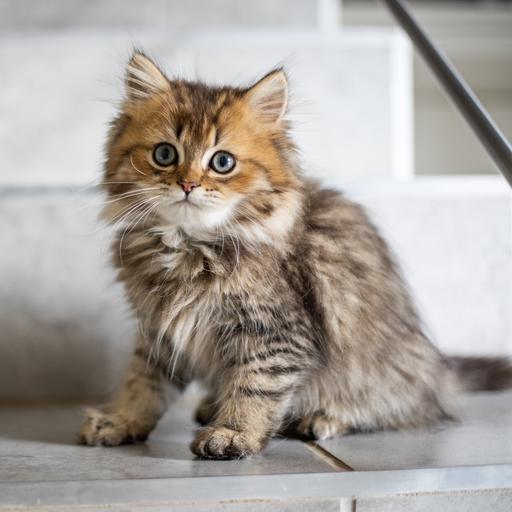} &
        \includegraphics[width=0.15\textwidth]{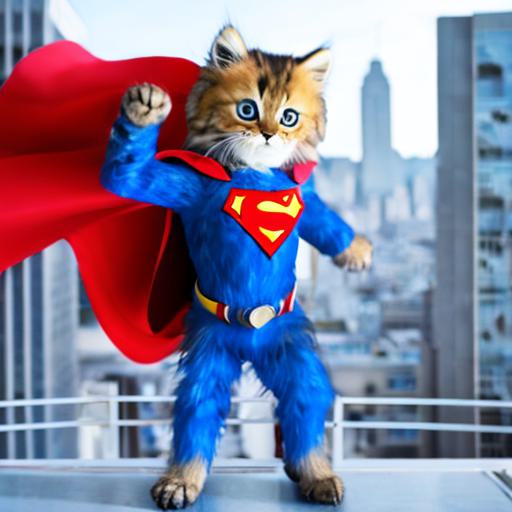} &
        \includegraphics[width=0.15\textwidth]{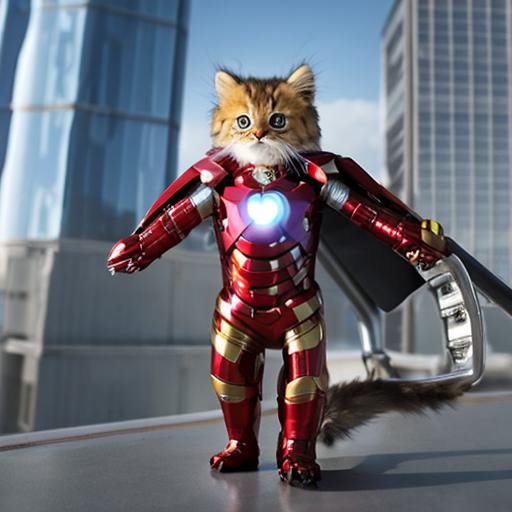} &
        \includegraphics[width=0.15\textwidth]{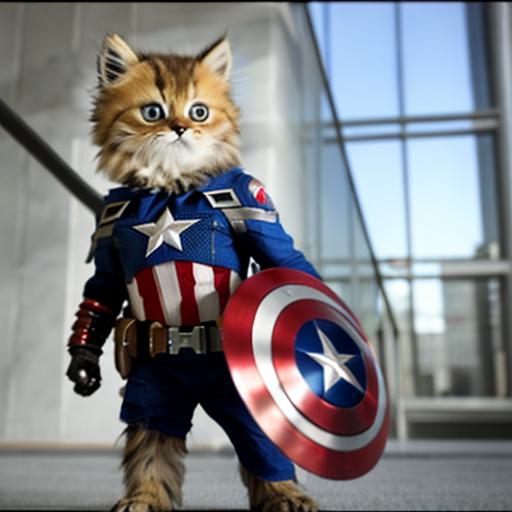} &
        \includegraphics[width=0.15\textwidth]{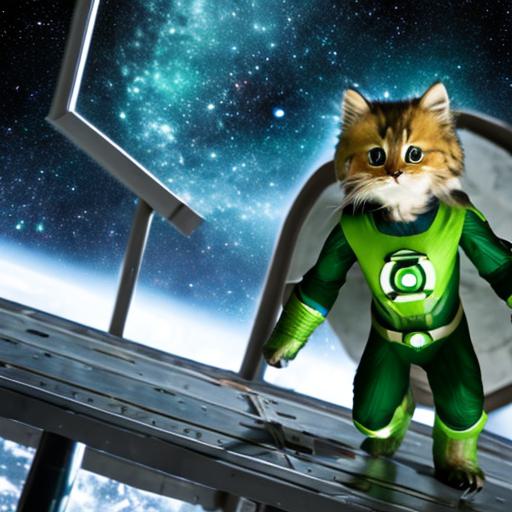} &
        \includegraphics[width=0.15\textwidth]{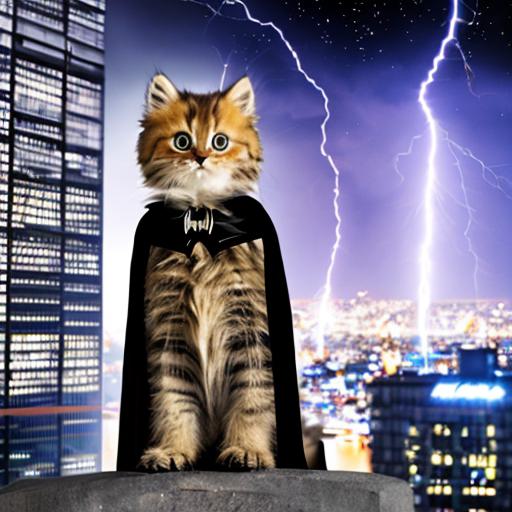} \\

        \begin{minipage}[t]{0.15\textwidth}\centering \vspace{0pt} Input Sample\end{minipage} &
        \begin{minipage}[t]{0.15\textwidth}\centering \vspace{-3mm}S* is wearing Superman's cape and the iconic blue suit with the 'S' emblem, standing on a rooftop in Metropolis at dawn\end{minipage} &
        \begin{minipage}[t]{0.15\textwidth}\centering \vspace{-3mm} A S* dressed as Iron Man, complete with a tiny suit of red and gold armor, standing confidently on a futuristic city rooftop\end{minipage} &
        \begin{minipage}[t]{0.15\textwidth}\centering \vspace{-3mm} At the Avengers headquarters, a S* dressed in a Captain America uniform patrols the area with authority\end{minipage} &
        \begin{minipage}[t]{0.15\textwidth}\centering \vspace{-3mm} S* wears a Green Lantern suit, standing on an outer space station's platform, with the vast starry sky and shining nebulae in the background\end{minipage} \vspace{1mm}&
        \begin{minipage}[t]{0.15\textwidth}\centering \vspace{-3mm} S* wears Batman's black cape, perched on the edge of a Gotham City skyscraper, with lightning in the night sky and city lights below\end{minipage} \\
        
        \includegraphics[width=0.15\textwidth,height=0.15\textwidth]{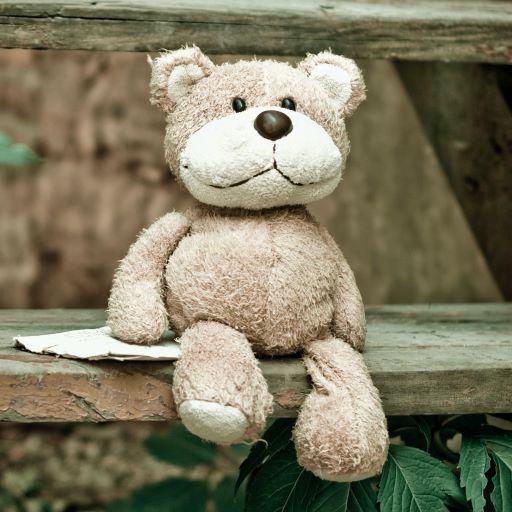} &
        \includegraphics[width=0.15\textwidth]{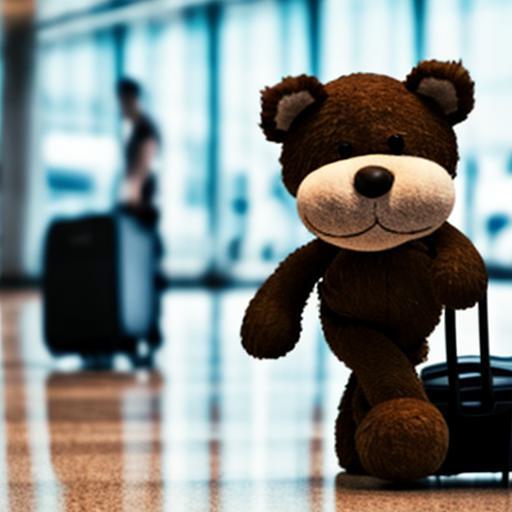} &
        \includegraphics[width=0.15\textwidth]{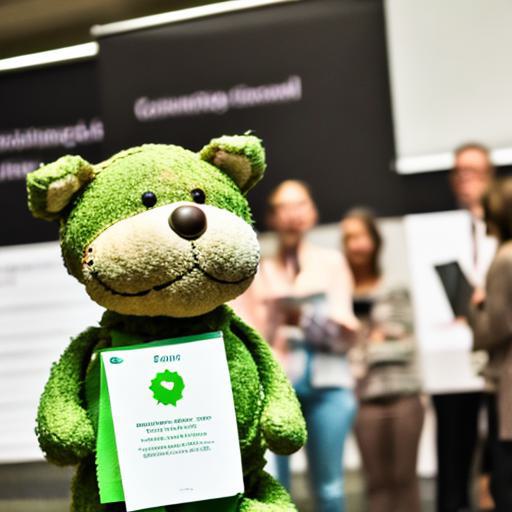} &
        \includegraphics[width=0.15\textwidth]{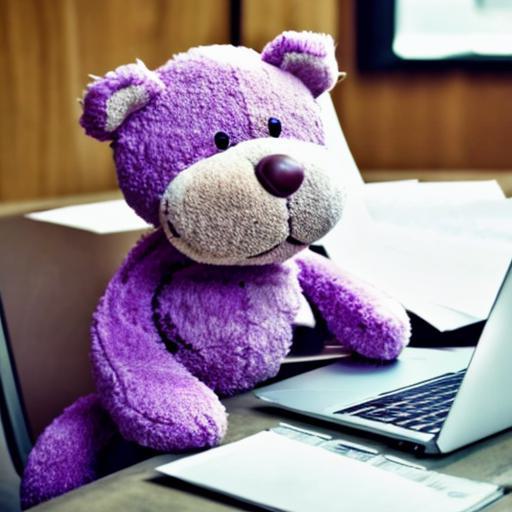} &
        \includegraphics[width=0.15\textwidth]{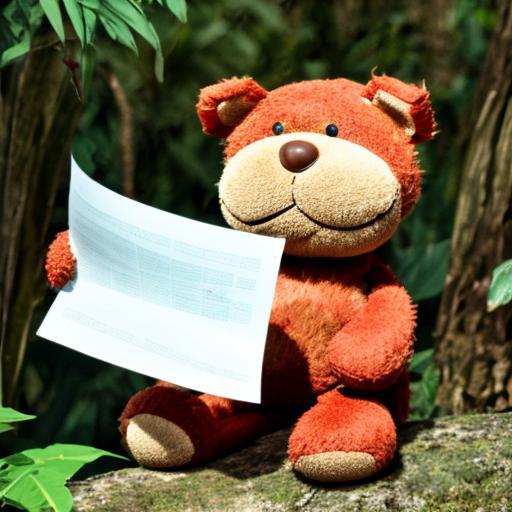}&
        \includegraphics[width=0.15\textwidth]{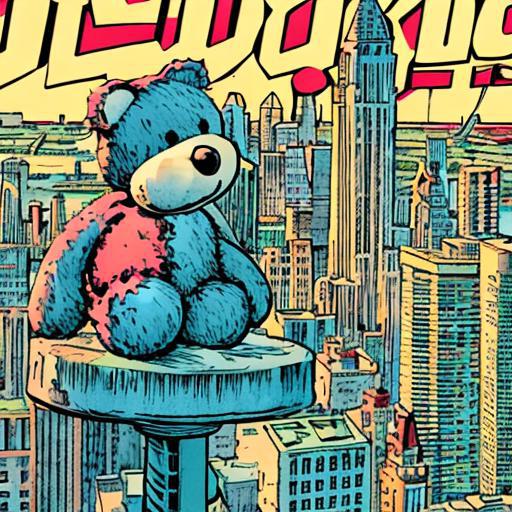} \\
        
        \begin{minipage}[t]{0.15\textwidth}\centering \vspace{0pt} Input Sample\end{minipage} &
        \begin{minipage}[t]{0.15\textwidth}\centering \vspace{-3mm} A black S* walking with his suitcase at the airport\end{minipage} &
        \begin{minipage}[t]{0.15\textwidth}\centering \vspace{-3mm} A green S* presenting a poster at a conference with people around\end{minipage} &
        \begin{minipage}[t]{0.15\textwidth}\centering \vspace{-3mm} A purple S* typing a paper on a laptop\end{minipage} &
        \begin{minipage}[t]{0.15\textwidth}\centering \vspace{-3mm} A red S* holding up his accepted paper in the jungle\end{minipage} &
        \begin{minipage}[t]{0.16\textwidth}\centering \vspace{-3mm} A S* atop a tall tower, looking out over a bustling metropolis illustrated in a vintage comic book style\end{minipage} \\
    \end{tabular}
    }
    \caption{Additional generated images by CoRe.}
    \label{fig:qualitative_evaluation_1}
\end{figure*}

\begin{figure*}[t]
    \centering
    \setlength{\tabcolsep}{0.1pt}
    {\footnotesize
    \begin{tabular}{c@{\hspace{0.12cm}} c@{\hspace{0.12cm}} c@{\hspace{0.08cm}} c@{\hspace{0.05cm}} c@{\hspace{0.1cm}} c}

        \includegraphics[width=0.15\textwidth,height=0.15\textwidth]{images/input_imgs/cat_toy.jpg} &
        \includegraphics[width=0.15\textwidth]{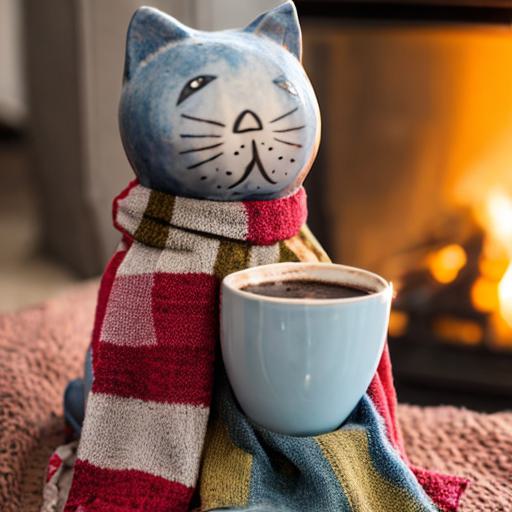} &
        \includegraphics[width=0.15\textwidth]{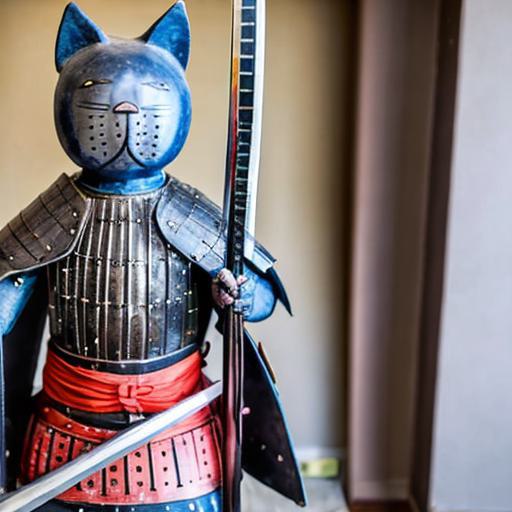} &
        \includegraphics[width=0.15\textwidth]{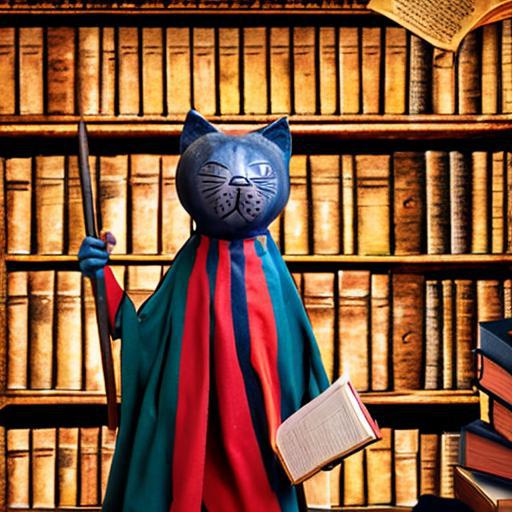} &
        \includegraphics[width=0.15\textwidth]{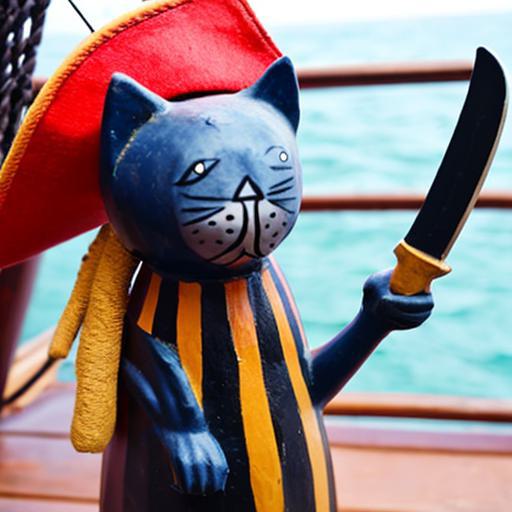}&
        \includegraphics[width=0.15\textwidth]{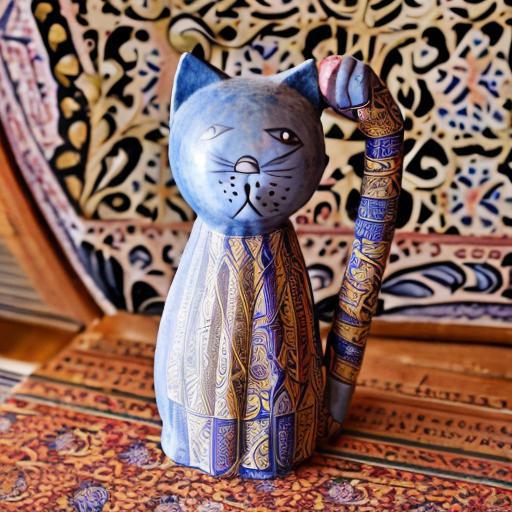} \\

        \begin{minipage}[t]{0.15\textwidth}\centering \vspace{0pt} Input Sample\end{minipage} &
        \begin{minipage}[t]{0.15\textwidth}\centering \vspace{-3mm} A S* wearing a cozy sweater and scarf, holding a cup of hot cocoa by the fireplace\end{minipage} &
        \begin{minipage}[t]{0.15\textwidth}\centering \vspace{-3mm} A S* in a full samurai armor, standing proudly with a katana\end{minipage} &
        \begin{minipage}[t]{0.17\textwidth}\centering \vspace{-3mm} S* dressed in Hogwarts robes, with a magical wand and a magical book inside an ancient library filled with towering bookshelves\end{minipage} \vspace{1mm}&
        \begin{minipage}[t]{0.15\textwidth}\centering \vspace{-3mm} A S* adorned in a pirate outfit, with a tricorn hat and a sword, standing on the deck of a pirate ship\end{minipage} &
        \begin{minipage}[t]{0.15\textwidth}\centering \vspace{-3mm} A S* in the intricate, decorative patterns of Islamic art\end{minipage} \\

\includegraphics[width=0.15\textwidth,height=0.15\textwidth]{images/input_imgs/elephant.jpg} &
        \includegraphics[width=0.15\textwidth]{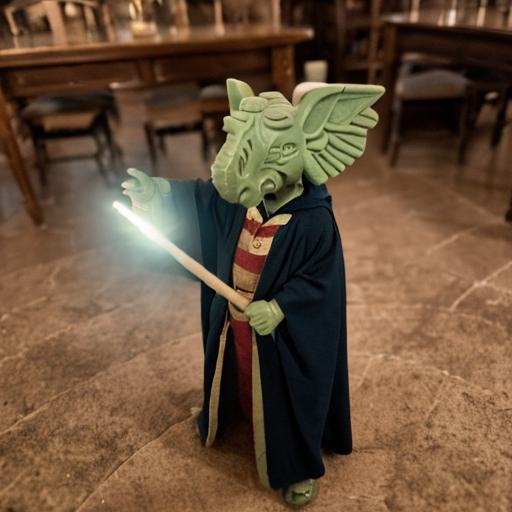} &
        \includegraphics[width=0.15\textwidth]{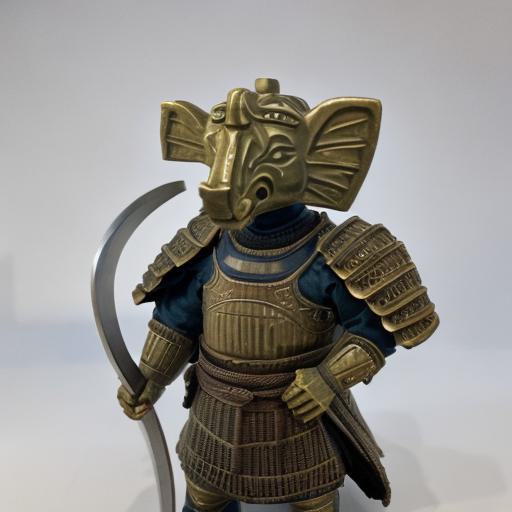} &
        \includegraphics[width=0.15\textwidth]{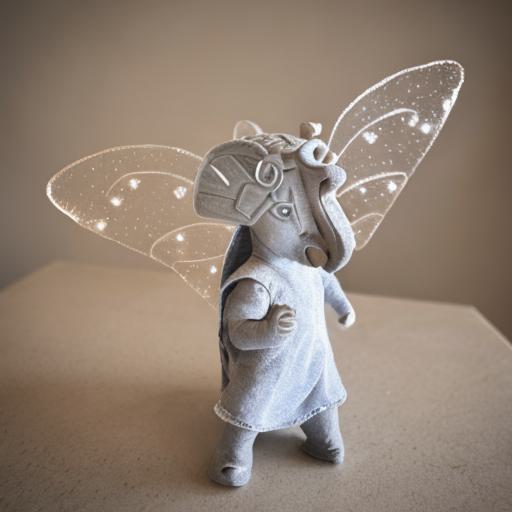}&
        \includegraphics[width=0.15\textwidth]{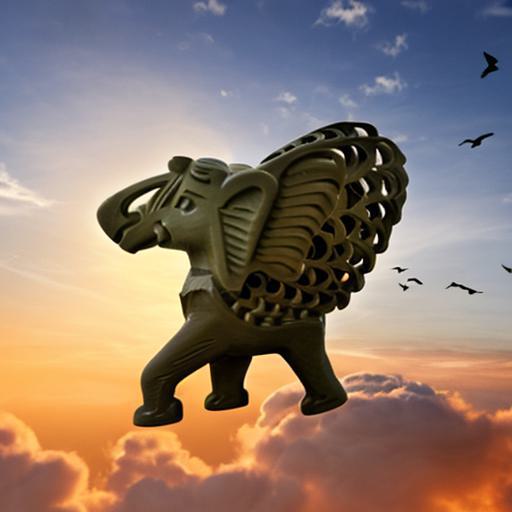} &
        \includegraphics[width=0.15\textwidth]{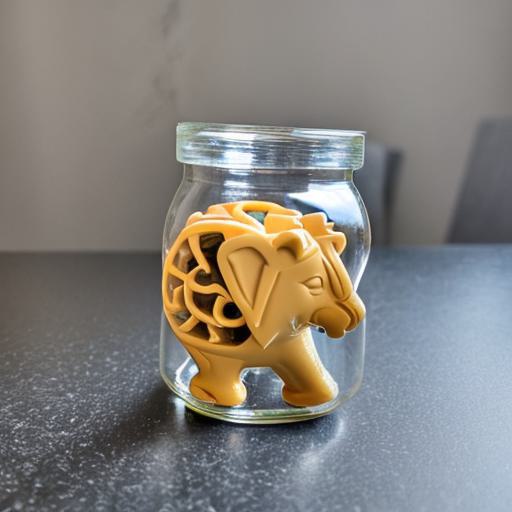} \\

        \begin{minipage}[t]{0.15\textwidth}\centering \vspace{0pt} Input Sample\end{minipage} &
        \begin{minipage}[t]{0.15\textwidth}\centering \vspace{-3mm} A S* dressed in Hogwarts robes, casting spells with a magical wand\end{minipage} &
        \begin{minipage}[t]{0.15\textwidth}\centering \vspace{-3mm} A S* in a full samurai armor, standing proudly with a katana\end{minipage} &
        \begin{minipage}[t]{0.15\textwidth}\centering \vspace{-3mm} A S* in a whimsical fairy costume, with delicate wings and sparkling dust\end{minipage} &
        \begin{minipage}[t]{0.15\textwidth}\centering \vspace{-3mm} A S* with wings, flying above the clouds at sunset, with a flock of birds soaring alongside it\end{minipage} \vspace{1mm}&
        \begin{minipage}[t]{0.15\textwidth}\centering \vspace{-3mm} A gummy candy in the shape of S* sits on the table in a glass candy jar\end{minipage} \\

\        
        \includegraphics[width=0.15\textwidth,height=0.15\textwidth]{images/input_imgs/headless_statue.jpg} &
        \includegraphics[width=0.15\textwidth]{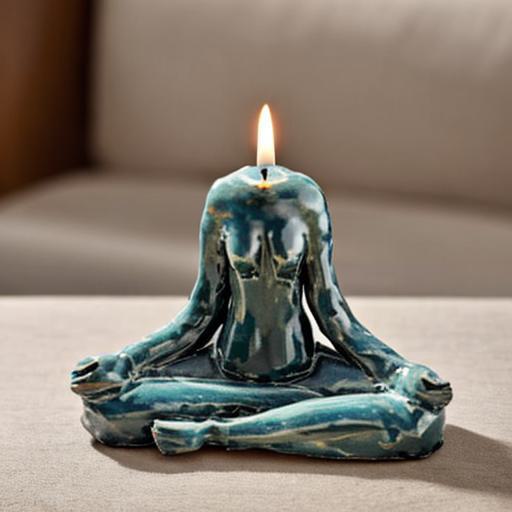} &
        \includegraphics[width=0.15\textwidth]{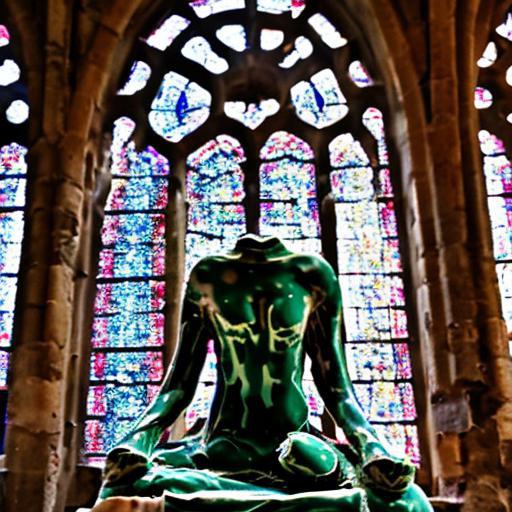} &
        \includegraphics[width=0.15\textwidth]{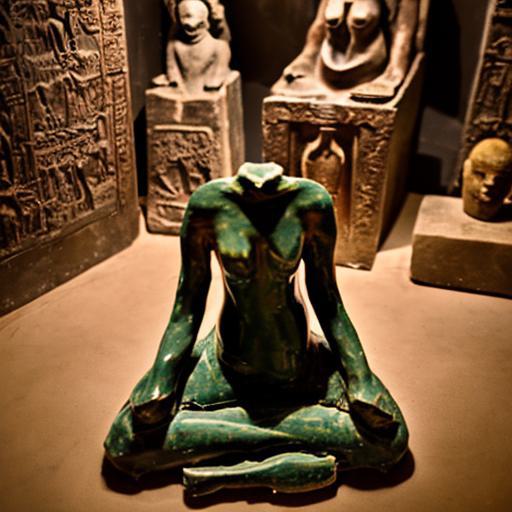}&
        \includegraphics[width=0.15\textwidth]{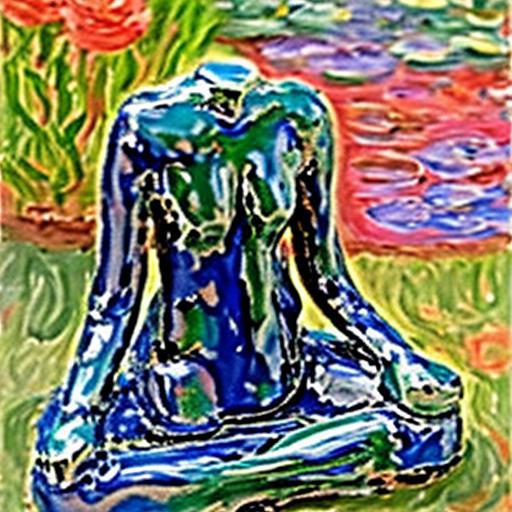} &
        \includegraphics[width=0.15\textwidth]{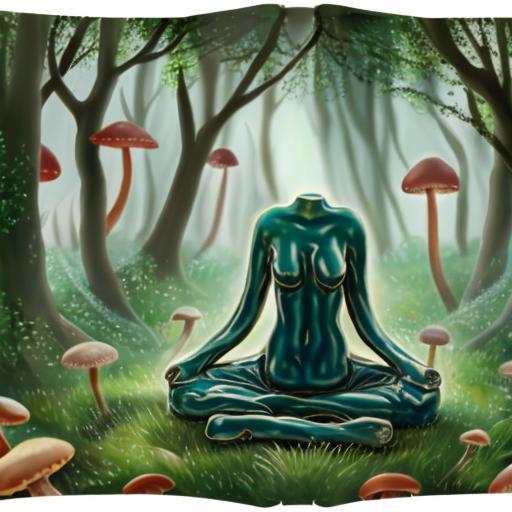} \\

        \begin{minipage}[t]{0.15\textwidth}\centering \vspace{0pt} Input Sample\end{minipage} &
        \begin{minipage}[t]{0.15\textwidth}\centering \vspace{-3mm} A S* serves as a candle holder with a flame at the top\end{minipage} &
        \begin{minipage}[t]{0.15\textwidth}\centering \vspace{-3mm} A S* in the ruins of an old cathedral, with shattered stained glass windows casting colorful light on it\end{minipage}\vspace{1mm} &
        \begin{minipage}[t]{0.15\textwidth}\centering \vspace{-3mm} A S* in a dimly lit museum hall, with other ancient artifacts and sculpture around it\end{minipage} &
        \begin{minipage}[t]{0.15\textwidth}\centering \vspace{-3mm} A colorful S* in the style of Monet, surrounded by a lush impressionist garden with light reflecting off its glossy surface\end{minipage} &
        \begin{minipage}[t]{0.15\textwidth}\centering \vspace{-3mm} A S* in a mystical forest glade, surrounded by glowing mushrooms and ethereal light\end{minipage} \\

\includegraphics[width=0.15\textwidth,height=0.15\textwidth]{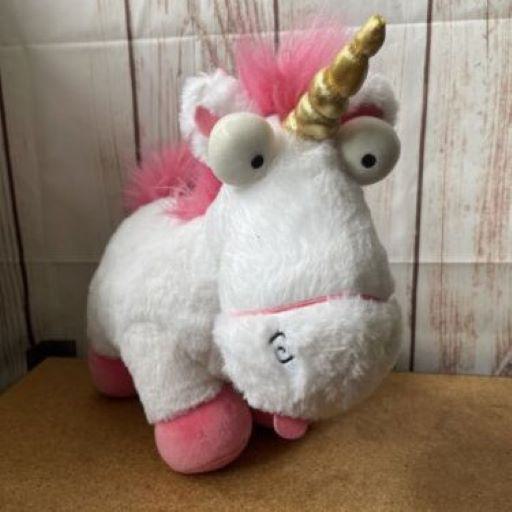} &
        \includegraphics[width=0.15\textwidth]{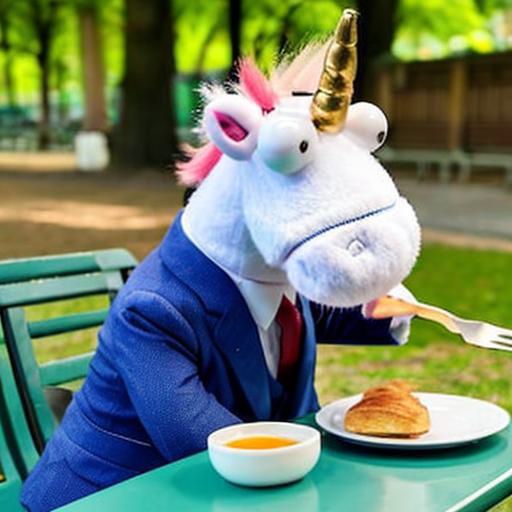} &
        \includegraphics[width=0.15\textwidth]{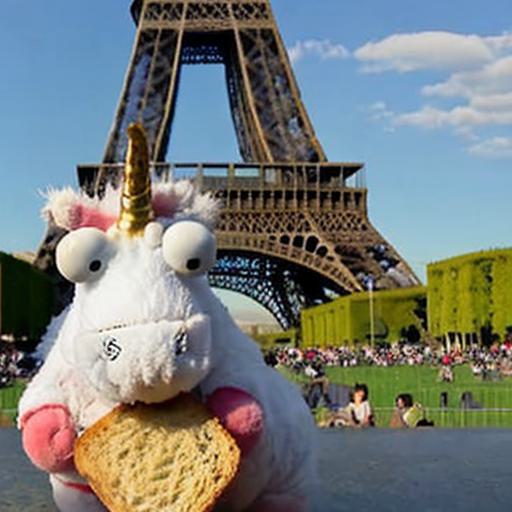} &
        \includegraphics[width=0.15\textwidth]{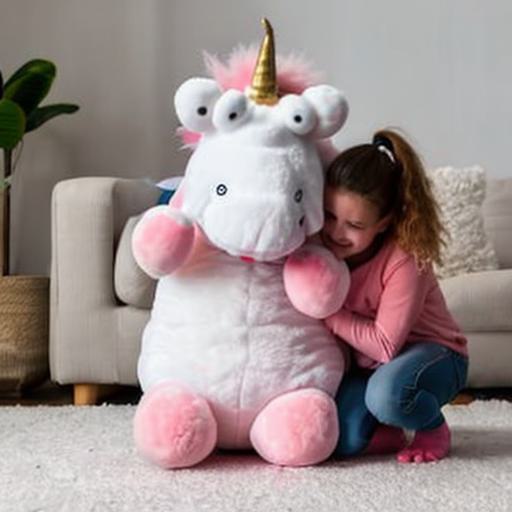} &
        \includegraphics[width=0.15\textwidth]{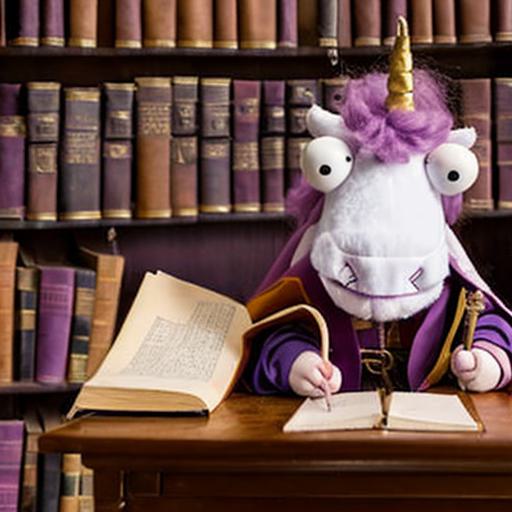}&
        \includegraphics[width=0.15\textwidth]{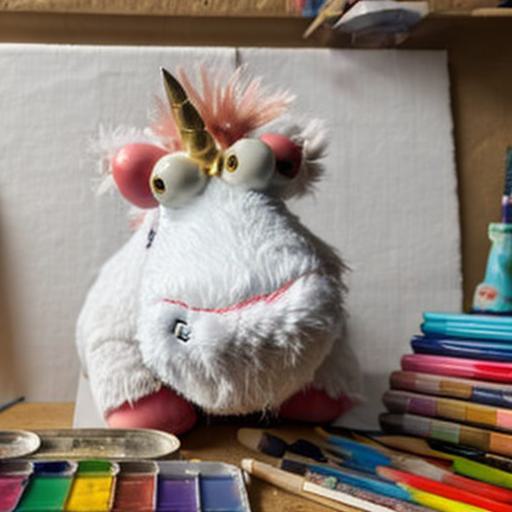} \\

        \begin{minipage}[t]{0.15\textwidth}\centering \vspace{0pt} Input Sample\end{minipage} &
        \begin{minipage}[t]{0.15\textwidth}\centering \vspace{-3mm} A S* in a suit is enjoying a breakfast with a fork in a park\end{minipage} &
        \begin{minipage}[t]{0.15\textwidth}\centering \vspace{-3mm} A S* is eating bread in front of the Eiffel Tower\end{minipage} &
        \begin{minipage}[t]{0.15\textwidth}\centering \vspace{-3mm} A S* is being hugged by a little girl on a plush carpet in the living room\end{minipage} \vspace{1mm}&
        \begin{minipage}[t]{0.15\textwidth}\centering \vspace{-3mm} A S* dressed as a purple wizard sits at a desk in a medieval library\end{minipage} &
        \begin{minipage}[t]{0.17\textwidth}\centering \vspace{-3mm} A curious S* peeking from beneath a canvas, with art supplies on a table above\end{minipage} \\
        \includegraphics[width=0.15\textwidth,height=0.15\textwidth]{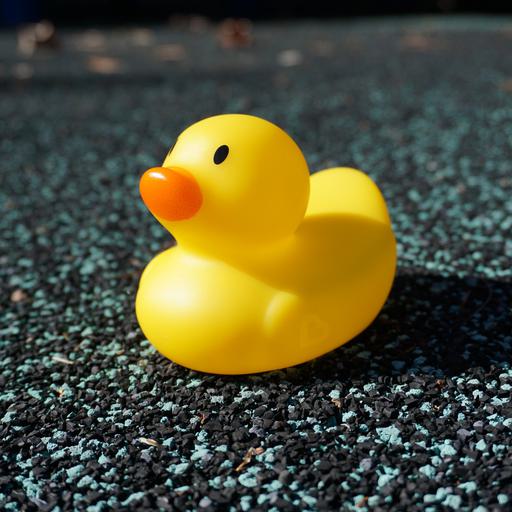} &
        \includegraphics[width=0.15\textwidth]{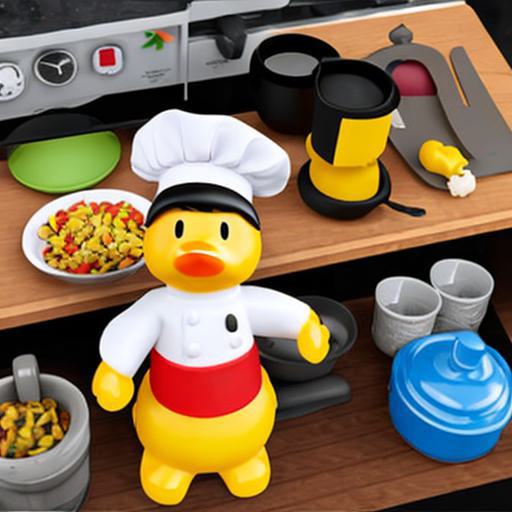} &
        \includegraphics[width=0.15\textwidth]{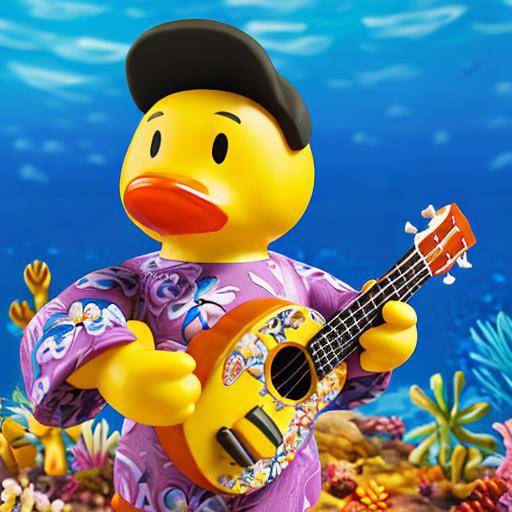} &
        \includegraphics[width=0.15\textwidth]{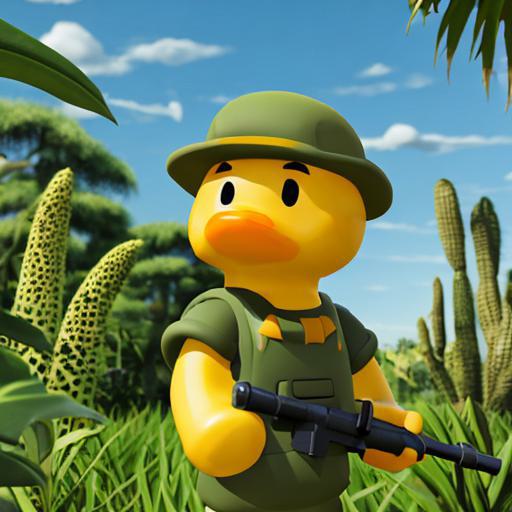} &
        \includegraphics[width=0.15\textwidth]{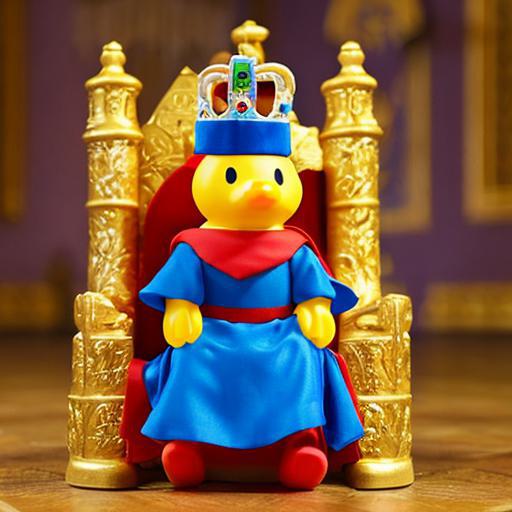} &
        \includegraphics[width=0.15\textwidth]{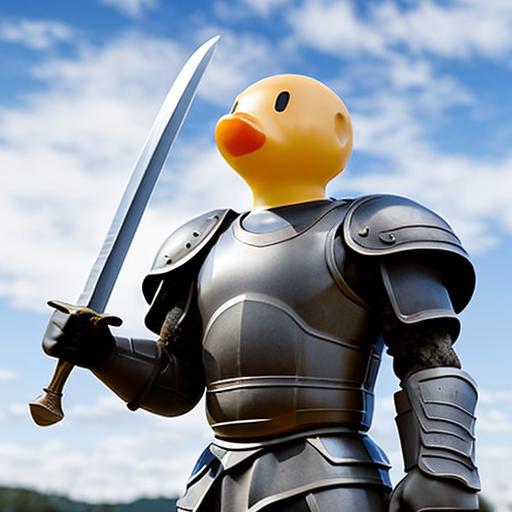}\\

        \begin{minipage}[t]{0.15\textwidth}\centering \vspace{0pt} Input Sample\end{minipage} &
        \begin{minipage}[t]{0.17\textwidth}\centering \vspace{-3mm} A S* dressed as a chef, with a tiny hat and apron, standing in a toy kitchen surrounded by miniature pots, pans, and a delicious-\\looking toy meal\end{minipage} &
        \begin{minipage}[t]{0.15\textwidth}\centering \vspace{-3mm} A S* wearing a Hawaiian shirt and flower lei, joyfully playing a ukulele under the sea\end{minipage} &
        \begin{minipage}[t]{0.15\textwidth}\centering \vspace{-3mm} A S* wearing a safari hat and holding a tranquilizer gun, standing among prehistoric plants\end{minipage} &
        \begin{minipage}[t]{0.15\textwidth}\centering \vspace{-3mm} A S* dressed as a royal prince, complete with a crown and cape, sitting on a miniature throne in a lavish toy castle\end{minipage} &
        \begin{minipage}[t]{0.15\textwidth}\centering \vspace{-3mm} A S* wearing a warrior's armor, holding a sword, ready to defend the kingdom\end{minipage} \\

    \end{tabular}
    }
    \caption{Additional generated images by CoRe.}
    \label{fig:qualitative_evaluation_2}

\end{figure*}

In Figures~\ref{fig:qualitative_evaluation_1} and~\ref{fig:qualitative_evaluation_2}, we present additional qualitative results generated by our method using a variety of prompts. 

\section{D\ \ \ Results for Test-Time Optimization}
\begin{figure*}[t]
    \centering
    \renewcommand{\arraystretch}{0.3}
    {
    \vspace{-0.5cm}
    \begin{tabular}{c@{\hspace{16pt}}c@{\hspace{8pt}}c@{\hspace{24pt}}c@{\hspace{16pt}}c@{\hspace{8pt}}c}
        
        Input & w/o TTO & w/ TTO & Input & w/o TTO & w/ TTO \\

        \raisebox{0.15\height}{\includegraphics[width=0.1\textwidth]{images/input_imgs/cat.jpg}} &
        \includegraphics[width=0.13\textwidth]{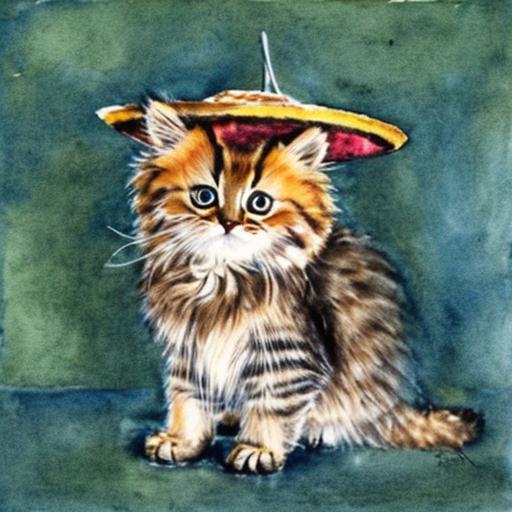} &
        \includegraphics[width=0.13\textwidth]{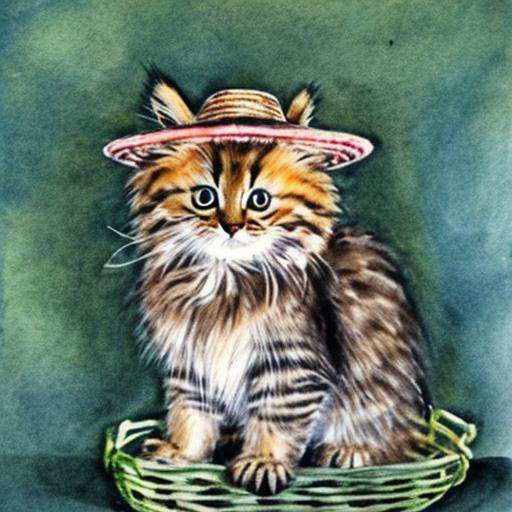} &
        
        \raisebox{0.15\height}{\includegraphics[width=0.1\textwidth]{images/input_imgs/teddy_bear.jpg}} &
        \includegraphics[width=0.13\textwidth]{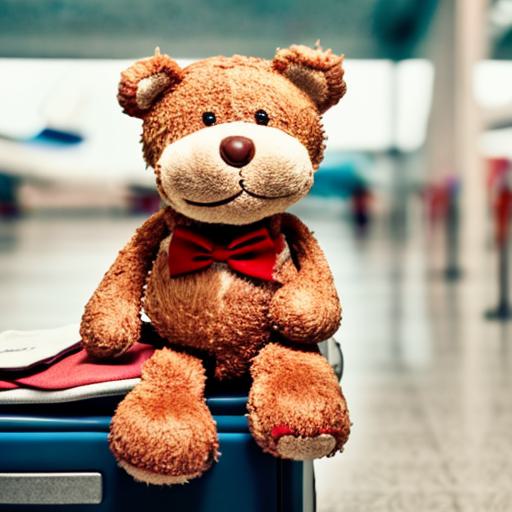} &
        \includegraphics[width=0.13\textwidth]{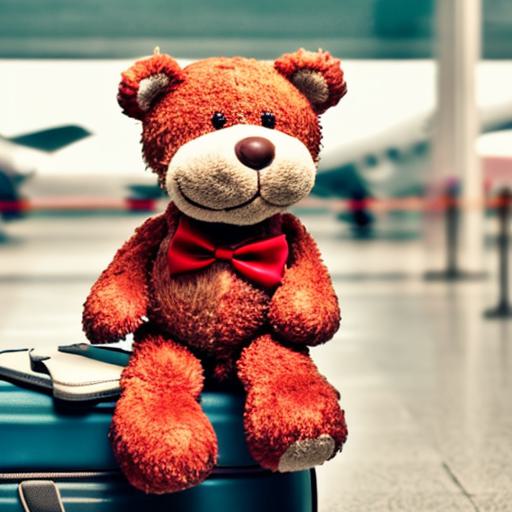} \\
        
        & \multicolumn{2}{@{\hspace{-16pt}}c}{Watercolor painting of a S*  }& 
        & \multicolumn{2}{c}{A \textcolor{red}{red} S* with a beautiful bowtie   } \\
        & \multicolumn{2}{@{\hspace{-16pt}}c}{wearing a sombrero in a \textcolor{red}{basket}}& 
        & \multicolumn{2}{c}{sitting on a suitcase at the airport \vspace{1mm}} \\
      
        \raisebox{0.15\height}{\includegraphics[width=0.1\textwidth]{images/input_imgs/fluffy.jpg}} &
        \includegraphics[width=0.13\textwidth]{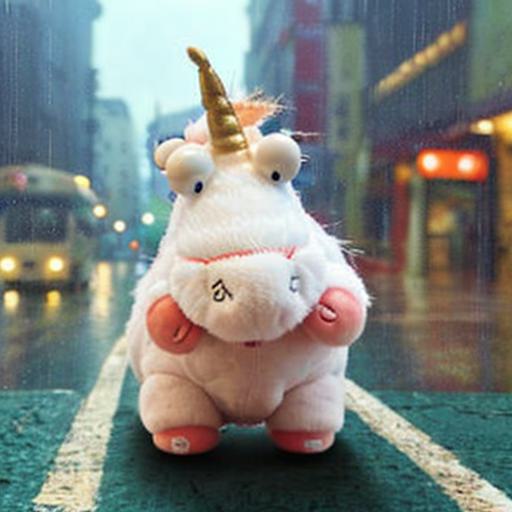} &
        \includegraphics[width=0.13\textwidth]{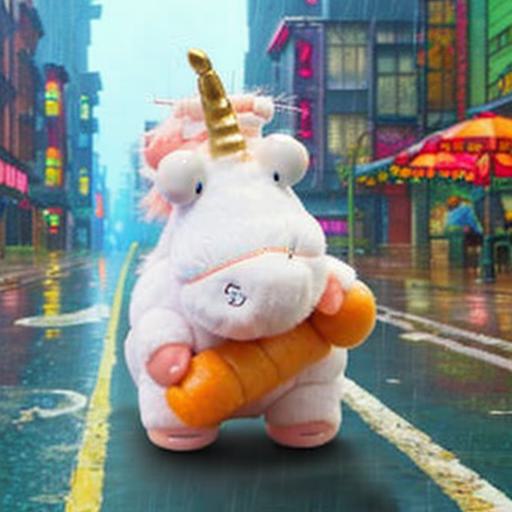} &
        
        \raisebox{0.15\height}{\includegraphics[width=0.1\textwidth]{images/input_imgs/cat_toy.jpg}} &
        \includegraphics[width=0.13\textwidth]{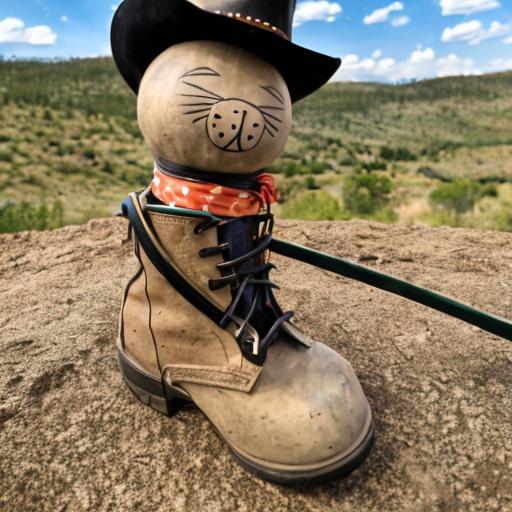} &
        \includegraphics[width=0.13\textwidth]{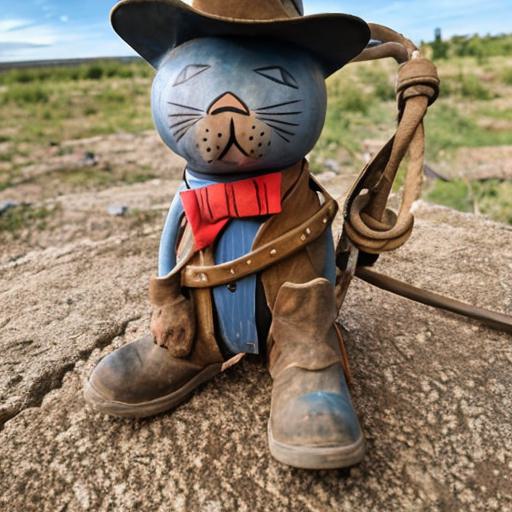} \\

        & \multicolumn{2}{@{\hspace{-16pt}}c}{A S* eating a \textcolor{red}{carrot} in } &
        & \multicolumn{2}{c}{A S* wearing a cowboy hat } \\
        & \multicolumn{2}{@{\hspace{-16pt}}c}{the rainy streets}&
        & \multicolumn{2}{c}{and \textcolor{red}{boots} holding a \textcolor{red}{lasso}\vspace{0.05mm} }\\
        &&&& \multicolumn{2}{c}{ standing in the Wild West \vspace{1mm}} \\
        
    \end{tabular}
    }
    \vspace{-0.2cm}
    \caption{Additional results for Test-Time Optimization (TTO).}
    \vspace{0.3cm}
    \label{fig:additional_test-time}
    
\end{figure*}

In Figure~\ref{fig:additional_test-time}, we present additional results of employing CoRe as a test-time optimization technique. 

\begin{figure}[t]
 \centering
 \includegraphics[width=1.\linewidth]{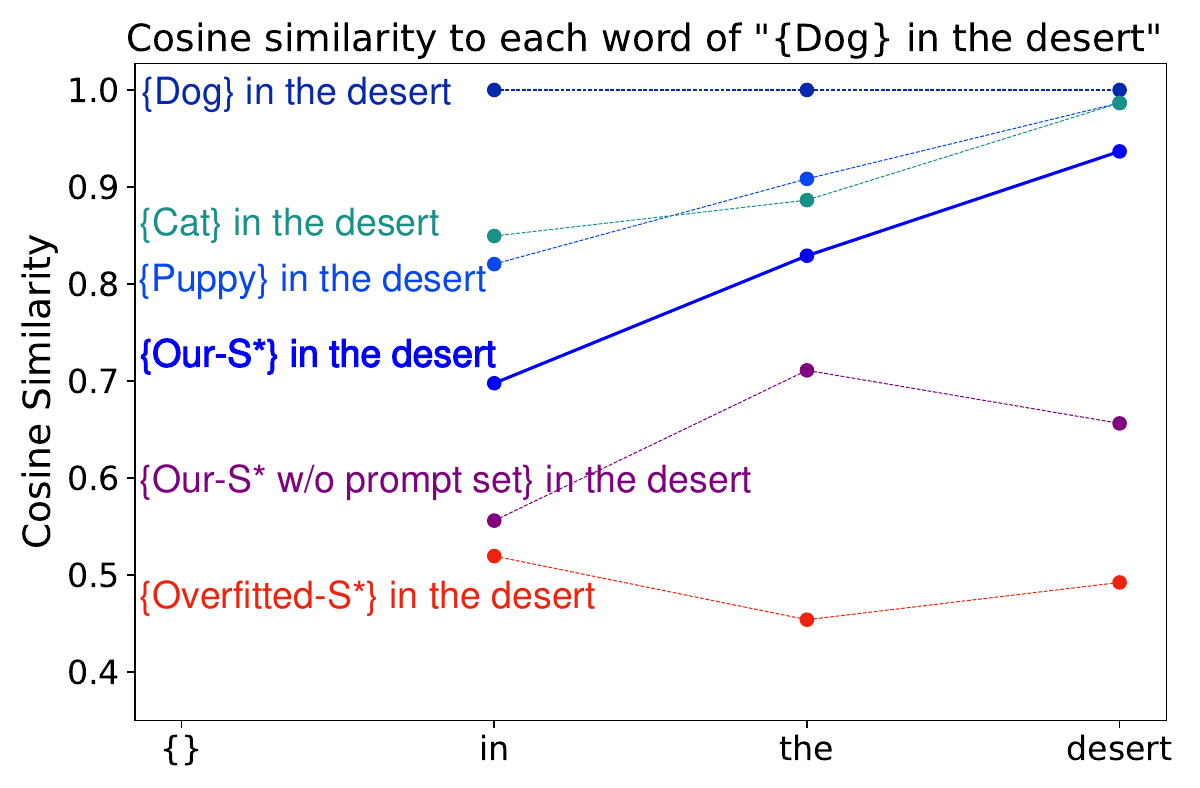}
 \includegraphics[width=1\linewidth]{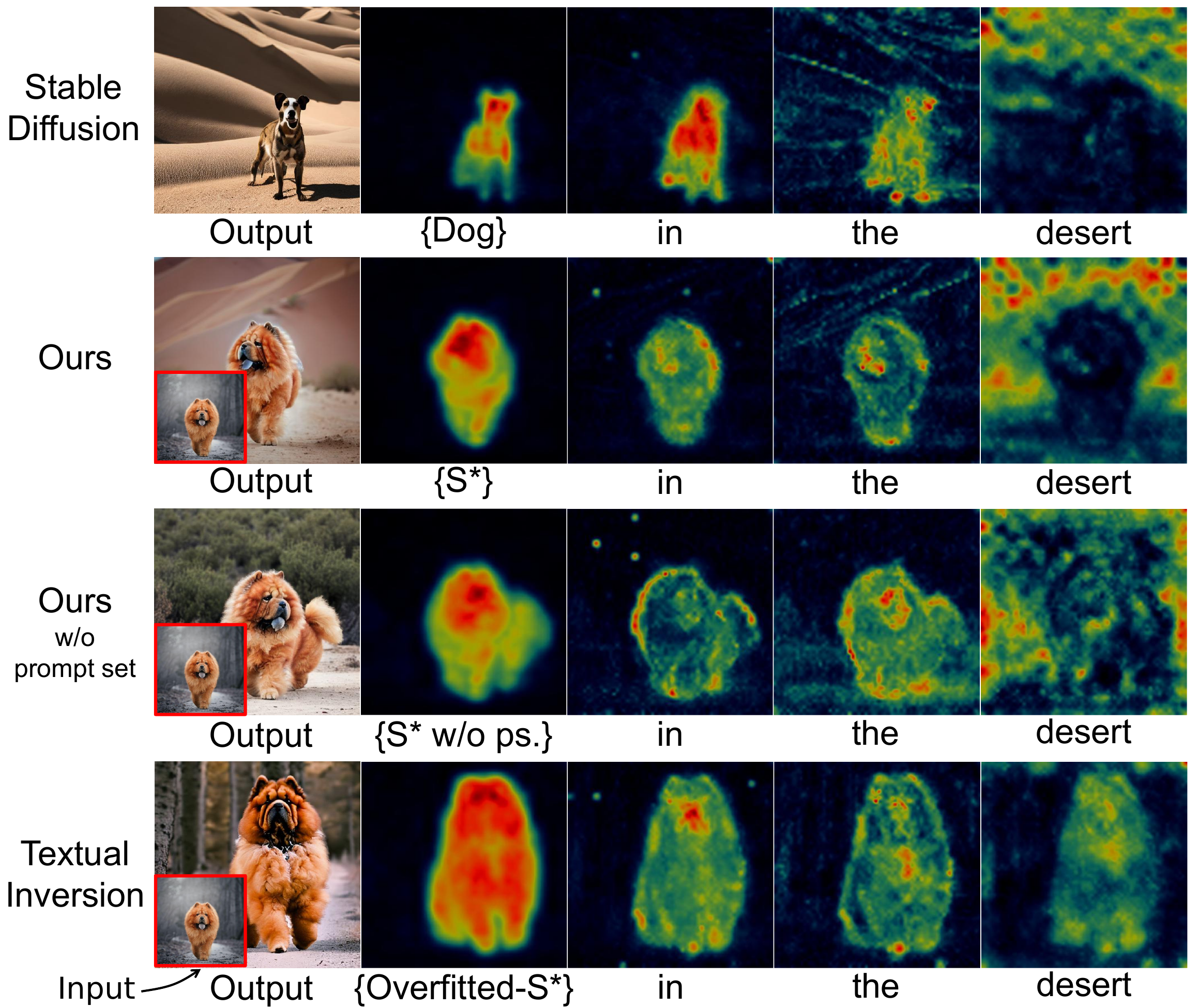}

\caption{The cosine similarity between the output embeddings of the prompt ``\{dog\} in the desert" and the object-switched prompts. Here, ``w/o prompt set'' means that the regularization prompt set is not used, applying the proposed regularization strategies directly to the training prompt ``a photo of \{\}''.}

\label{fig:appendix_motivation}
\end{figure}

\section{E\ \ \ Additional Ablation Study}

\begin{figure*}[b]
    \centering
    \renewcommand{\arraystretch}{0.3}
    \setlength{\tabcolsep}{3pt}
    {

    \begin{tabular}{c c c c c}

        \begin{tabular}{c} Input \end{tabular} &
        \begin{tabular}{c} w/o CER \end{tabular} &
        \begin{tabular}{c} w/o CAR\end{tabular} &
        \begin{tabular}{c} w/o Rescle\end{tabular} &
        \begin{tabular}{c} Full \end{tabular} \\

        \includegraphics[width=0.127\textwidth]{images/input_imgs/child_doll.jpg} &
        \includegraphics[width=0.127\textwidth]{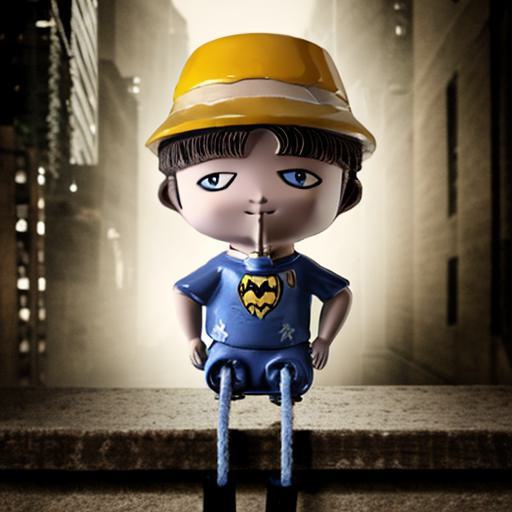} &
        \includegraphics[width=0.127\textwidth]{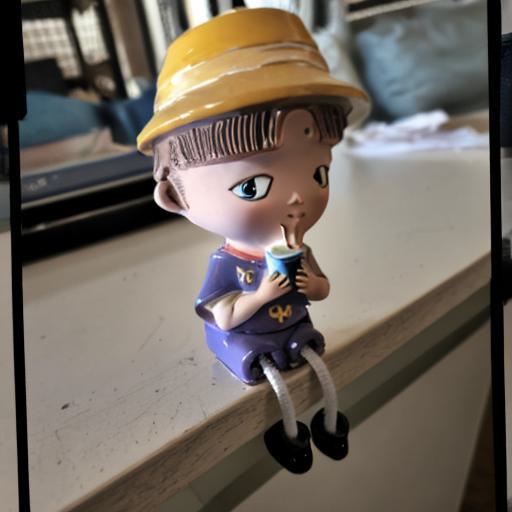} &
        \includegraphics[width=0.127\textwidth]{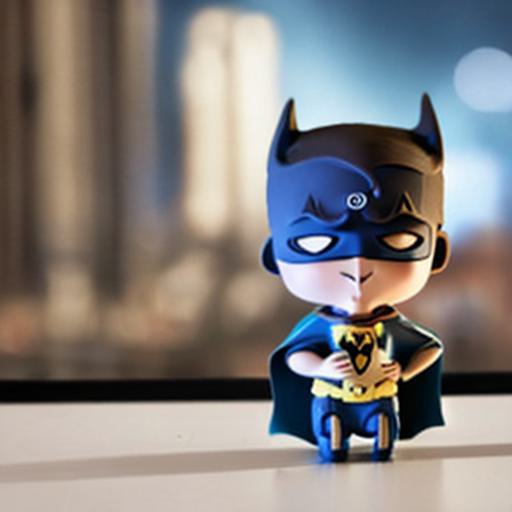}&
        \includegraphics[width=0.127\textwidth]{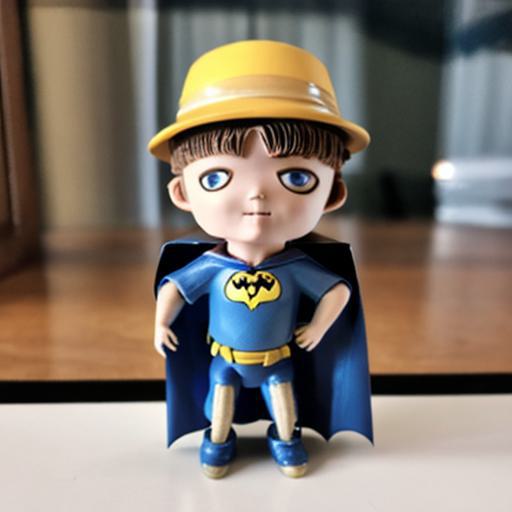} \\

        \raisebox{0.06\textwidth}{\begin{tabular}{c} A S* as\\ a Batman\end{tabular}}&
        \includegraphics[width=0.127\textwidth]{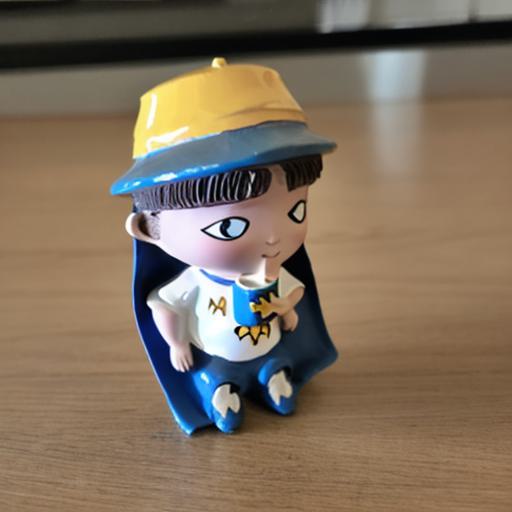} &
        \includegraphics[width=0.127\textwidth]{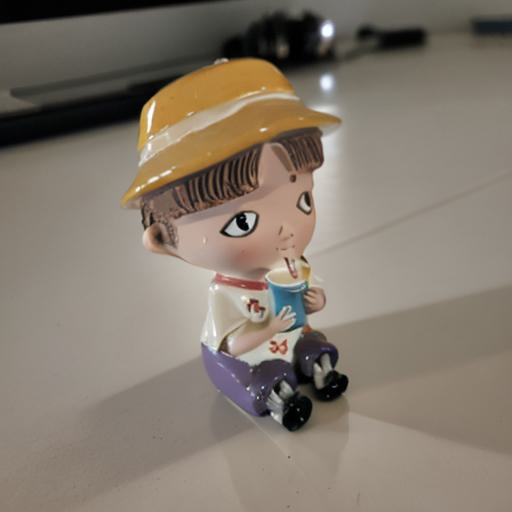} &
        \includegraphics[width=0.127\textwidth]{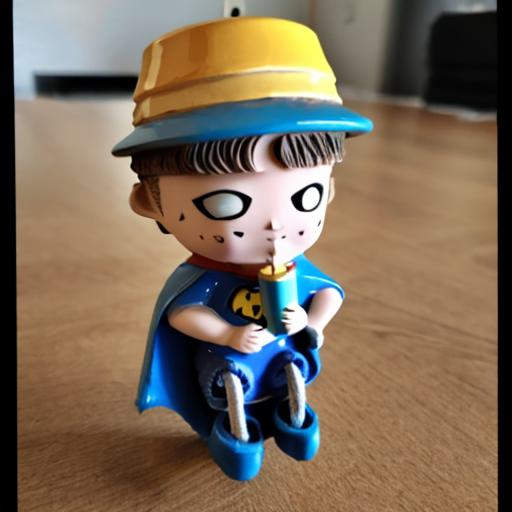}&
        \includegraphics[width=0.127\textwidth]{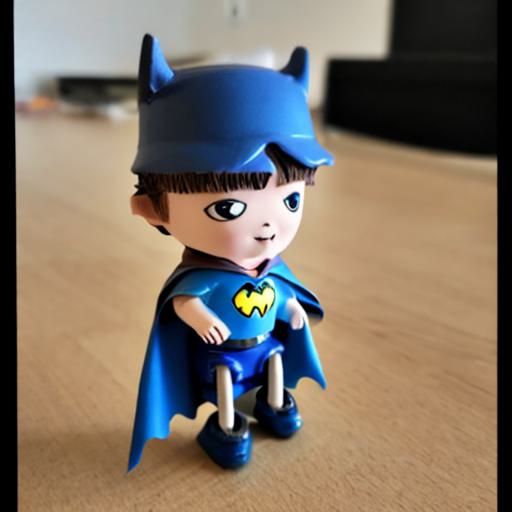} \\

        \includegraphics[width=0.127\textwidth]{images/input_imgs/asian_puppet.jpg} &
        \includegraphics[width=0.127\textwidth]{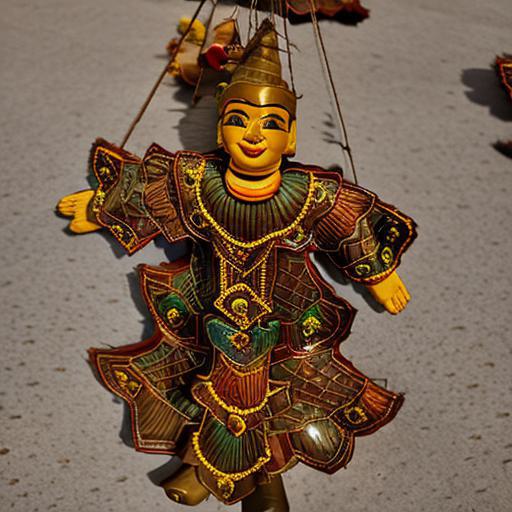} &
        \includegraphics[width=0.127\textwidth]{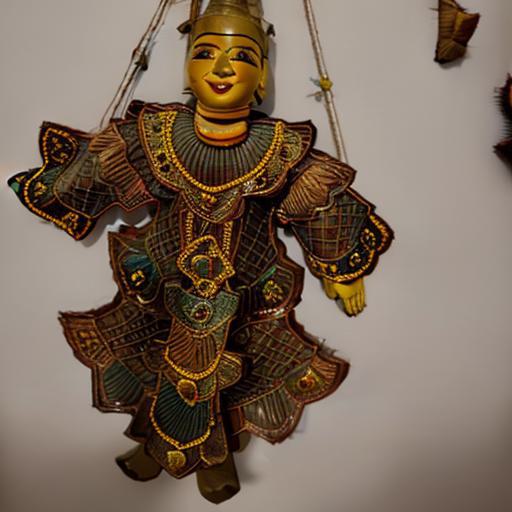} &
        \includegraphics[width=0.127\textwidth]{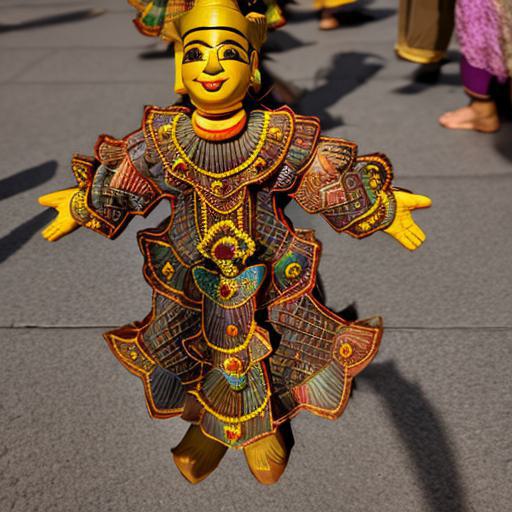}&
        \includegraphics[width=0.127\textwidth]{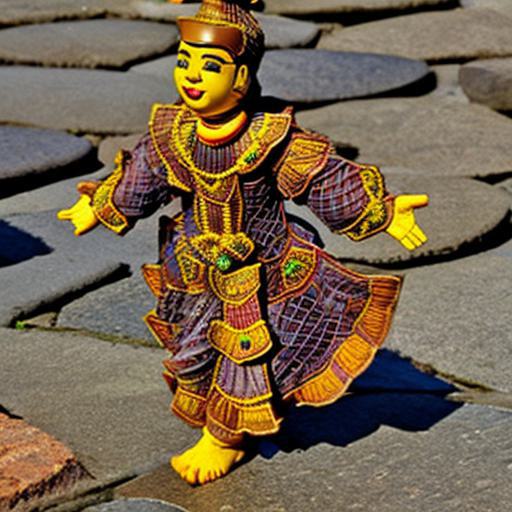} \\

        \raisebox{0.06\textwidth}{\begin{tabular}{c} A {S*} hopping\\ on stepping \\stones\end{tabular}}&
        \includegraphics[width=0.127\textwidth]{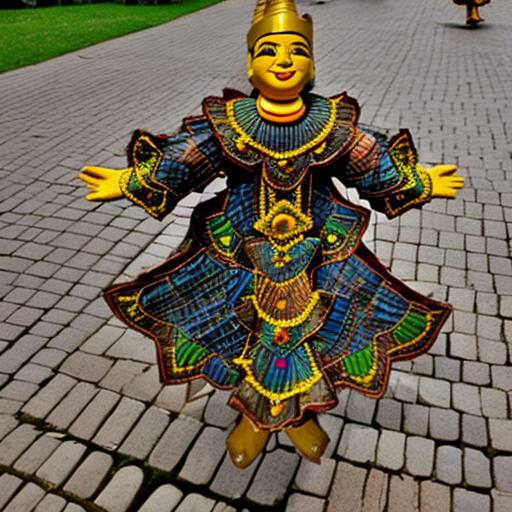} &
        \includegraphics[width=0.127\textwidth]{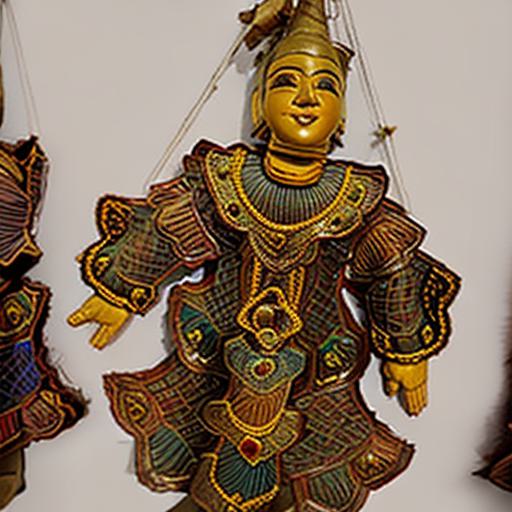} &
        \includegraphics[width=0.127\textwidth]{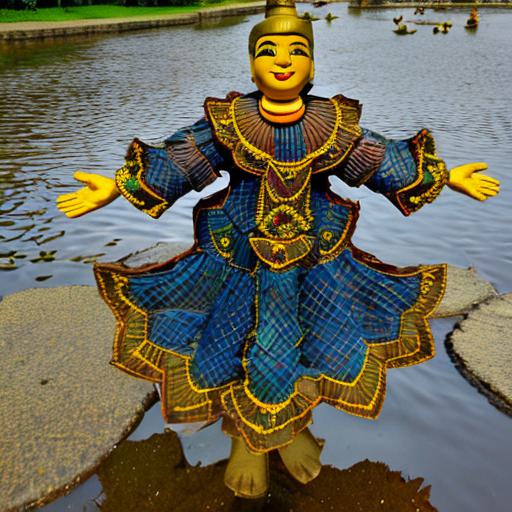}&
        \includegraphics[width=0.127\textwidth]{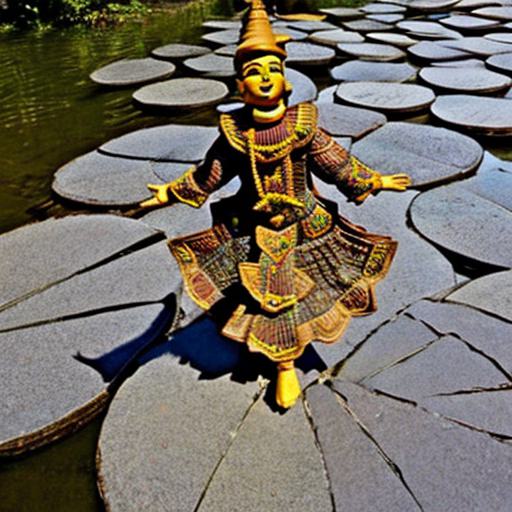} \\

        \includegraphics[width=0.127\textwidth]{images/input_imgs/grey_sloth.jpg} &
        \includegraphics[width=0.127\textwidth]{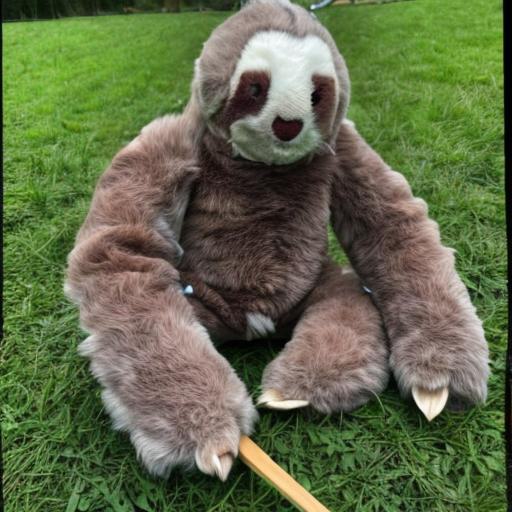} &
        \includegraphics[width=0.127\textwidth]{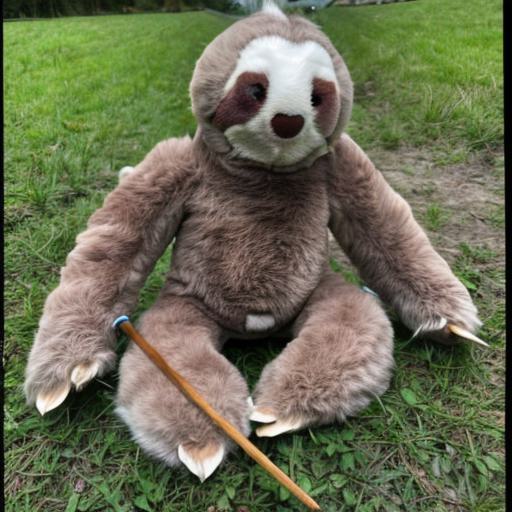} &
        \includegraphics[width=0.127\textwidth]{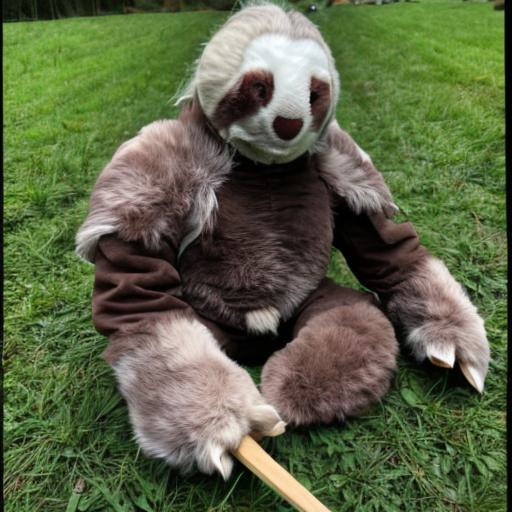}&
        \includegraphics[width=0.127\textwidth]{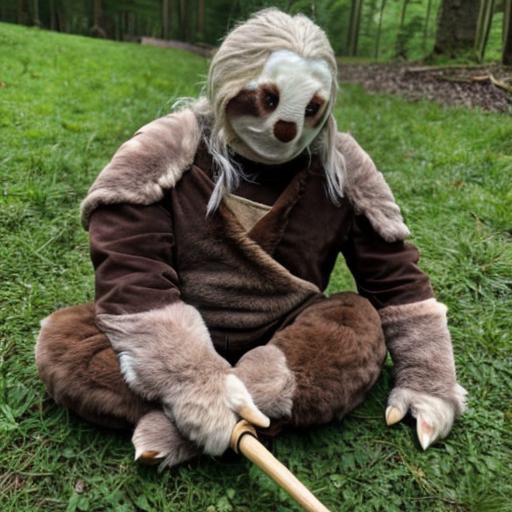} \\

        \raisebox{0.06\textwidth}{\begin{tabular}{c} A {S*} as\\ a witcher\end{tabular}}&
        \includegraphics[width=0.127\textwidth]{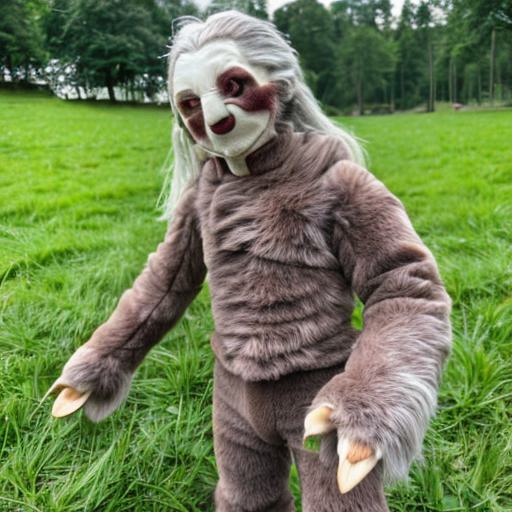} &
        \includegraphics[width=0.127\textwidth]{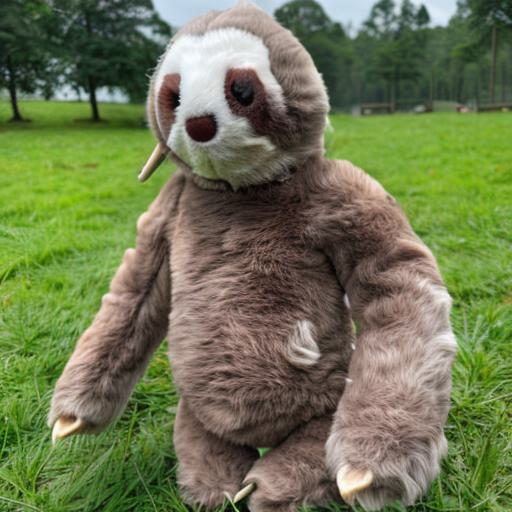} &
        \includegraphics[width=0.127\textwidth]{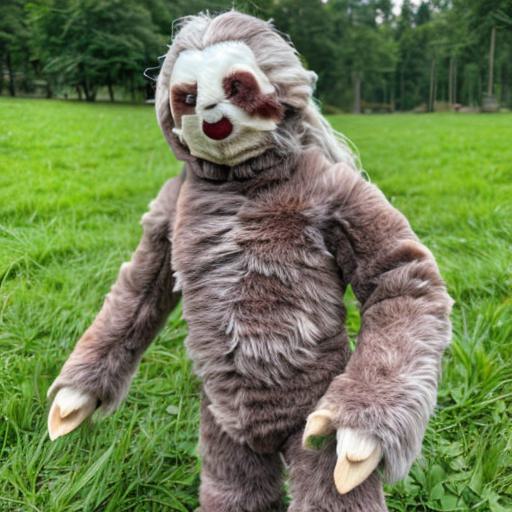}&
        \includegraphics[width=0.127\textwidth]{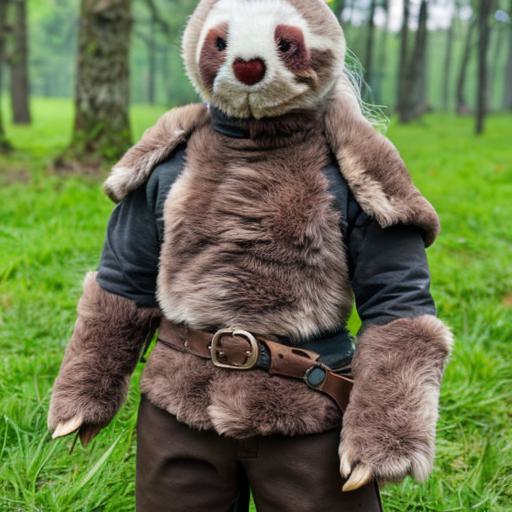} \\

    \end{tabular}
    }
    \caption{Additional ablation study. We compare models trained without Context Embedding Regularization (w/o CER), without Context Attention Regularization (w/o CAR), and without embedding rescaling strategy (w/o Rescale).}
    \label{fig:additional_ablation_study}
\end{figure*}

The results of the quantitative ablation study are presented in Table\ref{tab:ablation_evaluation}, which are consistent with the qualitative results in Figure~\ref{fig:ablation_study}. The absence of the context embedding regularization leads to degradation in both identity preservation and text alignment. The model without the context attention regularization achieves higher CLIP-I and DINO scores but a lower CLIP-T score, indicating a tendency to overfit the concept. The model without the embedding rescaling strategy exhibits slight degradation in both text alignment and identity preservation. Additionally, Figure~\ref{fig:additional_ablation_study} shows more qualitative results of the ablation study. 

\section{F\ \ \ Prompts for Quantitative Evaluation}
\label{sec:appendix_prompt_list}
Following~\cite{tewel2023keylocked}, we categorize the concepts into two groups: animate objects (e.g., ``cat'' and ``child doll'') and inanimate objects (e.g., ``clock'' and ``berry bowl''). Accordingly, we utilize two corresponding sets of prompts for quantitative evaluation, as detailed in Table~\ref{tab:prompt_list}.

\section{G\ \ \ Implementation Details}
\label{sec:appendix_implementation}
Our implementation is based on the publicly available Stable Diffusion v2-1. In the first training stage, the text embedding of $S_*$ is initialized with a super-category token. The regularization prompt set is detailed in the subsequent section. We optimize the text embedding of $S_*$ using CoRe for 300 steps with a batch size of 6 and a learning rate of 5e-3. The hyper-parameters $\lambda_{\text{emb}}$ and $\lambda_{\text{attn}}$ are set to 1.5e-4 and 0.05, respectively. Additionally, we apply the embedding rescaling strategy during the intermediate phase of the optimization (from 120 to 180 steps), as we empirically find that rescaling at the beginning or end phases can lead to degraded identity preservation, likely due to the information loss introduced by rescaling. In the second training stage, we fine-tune the entire U-Net for 1,000 steps with a batch size of 4 and a learning rate of 2e-6. For all the baseline methods, we utilize their official implementations and follow the hyper-parameters specified in their papers. All experiments are conducted on a single Nvidia A100 GPU.

\begin{figure}[h]
    \centering
    \includegraphics[width=1.0\linewidth]{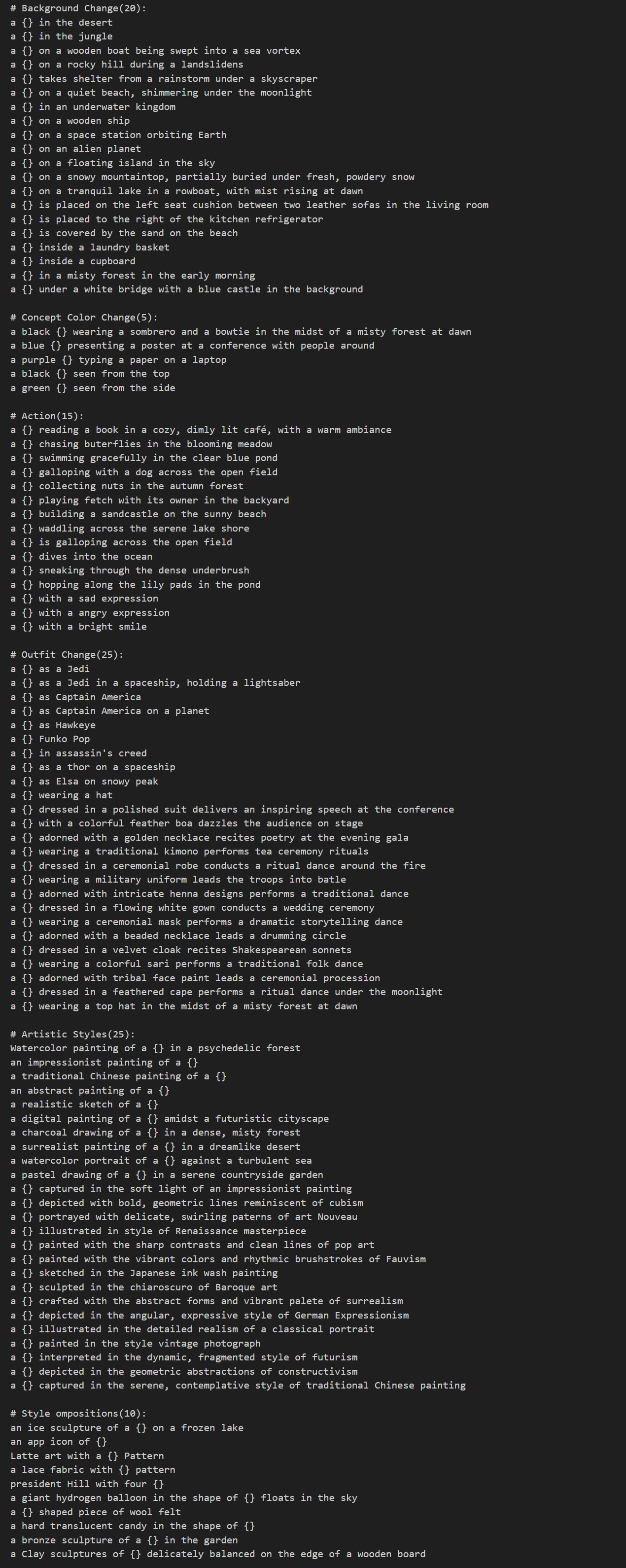}
    \caption{Prompt set for animate concepts.}
    \label{fig:prompt_animate}
\end{figure}

\begin{figure}[htbp]

   \includegraphics[width=1\linewidth]{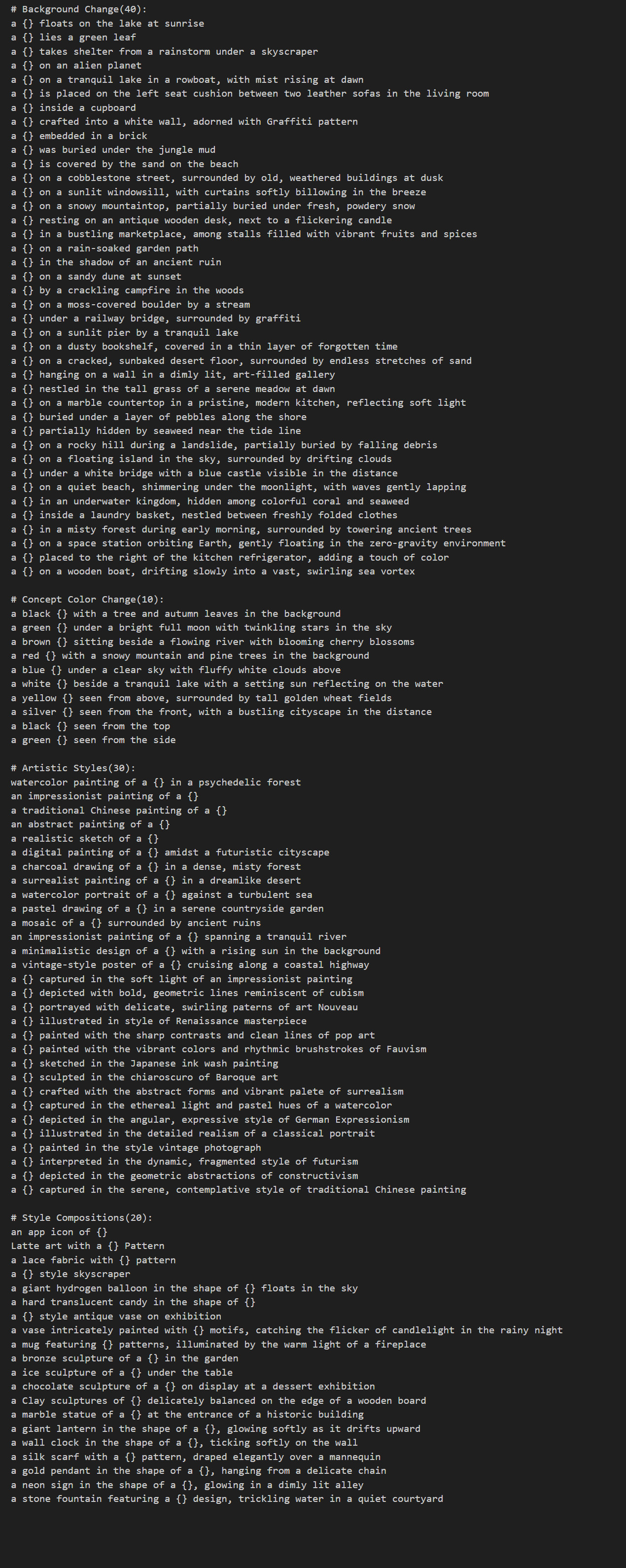} 
    \caption{Prompt set for inanimate concepts.}
    \label{fig:prompt_inanimate}
\end{figure}

\paragraph{Regularization Prompt Set.}
As previously mentioned, CoRe can be used with any prompt since it is applied only to the output embeddings and attention maps. To enhance the generalization of the new concept's text embedding, we design a regularization prompt set that covers a broad range of prompts. Similar to the prompts used for quantitative evaluation in Section~\ref{sec:appendix_prompt_list}, we employ two distinct prompts sets for animate and inanimate concepts, each containing 100 prompts. Specifically, the prompt set for inanimate concepts comprises four types of prompts: background change (e.g.,``a \{\} in the jungle''), concept color change (e.g.,``a black \{\} seen from the top''), artistic style (e.g.,``an abstract painting of a \{\}''), and style composition (e.g.,``a \{\} style skyscraper''). For animate concepts, two additional types are included: action and outfit change. The complete list of these prompt sets are provided in Figures~\ref{fig:prompt_inanimate} and~\ref{fig:prompt_animate}.

\begin{table}[t]
    \centering
    
    \caption{Quantitative ablation study. CLIP-I and DINO are used to evaluate identity preservation, while CLIP-T evaluates text alignment.}
        \begin{tabular}{lccc}
            \toprule
            Methods & CLIP-T$\uparrow$ & CLIP-I$\uparrow$ & DINO$\uparrow$ \\
            \midrule
            w/o CER & 0.2504& 0.6996& 0.5796\\
            w/o CAR & 0.2486& 0.7123&  0.6060\\
            w/o Rescale & 0.2561& 0.6968& 0.5805\\
            \midrule
            Full & 0.2568& 0.7054& 0.5842\\
            \bottomrule
        \end{tabular}
    \label{tab:ablation_evaluation}
\end{table}

\section{H\ \ \ User Study}
\begin{figure*}[t]
    \centering
    \includegraphics[width=0.72\linewidth]{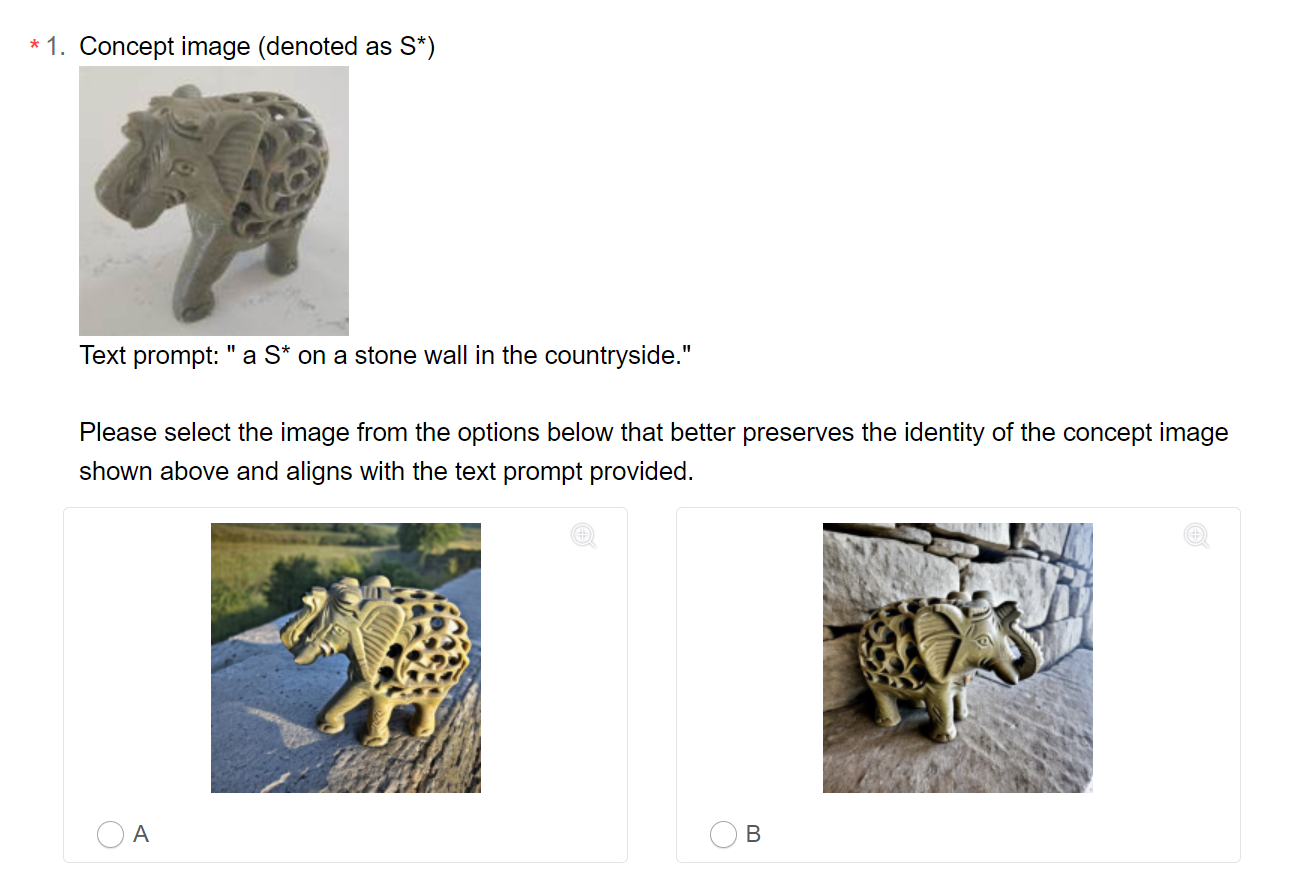}
    \vspace{-0.3cm}
    \caption{An example question of the user study.}
    \label{fig:user_study_example}
\end{figure*}

Figure~\ref{fig:user_study_example} illustrates an example question from the user study. Participants are presented with a concept image and a corresponding text prompt, followed by two generated images: one generated by our method and another by a baseline method. They are tasked with selecting the image that more accurately preserves the concept's identity and better aligns with the text prompt. The results are summarized in Table~\ref{tab:user_study}.



\clearpage
\renewcommand{\arraystretch}{1.3}
\begin{table*}[h]
\centering
\caption{Prompt list for quantitative evaluation, where ``\{\}'' represents $S_*$ or ``[V] class'' in some baselines.}
\begin{tabular}{p{8cm}p{8cm}}
\toprule
\textbf{Text prompts for Animate Objects} & \textbf{Text prompts for Inanimate Objects} \\ \midrule
a photo of a \{\} & a photo of a \{\} 
\\ \hline
a \{\} floats on a river & a \{\} floats on a river 
\\ \hline
a \{\} in Times Square & a \{\} in Times Square 
\\ \hline
a \{\} in the marketplace & a \{\} in the marketplace 
\\ \hline
a \{\} on a stone wall in the countryside & a \{\} on a stone wall in the countryside 
\\ \hline
a \{\} was buried at the bottom of the river 
& a \{\} was buried at the bottom of the river 
\\ \hline
a \{\} with the Eiffel Tower in the background 
& a \{\} with the Eiffel Tower in the background
\\ \hline
a red \{\} in the garden 
&  a red \{\} in the garden
\\ \hline
a pink \{\} by the lake 
& a pink \{\} by the lake
\\ \hline
a white \{\} seen from the bottom 
& a white \{\} seen from the bottom
\\ \hline
a curious \{\} exploring ancient ruins by the beach 
& a black \{\} seen from the back
\\ \hline
a brave \{\} crossing a shallow river in the wilderness& a purple \{\} on the dining table
\\ \hline
a \{\} leaping across rooftops in a city 
& a \{\} with a wheat field in the background
\\ \hline
a \{\} sitting on a floating island amidst the clouds, surrounded by flying birds 
& a \{\} in a snowy mountain landscape
\\ \hline
a cake with cream shaped like a \{\} on top 
& a \{\} depicted with rough, textured brushstrokes
\\ \hline
a \{\} reimagined as a character from a classic fairy tale 
& a \{\} depicted in the style of a Renaissance painting
\\ \hline
a \{\} wearing a superhero costume, complete with cape and mask & a \{\} painted like a Picasso abstract
\\ \hline
a \{\} painted in the style of an Abstract Expressionist painting& a \{\} painted in the style of an Abstract Expressionist painting
\\ \hline
a \{\} in the style of Van Gogh's Starry Night & a \{\} in the style of Van Gogh's Starry Night
\\ \hline
a \{\} with intricate Baroque carvings & a \{\} with intricate Baroque carvings
\\
\bottomrule
\end{tabular}
\label{tab:prompt_list}
\end{table*}

\clearpage

\clearpage

\end{document}